\documentclass[letterpaper, 10 pt, journal]{IEEEtran}
\IEEEoverridecommandlockouts

\usepackage[english]{babel}

\usepackage{cite}
\usepackage[hidelinks]{hyperref}
\hypersetup{
    colorlinks=true,
    linkcolor=green,
    filecolor=magenta,
    urlcolor=cyan,
    citecolor=blue
}
\usepackage{bm}
\usepackage{graphics}
\usepackage{epsfig}
\usepackage{multirow}
\usepackage{amssymb}
\usepackage{amsmath}

\usepackage{amsthm}
\usepackage{stmaryrd}
\usepackage{graphicx}
\graphicspath{{figures/}}
\usepackage[caption=false]{subfig}
\usepackage{float}

\usepackage[linesnumbered,ruled,vlined,algo2e,norelsize]{algorithm2e}
\usepackage{threeparttable}
\usepackage{booktabs}
\usepackage{optidef}
\def\matt#1{\begin{bmatrix}#1\end{bmatrix}}

\allowdisplaybreaks
\usepackage{soul} 

\newtheorem{assumption}{Assumption}

\newtheorem{remark}{Remark}

\begin{document}

\allowdisplaybreaks

\title{A Differentiable Dynamic Modeling Approach to Integrated Motion Planning and Actuator Physical Design for Mobile Manipulators}
\author{Zehui Lu, Yebin Wang
\thanks{Z. Lu is with the School of Aeronautics and Astronautics, Purdue University, IN 47907, USA. {\tt\small Email: lu846@purdue.edu} This work was done while Z. Lu was an intern at Mitsubishi Electric Research Laboratories (MERL), Cambridge, MA 02139, USA.}
\thanks{Y. Wang is with MERL. {\tt\small Email: yebinwang@ieee.org}}
}

\maketitle

\begin{abstract}
This paper investigates the differentiable dynamic modeling of mobile manipulators to facilitate efficient motion planning and physical design of actuators, where the actuator design is parameterized by physically meaningful motor geometry parameters.
These parameters impact the manipulator's link mass, inertia, center-of-mass, torque constraints, and angular velocity constraints, influencing control authority in motion planning and trajectory tracking control.
A motor's maximum torque/speed and how the design parameters affect the dynamics are modeled analytically, facilitating differentiable and analytical dynamic modeling.
Additionally, an integrated locomotion and manipulation planning problem is formulated with direct collocation discretization, using the proposed differentiable dynamics and motor parameterization.
Such dynamics are required to capture the dynamic coupling between the base and the manipulator. Numerical experiments demonstrate the effectiveness of differentiable dynamics in speeding up optimization and advantages in task completion time and energy consumption over established sequential motion planning approach.
Finally, this paper introduces a simultaneous actuator design and motion planning framework, providing numerical results to validate the proposed differentiable modeling approach for co-design problems. 
\end{abstract}

\IEEEpeerreviewmaketitle

\section{Introduction}\label{sec:introduction}

Mobile manipulators offer an evolution in robotic system architectures, enabling them to transition from automated systems to autonomous ones \cite{christensen2021roadmap}.
These systems combine mobility, provided by the mobile platform, with the manipulation capabilities of the mounted articulated arm, resulting in a versatile manipulation workspace \cite{thakar2023survey}.
Mobile manipulators have gained attention across various domains, including industrial settings like factories and warehouses \cite{yamashita2003motion,chen2018dexterous,thakar2020manipulator}, indoor environments such as healthcare \cite{li2017development}, and outdoor field applications like environment exploration \cite{stuckler2016nimbro,naazare2022online}, excavation \cite{jud2019autonomous}, and satellite service \cite{aghili2012prediction}.
Their flexibility makes them well-suited for undertaking and assisting with a wide range of tasks \cite{thakar2023survey}.
Nevertheless, mobile manipulators are expected to operate in complex, less structured, and dynamic environments, which presents unique challenges in system design, perception, motion (i.e. locomotion and manipulation) planning, and control.
These challenges have motivated researchers to explore various approaches in planning for mobile manipulators.

Motion planning is crucial in bridging the gap between perception and control in autonomous mobile manipulators across different deployment stages. This continuum is characterized by the level of feedback between perception and control, spanning from motion planning methodologies primarily used in offline settings to various forms of online replanning or closed-loop control suitable for navigating complex, less structured, and dynamic environments \cite{thakar2023survey}.
Motion planning typically involves predicting how the ego robot behaves based on a certain range of actions, and assessing the impacts of these actions on both the environment and the robot's subsequent decision-making process. The prediction usually relies on either a kinematic or dynamic model of the ego robot. The choice between these models depends on the required accuracy of the modeling and the specific mission requirements.

Over the past two decades, researchers have been focusing on developing computationally efficient methods for generating safe and robust motion planning in a complex environment.
The planning/control hierarchy of a robot typically consists of at least two levels \cite{WanHanAhn24}. A high-level planner, often a motion planner, generates a desired trajectory of velocity, acceleration, or torque for each actuator, which is then passed to a low-level controller. The low-level controller for each actuator is responsible for tracking the desired trajectory provided by the high-level planner.
However, the actual behavior of the robot may not always align exactly with what the motion planning algorithm anticipates due to the robot's modeling fidelity and the inherent dynamics of its actuators.
Based on existing literature, this discrepancy or uncertainty is somewhat manageable to some extent in specific domains. However, from the perspective of designers or manufacturers, this level of modeling and its subsequent motion planning is insufficient.
In other words, manufacturers must understand how critical design parameters influence the system dynamics, actuator capabilities, subsequent motion planning, and eventually closed-loop robot control. Given some performance metrics, these factors can assist in iteratively updating the design parameters to achieve optimal robot performance. The entire process is known as robot co-design.
Additionally, both the system dynamics and the impact of design parameters on the system dynamics are expected to be analytically modeled, which aims to facilitate solving some underlying optimization problems related to motion planning or robot design given existing numerical gradient-based optimization solvers.

To summarize, is it necessary to understand how major design parameters analytically influence the system dynamics and actuator capabilities, and how these factors affect motion planning subsequently.
This paper investigates the analytical and differentiable dynamic modeling of mobile manipulators with motor parameterization given physically meaningful motor geometry parameters, which enables integrated motion (locomotion and manipulation) planning and actuator design. This paper also studies how the design parameters affect the actuators' capabilities and subsequently the motion planning given the parameterized dynamics model.
The differentiability in this paper means that the gradient of the system dynamics regarding the underlying optimization decision variables, such as the system's states and inputs, is analytical.

\subsection{Literature Review}

\subsubsection{Motion Planning} \label{subsec:literature_motion_planning}
The mobile base and manipulator have significantly different dynamic characteristics, as the mobile base generally has higher inertia than the manipulator.
Despite this disparity, these two systems are strongly coupled, leading to complex dynamic behaviors.
These characteristics compound the complexity of the planning problem.

Various approaches have been employed to solve the motion planning problem for mobile manipulators. 
The planning algorithms are generally categorized into two classes, i.e. combined or separate planning, based on whether they consider the behavioral differences between the mobile base and the robot manipulator or not \cite{sandakalum2022motion}.

For separate planning, a complex task is often divided into a sequence of sub-tasks, and planning is conducted separately for each sub-task.
This approach essentially decouples the planning for the mobile base and the robot manipulator due to the lack of a dynamic model of the entire mobile manipulator. Nevertheless, it reduces the complexity of planning over a high-dimensional space. While various planning algorithms exist in the literature for locomotion (mobile base) \cite{tzafestas2018mobile} and manipulation planning \cite{berenson2009manipulation,berenson2011task,zucker2013chomp, rickert2014balancing, gold2022model, Holmes-RSS-20, Michaux-RSS-23,marcucci2023motion, michaux2023can, brei2023serving}, they have not been specifically implemented for mobile manipulators.
When motion planning is separately performed for the mobile base and the manipulator, the existing algorithms for each component can be utilized once a goal configuration is determined for both the mobile base and the manipulator.
The separation of locomotion and manipulation results in inefficient and suboptimal solutions even if optimal solutions are obtained for each sub-task \cite{sandakalum2022motion}. Furthermore, the separation may lead to infeasibility since poor base placement could render the final goal state unreachable \cite{correll2016analysis,sandakalum2022motion}.

As for the combined planning, existing methods can be categorized into several subsets based on utilizing different levels of model information.
First, some methods rely on the kinematic model of a simple mobile manipulator to design position-tracking controllers \cite{de2006kinematic} and perform trajectory planning \cite{tang2010differential}. Particularly, \cite{tan2003integrated} perform position-tracking control based on a kinematic model of the base and a dynamic model of the manipulator. 
Also, \cite{furuno2003trajectory} derive the dynamics for a two-wheel differential unicycle with a two-link manipulator, and proposes an optimization-based trajectory planning with constant joint velocity limits.

The majority of the literature relies on fast sampling/searching \cite{kavraki1996probabilistic,LaVKuf01,vannoy2008real,stilman2010global,cohen2010search,chitta2010planning,hornung2012navigation,rickert2014balancing,WanJhaAke17,pilania2018mobile,thakar2020manipulator,naazare2022online,honerkamp2023n} or optimizing \cite{berenson2008optimization,zucker2013chomp,schulman2014motion,jud2019autonomous,maric2019fast,ZhaWanZho19,thakar2020manipulator} over a generally high-dimensional configuration space or the task space (i.e. the Euclidean space) with kinematic constraints, where the collision avoidance is typically implemented by checking the intersection between the forward occupancy (volume) of the robot and the volume of obstacles.
Since implementing the forward occupancy or enforcing kinematic constraints only requires a kinematic model, there is no essential difference in doing so for a mobile manipulator or a fixed 6-DOF (degree of freedom) manipulator.
Consequently, a mobile manipulator is considered a single system despite the behavioral difference between the mobile base and the manipulator.
Due to the absence of dynamics or actuator information, the motion trajectory may not always be dynamically feasible. As a result, one can only assume that, given empirical kinematic constraints on joint velocities and accelerations, the desired inputs for actuators are always feasible.
Thus, at run-time, the actual trajectory may deviate from the expected one from motion planning.
On the other hand, these empirical kinematic constraints could also significantly affect the operation speed of robots.
This effect is especially noticeable for mobile manipulators, where limiting acceleration and velocity is necessary in the literature to minimize the jerk caused by sudden movements and the manipulator's sway while the base is in motion.

To reduce this discrepancy, \cite{kousik2017safe} and \cite{chen2021fastrack} employ a high-fidelity dynamic model for offline modeling error computation and a low-fidelity kinematic/dynamic model for real-time planning.
These methods incorporate pre-computed modeling errors between two models to reduce real-time computational burden and address safety concerns arising from inaccuracies in real-time planning.
Additionally, some methods such as \cite{michaux2023can} and \cite{brei2023serving} utilize two models of a fixed manipulator with low and high fidelity for real-time motion planning and control.
Therefore, a high-fidelity dynamic model for a mobile manipulator is essential for fast and high-precision motion planning in safety-critical scenarios, as such a model accurately characterizes the dynamic coupling between the base and the manipulator.
Moreover, from the manufacturers' perspective, understanding how major design parameters analytically affect the dynamics and actuator capabilities is necessary.
This understanding allows for the consideration of these factors in subsequent motion planning, ensuring that the planned trajectories are dynamically feasible given the physical limitations of the system and its actuators.

\subsubsection{Dynamic Modeling}

According to the above literature review on motion planning, dynamic modeling for mobile manipulators is essential for accurate and safe motion planning, as well as robot design problems.
The literature on dynamic modeling of mobile manipulators can be generally categorized into two classes, i.e. modeling for a particular type of mobile manipulators or general multiple rigid bodies.

Regarding the dynamic modeling for some particular mobile manipulators, \cite{tang2010differential} and \cite{de2006kinematic} develop kinematic models for differential unicycles with a two-link/three-link manipulator. \cite{tan2003integrated} combine the kinematics of a nonholonomic cart with the dynamics of a three-link manipulator.
\cite{furuno2003trajectory} discuss how to model the dynamics of a differential unicycle with a two-link manipulator.
\cite{korayem2018derivation} develop a dynamic model for a nonholonomic cart with a two-flexible-link manipulator by recursive Gibbs–Appell (Lagrange) formulations.
The above works model the kinematics or dynamics of the mobile base as a simple differential unicycle, where the contact between the wheels and the ground is not considered.
In contrast, \cite{seegmiller2016high} investigate the high-fidelity modeling of a four-wheeled robot, including the modeling of contact forces between the wheels and the ground, as well as how the terrain affects the robot's dynamics.

As for the analytical modeling methodologies for general multibody dynamics in the literature, \cite{featherstone2014rigid} proposes the Articulated Body Algorithm (ABA) to efficiently and analytically generate forward dynamics for a tree-like chain of articulated links.
\cite{Carpentier-RSS-18} investigate the analytical and symbolic computation of derivatives for the analytical forward dynamics given the ABA.
\cite{carpentier2019pinocchio} propose Pinocchio, a fast implementation of rigid body dynamics algorithms and their analytical derivatives. Pinocchio can take a robot's $\tt{urdf}$ configuration file as input to generate its analytical dynamics. While Pinocchio is widely used in the research community, it cannot analytically embed actuators into the system dynamics, as they are implicitly included in the mass, inertia, and center-of-mass (CoM) of each link in an $\tt{urdf}$ file. 
\cite{stein2023application} propose a revised ABA algorithm for calculating the forward dynamics of a fixed 6-DOF manipulator, where the motor dynamics are coupled with the manipulator dynamics. Although the dynamics' fidelity is relatively high, combining the slow manipulator dynamics ($\sim$ 100 Hz) with the fast motor dynamics (over 2000 Hz) can significantly slow down any computation that involves dynamics forward propagation as it requires a smaller discretization.
Hence, for the sake of computational efficiency, it is necessary to analytically understand the motors' capacity and how they affect the entire robot without including their faster dynamics.
Therefore, this paper aims to investigate how to include this information analytically into the dynamics of a mobile manipulator.

\subsection{Paper Organization and Contributions}

The rest of this paper is organized as follows:
Section \ref{sec:prelim} introduces necessary notations and preliminary definitions;
Section \ref{sec:define_mobile} presents the definition of a class of mobile manipulators and all its necessary components;
Section \ref{sec:motor_para} introduces the parameterization of servomotors given motor geometry parameters, as well as presents both the numerical and analytical modeling of motor torque capacity;
Section \ref{sec:system_modeling} presents the forward and inverse dynamics modeling of a mobile manipulator and cross-validates the algorithms;
Section \ref{sec:joint_locomotion_planning} formulates an integrated locomotion and manipulation planning optimization problem with discretization by direct collocation;
Section \ref{sec:numerical_exp} presents some numerical experiments for the integrated planning method and a benchmark method;
Section \ref{sec:simultaneous_design} showcases a simultaneous actuator design and motion planning framework with some numerical results;
Section \ref{sec:conclusion} discusses the limitations of the proposed modeling approach and concludes this paper.

The contributions of this paper are summarized as follows:
\begin{enumerate}
\item An analytical modeling approach for a manipulator's actuators based on motor design parameters;
\item A differentiable and analytical modeling approach for mobile manipulators given actuators parameterized by motor design parameters;
\item An analytical modeling approach for a motor's maximum speed/torque as a function of its design parameters;
\item An integrated locomotion and manipulation planning approach with motor torque/speed constraints and direct collocation discretization;
\item A framework of simultaneous actuator design and integrated motion planning for mobile manipulator co-design.
\end{enumerate}

\section{Preliminaries} \label{sec:prelim}
This section introduces necessary notations and preliminary definitions.

\subsection{Notations}
Denote $\mathbb{R}$ as the real number set and $\mathbb{R}_+$ as the positive real number set. Denote $\mathbb{Z}_+$ as the positive integer set.
Denote $\mathbb{E}^{m}$ as the $m$-dimensional Euclidean vector space.
Denote $\mathrm{SO}(3)$ as the Special Orthogonal Group associated with $\mathbb{E}^{3}$.
Denote $\mathrm{SE}(3)$ as the Special Euclidean Group associated with $\mathbb{E}^{3}$.
For $\boldsymbol{x}, \boldsymbol{y} \in \mathbb{R}^n$, $\boldsymbol{x} \leq \boldsymbol{y}$ indicates element-wise inequality. Let $\text{col}\{ \boldsymbol{v}_1, \cdots, \boldsymbol{v}_a \}$ denote a column stack of elements $\boldsymbol{v}_1, \cdots, \boldsymbol{v}_a $, which may be scalars, vectors or matrices, i.e. $\text{col}\{ \boldsymbol{v}_1, \cdots, \boldsymbol{v}_a \} \triangleq {\matt{{\boldsymbol{v}_1}^{\top} & \cdots & {\boldsymbol{v}_a}^{\top}}}^{\top}$. 
Let $\boldsymbol{0}_n, \boldsymbol{1}_n \in \mathbb{R}^{n}$ denote a zero and an one vector.
Let $\boldsymbol{I}_n \in \mathbb{R}^{n \times n}$ denote an identity matrix.
Let $\llbracket a,b \rrbracket$ denote a set of all integers between $a \in \mathbb{Z}$ and $b \in \mathbb{Z}$, with both ends included.
Denote $\mathrm{diag}(a_1, \cdots, a_n)$ as a diagonal matrix in $\mathbb{R}^{n\times n}$ with diagonal elements $a_1, \cdots, a_n$.
Given a matrix $\boldsymbol{A} \in \mathbb{R}^{n \times m}$, $\boldsymbol{A}[a:b, c] \in \mathbb{R}^{(b-a+1)}$ ($1 \leq a \leq b \leq n, 1 \leq c \leq m$) represents a vector slice of $\boldsymbol{A}$ from $c$-th column, $a$-th row until $b$-th row; $\boldsymbol{A}[a:b, c:d] \in \mathbb{R}^{(b-a+1) \times (d-c+1)}$ ($1 \leq a \leq b \leq n, 1 \leq c \leq d \leq m$) represents a matrix slice of $\boldsymbol{A}$ from $c$-th column until $d$-th column, $a$-th row until $b$-th row.

\subsection{Homogeneous Transformation} \label{subsec:prelim:homo_transformation}
Let $\boldsymbol{T}_{L_1, L_2} \in \mathrm{SE}(3)$ be a homogeneous transformation. Its subscript indicates that $\boldsymbol{T}_{L_1, L_2}$ is a transformation from an inertia frame $\{L_2\}$ to an inertia frame $\{L_1\}$. Let $\boldsymbol{p}_{L_2}$ be a homogeneous coordinate of a point in $\mathbb{R}^3$ in the frame $\{L_2\}$, then $\boldsymbol{T}_{L_1, L_2} \cdot \boldsymbol{p}_{L_2}$ is the homogeneous coordinate of the same point in the frame $\{L_1\}$.
Let $\boldsymbol{T}_{L_2, L_3} \in \mathrm{SE}(3)$ be a homogeneous transformation from an inertia frame $\{L_3\}$ to the inertia frame $\{L_2\}$. Then $\boldsymbol{T}_{L_1, L_3} = \boldsymbol{T}_{L_1, L_2} \cdot \boldsymbol{T}_{L_2, L_3}$ is the homogeneous transformation from the frame $\{L_3\}$ to the frame $\{L_1\}$. Denote $\boldsymbol{M}_{L_1,L_2} \in \mathrm{SE}(3)$ be a homogeneous transformation with the same linear translation of $\boldsymbol{T}_{L_1, L_2}$ and zero rotation, i.e. $\boldsymbol{M}_{L_1,L_2}[1:3, 1:3] = \boldsymbol{I}_3$ and $\boldsymbol{M}_{L_1,L_2}[1:3, 4] = \boldsymbol{T}_{L_1,L_2}[1:3, 4]$.

\subsection{Euclidean Cross Operator}
For two vectors $\boldsymbol{a}, \boldsymbol{b} \in \mathbb{R}^3$, the cross product $\boldsymbol{a} \times \boldsymbol{b}$ is equivalent to a linear operator $[\boldsymbol{a}\times]$ that maps $\boldsymbol{b}$ to $\boldsymbol{a} \times \boldsymbol{b}$.
The operator $\boldsymbol{a}\times$ is given by
\begin{equation} \label{eq:prelim:cross_euclidean}
\boldsymbol{a}\times = \matt{x \\ y \\ z} \times = \matt{0 & -z & y \\ z & 0 & -x \\ -y & x & 0}.
\end{equation}
If $\lambda \in \mathbb{R}$, then $(\lambda \boldsymbol{a}) \times = \lambda (\boldsymbol{a}\times)$. The matrix in \eqref{eq:prelim:cross_euclidean} is a $3 \times 3$ skew-symmetric matrix representation of vector $\boldsymbol{a}$.

\subsection{Spatial Cross Operator}
Given a homogeneous transformation $\boldsymbol{T} = \matt{\boldsymbol{R} & \boldsymbol{p} \\ \boldsymbol{0}_{1\times3} & 1} \in \mathrm{SE}(3)$, its adjoint representation $[\mathrm{Ad}_{\boldsymbol{T}}]$ is
\begin{equation}
[\mathrm{Ad}_{\boldsymbol{T}}] = \matt{\boldsymbol{R} & \boldsymbol{0}_{3 \times 3} \\ [\boldsymbol{p}\times] \boldsymbol{R} & \boldsymbol{R}} \in \mathbb{R}^{6 \times 6},
\end{equation}
where $\boldsymbol{R} \in \mathrm{SO}(3)$ and $\boldsymbol{p} \in \mathbb{R}^3$. $[\mathrm{Ad}_{\boldsymbol{T}}]$ is a linear operator and has the following properties:
\begin{equation}
[\mathrm{Ad}_{\boldsymbol{T}^{-1}}] = [\mathrm{Ad}_{\boldsymbol{T}}]^{-1}, \quad [\mathrm{Ad}_{\boldsymbol{T}_1 \boldsymbol{T}_2}] = [\mathrm{Ad}_{\boldsymbol{T}_1}] [\mathrm{Ad}_{\boldsymbol{T}_2}].
\end{equation}
Denote a spatial twist $\boldsymbol{\mathcal{V}} = \matt{\boldsymbol{\omega} \\ \boldsymbol{v}} \in \mathbb{R}^6$ with angular velocity $\boldsymbol{\omega} \in \mathbb{R}^3$ and linear velocity $ \boldsymbol{v} \in \mathbb{R}^3$ in its inertia frame $\{L_2\}$.
Given a homogeneous transformation $\boldsymbol{T}_{L_1, L_2}$,
its adjoint operator applying on a spatial twist, i.e. $[\mathrm{Ad}_{\boldsymbol{T}_{L_1, L_2}}] \boldsymbol{\mathcal{V}}$, is a transformation of the spatial twist from the inertia frame $\{L_2\}$ to the inertia frame $\{L_1\}$.

A spatial cross operator $[\boldsymbol{\mathcal{V}} \times^*]$ is given by
\begin{equation}
[\boldsymbol{\mathcal{V}} \times^*] \triangleq [\mathrm{ad}_{\boldsymbol{\mathcal{V}}}] = \matt{ \boldsymbol{\omega}\times & \boldsymbol{0}_{3\times 3} \\ \boldsymbol{v}\times & \boldsymbol{\omega}\times } \in \mathbb{R}^{6 \times 6}.
\end{equation}
The spatial cross operator $[\boldsymbol{\mathcal{V}} \times^*]$ can be viewed as a differentiation operator that maps from a spatial force to the derivative of the spatial force.

\subsection{Screw Axis to Homogeneous Transformation}
Given a screw axis $\boldsymbol{\mathcal{S}} = \matt{\boldsymbol{\omega} \\ \boldsymbol{v}} \in \mathbb{R}^6$ with $||\boldsymbol{\omega}|| = 1$, for any angular distance $\theta \in \mathbb{R}$ traveled around the axis $\boldsymbol{\mathcal{S}}$, the corresponding homogeneous transformation matrix $\boldsymbol{T}(\boldsymbol{\mathcal{S}}, \theta) \in \mathrm{SE}(3)$ is
\begin{equation} \label{eq:screw_axis_trans}
\boldsymbol{T}(\boldsymbol{\mathcal{S}}, \theta) = e^{[\boldsymbol{\mathcal{S}}]\theta} = \matt{ \boldsymbol{R}(\boldsymbol{\omega}, \theta) & \boldsymbol{p}(\boldsymbol{\mathcal{S}}, \theta) \\ \boldsymbol{0}_{1\times 3} & 1},
\end{equation}
where $\boldsymbol{R}(\boldsymbol{\omega}, \theta) = \boldsymbol{I}_{3} + \mathrm{sin}\theta [\boldsymbol{\omega} \times] + (1-\mathrm{cos}\theta) [\boldsymbol{\omega} \times]^2$, and $\boldsymbol{p}(\boldsymbol{\mathcal{S}}, \theta) = (\boldsymbol{I}_{3}\theta + (1-\mathrm{cos}\theta)[\boldsymbol{\omega} \times] + (\theta-\mathrm{sin}\theta)[\boldsymbol{\omega} \times]^2)\boldsymbol{v}$.
The matrix exponential $e^{[\boldsymbol{\mathcal{S}}]\theta}$ converts a rotation around the screw axis into a homogeneous transformation in $\mathrm{SE}(3)$.

\section{Mobile Manipulator} \label{sec:define_mobile}
\begin{figure}[!ht]
\centering
\includegraphics[width=0.45\linewidth]{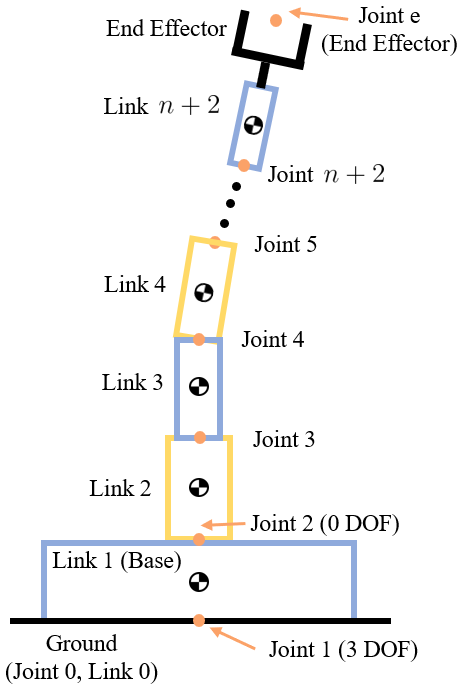}
\caption{Definition of a mobile manipulator. Joint 3 to Joint $n+2$ are all 1-DOF joints; Joint e is only for notation and has 0-DOF.}
\label{fig:mobile_manipulator_def}
\end{figure}

This paper considers a class of mobile manipulators given in Fig. \ref{fig:mobile_manipulator_def}. Suppose that the degree of freedom (DOF) of the manipulator is $n \in \mathbb{Z}_+$. Link 2 is the first (fixed) link of the manipulator. Joint 3 is the first joint of the manipulator. Link 2 is installed rigidly on Link 1, a mobile base, and thus Joint 2 is a 0-DOF fixed joint.
Joint 1 is defined as a 3-DOF planar joint, which includes the linear translation along the x-axis and the y-axis and the rotation around the z-axis. Joint 1 connects Link 1 with the ground, which is also denoted as Joint 0 and Link 0. From Joint 3 to Joint $2+n$, each joint is a 1-DOF revolute joint along a screw axis.

\begin{figure}[!ht]
\centering
\includegraphics[width=0.50\linewidth]{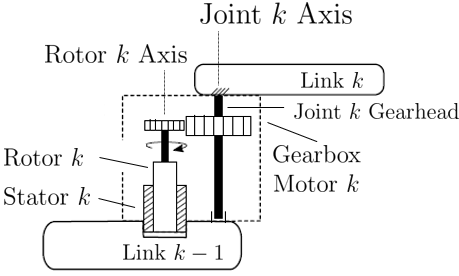}
\caption{Scheme of a geared motor between Link $k-1$ and Link $k$ ($k=3, \cdots, n+2$).}
\label{fig:geared_motor_scheme}
\end{figure}

As illustrated in Fig.~\ref{fig:geared_motor_scheme}, for each Joint $k$, $k = 3, \cdots, n+2$, there is a motor installed, denoted by Motor $k (k \geq 3)$. In detail, for each $k = 3, \cdots, n+2$, the stator of Motor $k$ is installed rigidly on Link $k-1$ and the rotor of Motor $k$ connects with Link $k$ through Joint $k$'s gearbox.

\begin{remark} \label{remark:mobile_base}
This paper models the mobile base as one rigid body with one 3-DOF planar joint on $\mathrm{SE}(2)$ because this paper focuses on the modeling methodology and algorithm of the forward and inverse dynamics of the entire chain of rigid bodies given motor parameterization.
To model the full 6-DOF motion of a mobile base, including pitch and roll movements, one has to define a particular mechanical configuration for the mobile base, including how motors are attached to the wheels, and how wheels are attached to the base and interact with the ground, which introduces extra modeling complexity and is out of the scope of this paper.
\end{remark}

\subsection{Joints}
Before introducing the forward dynamics, the kinematics of each joint and link need to be defined first. If Joint 1 is just a naive 3-DOF joint allowing movement in $\mathrm{SE}(2)$, its motion subspace matrix in Joint 1's inertia frame $\{J_1\}$ is given by
\begin{equation} \label{eq:joint_1_motion_subspace_naive}
\boldsymbol{S}_{J_1}= \matt{0 & 0 & 1 & 0 & 0 & 0 \\ 0 & 0 & 0 & 1 & 0 & 0 \\ 0 & 0 & 0 & 0 & 1 & 0}^{\top} \in \mathbb{R}^{6 \times 3}.
\end{equation}
The position and velocity variables of Joint 1 are given by
\begin{equation}
\boldsymbol{q}_1 \triangleq \matt{\theta_z & p_x & p_y}^{\top}, \quad \dot{\boldsymbol{q}}_1 \triangleq \matt{\omega_z & v_x & v_y}^{\top}.
\end{equation}

Since Joint 2 is a fixed joint with 0-DOF, its motion subspace matrix in Joint 2's inertia frame $\{J_2\}$ is given by
\begin{equation} \label{eq:joint_2_motion_subspace}
\boldsymbol{S}_{J_2} = \boldsymbol{0}_{6}.
\end{equation}
And its position and velocity variables $\boldsymbol{q}_2$ are given by
\begin{equation}
\boldsymbol{q}_2 \equiv 0 \equiv \dot{\boldsymbol{q}}_2.
\end{equation}

For Joint $k = 3, \cdots, n+2$, its position and velocity variables are given by $\boldsymbol{q}_k \triangleq \theta_k \in \mathbb{R}$ and $\dot{\boldsymbol{q}}_k \triangleq \dot{\theta}_k \in \mathbb{R}$, where $\theta_k$ and $\dot{\theta}_k$ are the joint angular position and velocity around the screw axis of Joint $k$, respectively. For Joint $k = 3, \cdots, n+2$, its motion subspace matrix in Joint $k$'s inertia frame $\{J_k\}$ is equivalent to its screw axis. For Joint $k$ rotating around its y-axis and z-axis of the frame $\{J_k\}$, $\boldsymbol{S}_{J_k} = \matt{ 0 & 1 & 0 & 0 & 0 & 0 }^{\top}$ and $\boldsymbol{S}_{J_k} = \matt{ 0 & 0 & 1 & 0 & 0 & 0 }^{\top}$, respectively.

By \cite[Chapter~3.5]{featherstone2014rigid}, the apparent derivative of Joint $k$'s motion subspace matrix $\boldsymbol{S}_{J_k}$ is defined by
\begin{equation} \label{eq:apparent_deri}
\mathring{\boldsymbol{S}}_{J_k} = \frac{\partial \boldsymbol{S}_{J_k}}{\partial t} + \sum_{j=1}^{n_k} \frac{\partial \boldsymbol{S}_{J_k}}{\partial \boldsymbol{q}_k [j]} \dot{\boldsymbol{q}}_k [j],
\end{equation}
where $n_k$ is the dimension of Joint $k$'s position variable $\boldsymbol{q}_k$; $\boldsymbol{q}_k [j]$ denotes the $j$-th element of $\boldsymbol{q}_k$.
Denote $\mathring{\boldsymbol{S}}_{J_k}$ as the apparent derivative of $\boldsymbol{S}_{J_k}$ in the frame $\{ J_k \}$, and we have
\begin{equation} \label{eq:motion_subspace_deri}
\mathring{\boldsymbol{S}}_{J_1} = \boldsymbol{0}_{6 \times 3}, \quad \mathring{\boldsymbol{S}}_{J_k} = \boldsymbol{0}_{6}, \ \forall k = 2, \cdots, n+2.
\end{equation}

Given the class of mobile manipulators defined in Fig. \ref{fig:mobile_manipulator_def}, the linear transformation from $\{J_k\}$ to $\{J_{k-1}\}$ is given by
\begin{equation}
\boldsymbol{M}_{J_{k-1},J_k}= \matt{\boldsymbol{I}_{3} & \boldsymbol{p}_{J_{k-1},J_k} \\ \boldsymbol{0}_{1\times 3} & 1}, k=3, \cdots, n+2,
\end{equation}
where $\boldsymbol{p}_{J_{k-1},J_k} \in \mathbb{R}^3$ indicates the position coordinates of Joint $k$ in the frame $\{J_{k-1}\}$.

\subsection{Links}
Given a specific kinematic chain, the homogeneous transformation from Link 1's inertia frame $\{L_1\}$ to Link 0's inertia frame $\{L_0\}$ (or the global frame), is given by
\begin{equation}
\boldsymbol{T}_{L_0, L_1}(\boldsymbol{q}_1) = \matt{\mathrm{cos}(\theta_z) & -\mathrm{sin}(\theta_z) & 0 & p_x \\ \mathrm{sin}(\theta_z) & \mathrm{cos}(\theta_z) & 0 & p_y \\ 0 & 0 & 1 & h_1 \\ 0 & 0 & 0 & 1},
\end{equation}
where $h_1 > 0$ is the height of CoM of the mobile base (Link 1) from the global frame. $\boldsymbol{T}_{L_1, L_0} = (\boldsymbol{T}_{L_0, L_1})^{-1}$.

The transformation from Link $k$'s inertia frame $\{L_k\}$ to the frame $\{J_k\}$ is given by
\begin{equation}
\boldsymbol{M}_{J_k, L_k} = \matt{\boldsymbol{I}_{3} & \boldsymbol{p}_{J_k, L_k} \\ \boldsymbol{0}_{1\times 3} & 1}, k = 1, \cdots, n+2,
\end{equation}
where $\boldsymbol{p}_{J_k, L_k} \in \mathbb{R}^3$ indicates the position coordinates of Link $k$ in the frame $\{J_k\}$.
Note that for $k = 1$, $\boldsymbol{p}_{J_1, L_1}[3] = h_1$.

The homogeneous transformation from $\{J_2\}$ to $\{L_1\}$ is given by
\begin{equation}
\boldsymbol{T}_{L_1, J_2} = \boldsymbol{M}_{L_1, J_2} = \matt{\boldsymbol{I}_3 & \boldsymbol{p}_{L_1, J_2} \\ \boldsymbol{0}_{1\times 3} & 1},
\end{equation}
where $\boldsymbol{p}_{L_1, J_2} \in \mathbb{R}^3$ is the position coordinates of Joint 2 in the frame $\{L_1\}$.

The homogeneous transformation from the frame $\{L_2\}$ to the frame $\{L_1\}$ is given by
\begin{equation}
\boldsymbol{T}_{L_1, L_2} = \boldsymbol{T}_{L_1, J_2} \cdot \boldsymbol{T}_{J_2, L_2} = \matt{\boldsymbol{I}_3 & \boldsymbol{p}_{L_1, J_2} + \boldsymbol{p}_{J_2, L_2} \\ \boldsymbol{0}_{1\times 3} & 1}.
\end{equation}
Thus, $\boldsymbol{T}_{L_2, L_1} = (\boldsymbol{T}_{L_1, L_2})^{-1} = \boldsymbol{M}_{L_2, L_1}$ since there is no rotation between Link 1 and Link 2.

For $k=3,\cdots, n+2$, the transformation of the frame $\{L_{k-1}\}$ to the frame $\{L_k\}$ is given by
\begin{equation}
\begin{split}
\boldsymbol{M}_{L_k, L_{k-1}} &= (\boldsymbol{M}_{J_k, L_k})^{-1} (\boldsymbol{M}_{J_{k-1}, J_k})^{-1} \boldsymbol{M}_{J_{k-1}, L_{k-1}} \\
&=(\boldsymbol{M}_{J_{k-1}, J_k} \boldsymbol{M}_{J_k, L_k} )^{-1} \boldsymbol{M}_{J_{k-1}, L_{k-1}}.
\end{split}
\end{equation}

\subsection{Inertia of Links}
The spatial inertia matrix $\hat{\boldsymbol{G}}_{L_k} \in \mathbb{R}^{6\times 6}$ of Link $k$ in the frame $\{L_k\}$ is given by
\begin{equation}
\hat{\boldsymbol{G}}_{L_k} = \matt{\boldsymbol{I}_{k,\mathrm{G}} & \boldsymbol{0}_{3\times 3} \\ \boldsymbol{0}_{3\times 3} & m_k \boldsymbol{I}_{3} },
\end{equation}
where $\boldsymbol{I}_{k,\mathrm{G}} \in \mathbb{R}^{3\times 3}$ is the inertia tensor in matrix form for Link $k$; $m_k$ is the mass of Link $k$.

The stator of Motor $k (k \geq 3)$ is mounted on Link $k-1$ and hence each link's inertia should add the effect from the stator.
The mass and inertia of a manipulator's link are typically greater than the ones of a motor stator due to the requirements of the mechanical design of the manipulator. Therefore, to simplify the parameterization of a manipulator, this paper adopts the following assumption on the CoM of Link $k-1$.
\begin{assumption}[\bf{Constant Link CoM}] \label{assump:constant_link_com}
For $k \geq 3$, the CoM of Link $k-1$ with Motor $k$'s stator installed on the end of Link $k-1$ stays the same as the CoM of Link $k-1$.
\end{assumption}

Suppose the inertia tensor matrix for the stator of Motor $k$ ($k \geq 3$) is $\boldsymbol{I}_{k,\mathrm{S}} \in \mathbb{R}^{3\times 3}$ and the stator mass is $m_{k, \mathrm{S}}$.
With Assumption \ref{assump:constant_link_com}, the spatial inertia matrix $\boldsymbol{G}_{L_{k-1}}$ of Link $k-1$ in the frame $\{L_{k-1}\}$ is given by
\begin{equation}
\boldsymbol{G}_{L_{k-1}} = \hat{\boldsymbol{G}}_{L_{k-1}} +  \matt{\hat{\boldsymbol{I}}_{k,\mathrm{S}} & \boldsymbol{0}_{3\times 3} \\ \boldsymbol{0}_{3\times 3} & m_{k, \mathrm{S}} \boldsymbol{I}_{3} }, \ k = 3, \cdots, n+2,
\end{equation}
where $\hat{\boldsymbol{I}}_{k,\mathrm{S}} = \boldsymbol{I}_{k,\mathrm{S}} + m_{k, \mathrm{S}} \cdot \mathrm{diag}(l_{\mathrm{x}}, l_{\mathrm{y}}, l_{\mathrm{z}})$;
$l_{\mathrm{x}} = (\boldsymbol{d}_{k,\mathrm{c2t}}[2])^2 + (\boldsymbol{d}_{k,\mathrm{c2t}}[3])^2$,
$l_{\mathrm{y}} = (\boldsymbol{d}_{k,\mathrm{c2t}}[1])^2 + (\boldsymbol{d}_{k,\mathrm{c2t}}[3])^2$,
$l_{\mathrm{z}} = (\boldsymbol{d}_{k,\mathrm{c2t}}[1])^2 + (\boldsymbol{d}_{k,\mathrm{c2t}}[2])^2$; $\boldsymbol{d}_{k,\mathrm{c2t}} \in \mathbb{R}^3$ is the position coordinate of Link $k-1$'s link tip in the frame $\{L_k\}$, i.e. the location where Motor $k$'s stator is mounted on.
For Link 1 and Link $n+2$, $\boldsymbol{G}_{L_1} = \hat{\boldsymbol{G}}_{L_1}$, $\boldsymbol{G}_{L_{n+2}} = \hat{\boldsymbol{G}}_{L_{n+2}}$.

\subsection{Homogeneous Transformation} \label{subsec:homo_transformation_define}
To calculate rigid-body mechanics, Joint $k$'s motion subspace matrix $\boldsymbol{S}_{J_k}$ in the frame $\{J_k\}$ needs to be transformed into the matrix $\boldsymbol{A}_{L_k}$ in the frame $L_k$, i.e.
\begin{equation}
\boldsymbol{A}_{L_k} = [\mathrm{Ad}_{(\boldsymbol{M}_{J_k,L_k})^{-1}}] \boldsymbol{S}_{J_k}, \  k = 1, \cdots, n+2.
\end{equation}
Similarly,
\begin{equation}
\mathring{\boldsymbol{A}}_{L_k} = [\mathrm{Ad}_{(\boldsymbol{M}_{J_k,L_k})^{-1}}] \mathring{\boldsymbol{S}}_{J_k}, \  k = 1, \cdots, n+2.
\end{equation}
Note that $\boldsymbol{A}_{L_2} = \boldsymbol{0}_6$, $\mathring{\boldsymbol{A}}_{L_k} = \boldsymbol{0}_6$ for all $k = 2, \cdots, n+2$.

Determining the transformation between the screw axis $\boldsymbol{S}_{R_k} \in \mathbb{R}^6$ of Motor $k$'s rotor and the screw axis of Joint $k$ ($k \geq 3$) requires a specific mechanical configuration of a gearbox, which further introduces the modeling complexity.
Due to the absence of detailed modeling of gearboxes, this paper adopts the following assumption. Since the gearbox of robotic manipulators is typically a compact harmonic drive, the transformation between rotor axes and joint axes can be neglected.

\begin{assumption}[\bf{Coincident Axes}] \label{assump:coincident_axes}
For $k \geq 3$, the screw axis of Joint $k$ is the same as the screw axis of Motor $k$'s rotor.
\end{assumption}

Assumption \ref{assump:coincident_axes} assumes that $\boldsymbol{M}_{R_k, J_k} = \boldsymbol{I}_4$ and the frame $\{R_k\}$ is equivalent to the frame $\{J_k\}$, for $k \geq 3$. Nevertheless, one can readily define $\boldsymbol{M}_{R_k, J_k}$ differently according to the modeling of harmonic drives.

The inertia of rotors is modeled as follows. Denote the gear ratio at Joint $k$ as $Z_k > 0$, for $k \geq 3$. The motion subspace matrix of Rotor $k$ in the frame $\{R_k\}$ is given by
\begin{equation}
\boldsymbol{A}_{R_k} = Z_k [\mathrm{Ad}_{\boldsymbol{M}_{R_k,J_k}}] \boldsymbol{S}_{J_{k}} = Z_k \boldsymbol{S}_{J_k}.
\end{equation}
According to the definition of apparent derivatives in \eqref{eq:apparent_deri},
\begin{equation}
\mathring{\boldsymbol{A}}_{R_k} = Z_k [\mathrm{Ad}_{\boldsymbol{M}_{R_k,J_k}}] \mathring{\boldsymbol{S}}_{R_{k}} = Z_k \mathring{\boldsymbol{S}}_{J_k} = \boldsymbol{0}_6, \ \forall k \geq 3.
\end{equation}
The spatial inertia matrix $\boldsymbol{G}_{R_k} \in \mathbb{R}^{6\times 6}$ of Motor $k$'s rotor in the frame $\{R_k\}$ is given by
\begin{equation}
\boldsymbol{G}_{R_k} = \matt{\boldsymbol{I}_{k,\mathrm{R}} & \boldsymbol{0}_{3\times 3} \\ \boldsymbol{0}_{3\times 3} & m_{k,\mathrm{R}} \boldsymbol{I}_{3} },
\end{equation}
where $\boldsymbol{I}_{k,\mathrm{R}} \in \mathbb{R}^{3\times 3}$ is the rotor's inertia tensor in matrix form; $m_{k,\mathrm{R}}$ is the mass of Motor $k$'s rotor.
The transformation from the frame $\{L_{k-1}\}$ to the frame $\{R_k\}$ is
\begin{equation}
\begin{split}
\boldsymbol{M}_{R_k,L_{k-1}} &= \boldsymbol{M}_{R_k,J_k} \boldsymbol{M}_{J_k,L_{k}} \boldsymbol{M}_{L_k,L_{k-1}} \\
&= \boldsymbol{M}_{R_k,L_{k}} \boldsymbol{M}_{L_k,L_{k-1}}, \ \forall k \geq 3.
\end{split}
\end{equation}

Finally, for $k \geq 3$,
\begin{subequations}
\begin{align}
\boldsymbol{T}_{L_k, L_{k-1}} &= e^{[-\boldsymbol{A}_{L_k}]\theta_{k}} \cdot \boldsymbol{M}_{L_k,L_{k-1}}, \\
\boldsymbol{T}_{R_k, L_{k-1}} &= e^{[-\boldsymbol{A}_{R_k}]\theta_{k}} \cdot \boldsymbol{M}_{R_k,L_{k-1}},
\end{align}
\end{subequations}
where $e^{[\cdot]\theta_{k}}$ is given by \eqref{eq:screw_axis_trans}.

The homogeneous transformations from the inertia frame $\{L_k\}$ to the global inertia frame $\{L_0\}$ (equivalent to $\{J_0\}$) are given by
\begin{equation}
\boldsymbol{T}_{L_1} \triangleq \boldsymbol{T}_{L_0, L_1}, \boldsymbol{T}_{L_k} \triangleq \boldsymbol{T}_{L_0, L_k} = \boldsymbol{T}_{L_{k-1}} (\boldsymbol{T}_{L_k, L_{k-1}})^{-1}.
\end{equation}

The homogeneous transformations from the inertia frame $\{J_k\}$ to the global inertia frame $\{L_0\}$ are given by
\begin{equation}
\boldsymbol{T}_{J_k} \triangleq \boldsymbol{T}_{L_0, J_k} = \boldsymbol{T}_{L_k} (\boldsymbol{M}_{J_k, L_k})^{-1}.
\end{equation}

Given the homogeneous transformation $\boldsymbol{T}_{J_{n+2}, J_{\mathrm{e}}}$ from the end effector frame $\{J_{\mathrm{e}}\}$ to the frame $\{L_{n+2}\}$, the transformation from the frame $\{J_{\mathrm{e}}\}$ to the global inertia frame $\{L0\}$ is given by
\begin{equation}
\boldsymbol{T}_{J_{\mathrm{e}}} \triangleq \boldsymbol{T}_{L_0, J_{\mathrm{e}}} = \boldsymbol{T}_{J_{n+2}} \cdot \boldsymbol{T}_{J_{n+2}, J_{\mathrm{e}}}.
\end{equation}
For every Joint $k$ and the end effector, its position coordinates in the global inertia frame are given by $\boldsymbol{T}_{J_k}[1:3, 4]$ and $\boldsymbol{T}_{J_{\mathrm{e}}}[1:3, 4]$, respectively.

\section{Motor Parameterization} \label{sec:motor_para}
\begin{figure}
\centering
\includegraphics[width=0.30\textwidth]{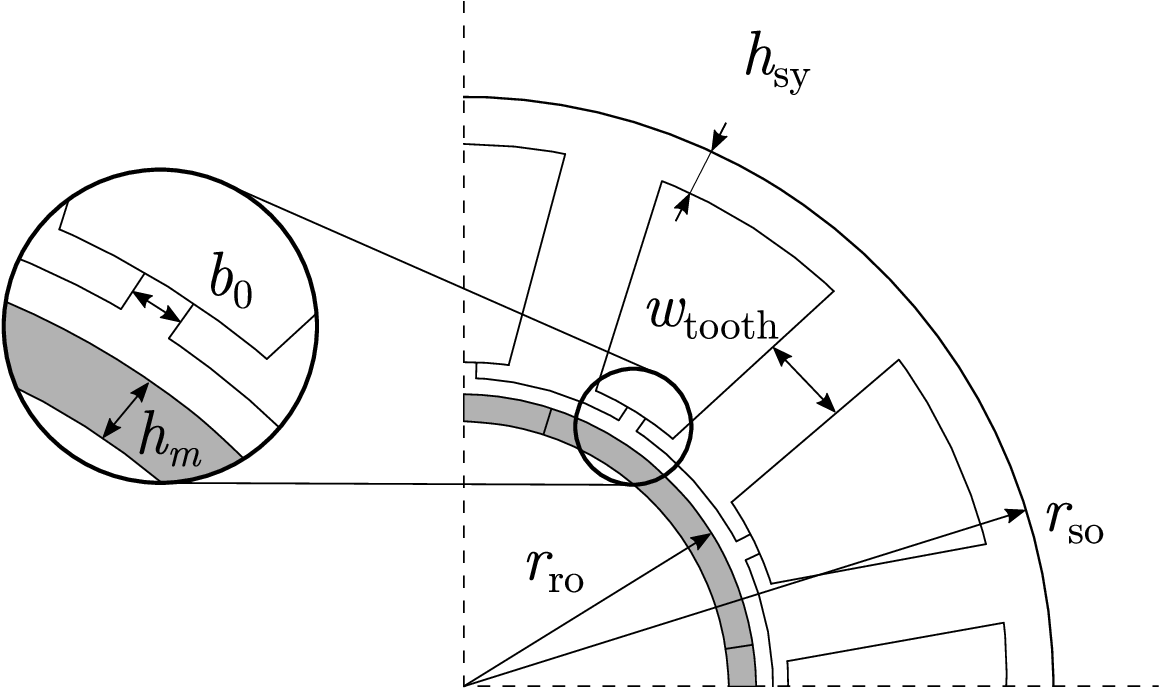}
\caption{The cross-section of an SPMSM design. The core axial length $l$ is not illustrated in this figure.}
 \label{fig:0_Motor}
\end{figure}

Surface-mounted permanent magnet synchronous motors (SPMSM) are typically used as the actuators (servomotors) of robotic manipulators.
This section introduces the modeling of a class of SPMSMs admitting the design parameterization summarized in Table \ref{table:motor_design_var}.
Fig. \ref{fig:0_Motor} illustrates the physical meaning of the design variables, where the axial length $l$ is not shown. Denote an arbitrary motor's design variables as
\begin{equation*}
\boldsymbol{\beta} \triangleq \text{col}\{l, r_{\mathrm{ro}}, r_{\mathrm{so}}, h_{\mathrm{m}}, h_{\mathrm{sy}}, w_{\mathrm{tooth}}, b_{0}\} \in \mathbb{R}^{7}.
\end{equation*}

This section also introduces how motor parameterization could affect the dynamic modeling and motion planning of the mobile manipulator.

\begin{table}
\centering
\begin{threeparttable}
\caption{SPMSM Design Parameters} \label{table:motor_design_var}
\begin{tabular}{c c | c c}
\hline
Parameter & Description & Parameter & Description \\
\hline
$l$ & Axial length of core & $h_{\mathrm{sy}}$ & Stator yoke \\
$r_{\mathrm{ro}}$ & Outer radius of rotor & $w_{\mathrm{tooth}}$ & Width of tooth \\
$r_{\mathrm{so}}$ & Outer radius of stator & $b_{0}$ & Slot opening \\
$h_{\mathrm{m}}$ & Height of magnet \\
\hline
\end{tabular}
\begin{tablenotes}
\small
\item Units of all design parameters are mm
\end{tablenotes}
\end{threeparttable}
\centering
\end{table}

\subsection{Magnetic Equivalent Circuit Modeling} \label{subsec:mec_modeling}
This subsection presents the magnetic equivalent circuit (MEC) modeling \cite{Higuchi2017} for an SPMSM as depicted in Fig.~\ref{fig:0_Motor}. To avoid confusion on whether a parameter is applied universally or just on one particular motor, this subsection introduces an index $j$ to indicate a particular motor. $\boldsymbol{\beta}_j \in \mathbb{R}^7$ denotes the design parameter associated with motor $j$.
$\text{gcd}(a,b)$ denotes the greatest common divisor of two positive integers $a$ and $b$.
The constant parameters for an SPMSM are:
\begin{itemize}
\item Number of slots $Q = 12$ 
\item Number of slots per phase $q_1 = Q/3$
\item Number of pole pairs $p = 4$
\item Number of slots per pole per phase $q_{\mathrm{pm}}=\frac{q_1}{\text{gcd}(q_1, 2p)}$
\item Number of winding turns per tooth $n_{\mathrm{s}}=50$
\item Number of coils connected in parallel $C_{\mathrm{p}} = 1$
\item Gear ratio at the $j$-th joint $Z_j = 50 \ \forall j$
\item Height of tooth tip $h_{\mathrm{tip}}=2$ mm
\item Width of air gap $\delta=0.5$ mm
\item Magnet width in electric angle $\alpha_{\mathrm{m}} = \pi$
\item Remanent flux density of the magnet $B_{\mathrm{r}} = 1.38$ T
\item Maximum limitation for flux density $B_{\mathrm{max}} = 1.5$ T
\item Mass density of iron $\rho_{\mathrm{iron}} = 7.8\cdot 10^{-6}$ kg/mm$^3$
\item Mass density of copper $\rho_{\mathrm{cu}} = 8.93\cdot 10^{-6}$ kg/mm$^3$
\item Electric resistivity of copper winding $\rho_{\mathrm{e}} = 1.8\cdot 10^{-5} \ \Omega \cdot$mm
\item Permeability of air $\mu_0 = 4\pi\cdot 10^{-7}$ N/A$^2$
\item Relative recoil permeability of the magnet $\mu_{\mathrm{r}} = 1.05$
\item Filling factor $f_{\mathrm{f}} = 0.55$ 
\end{itemize}

\subsubsection{Geometric Parameters}
According to Fig. \ref{fig:0_Motor},  the expression for the slot height is
\begin{equation} \label{eq:ap:hss}
h_{\mathrm{ss},j}=r_{\mathrm{so},j}-h_{\mathrm{sy},j}-r_{\mathrm{ro},j}-\delta-h_{\mathrm{tip}}.
\end{equation}
For a rectangular tooth cross-section, one can compute the slot width as $b_{\mathrm{ss},j} = A_{\mathrm{slot},j} / h_{\mathrm{ss},j}$, where $A_{\mathrm{slot},j}$ is the slot area, i.e.
\begin{equation} \label{eq:ap:area_slot}
A_{\mathrm{slot},j} = \textstyle\frac{\pi ((r_{\mathrm{so},j}-h_{\mathrm{sy},j})^2-(r_{\mathrm{ro},j}+\delta+h_{\mathrm{tip},j})^2)}{Q} - w_{\mathrm{tooth},j} h_{\mathrm{ss},j}.
\end{equation}
The cross-section area of the stator core is given by
\begin{equation} \label{eq:ap:area_stator_core}
A_{\mathrm{so},j} = \pi r_{\mathrm{so},j}^2 - \pi (r_{\mathrm{ro},j}+\delta)^2 - Q (A_{\mathrm{slot},j} + b_{0} h_{\mathrm{tip}}).
\end{equation}
Thus the volume of the stator core is given by
\begin{equation} \label{eq:ap:volume_stator_core}
V_j = A_{\mathrm{so},j}l_j.
\end{equation}
The copper area is given by $A_{\mathrm{cu},j} = A_{\mathrm{slot},j} f_f$. 
For concentrated windings and assuming one winding is a complete turn around a tooth, the area of a single coil is given by $A_{\mathrm{coil},j} = A_{\mathrm{cu},j} / (2n_{\mathrm{s}})$.
The minimal wire diameter is \begin{equation} \label{eq:ap:d_wire}
D_{\mathrm{wire},j} = \textstyle\sqrt{4A_{\mathrm{coil},j}/\pi}.
\end{equation}
The arc span per slot can be determined by $\tau_{\mathrm{s},j} = 2\pi(r_{\mathrm{ro},j} + \delta)/Q$. The average length of the coil end-winding $l_{\mathrm{end,av},j}$ and the total coil length $l_{\mathrm{coil},j}$ are given by
\begin{equation*}
l_{\mathrm{end,av},j} = (w_{\mathrm{tooth},j} (2-\pi/2) + \pi \tau_{\mathrm{s},j} / 2)/2, \ l_{\mathrm{coil},j} = 2l_j + 2 l_{\mathrm{end,av},j}.
\end{equation*}
Then the weight of the stator and rotor are given by:
\begin{equation} \label{eq:ap:mass_motor}
\begin{split}
m_{\mathrm{rotor},j} = \ &\rho_{\mathrm{iron}}\pi r_{\mathrm{ro},j}^2 l_j, \\
m_{\mathrm{stator},j} = \ &\rho_{\mathrm{iron}}\pi r_{\mathrm{so},j}^2 l_j - \rho_{\mathrm{iron}}\pi (r_{\mathrm{ro},j}+\delta)^2 l_j \\
&- \rho_{\mathrm{iron}}A_{\mathrm{slot},j} l_j Q + \rho_{\mathrm{cu}}A_{\mathrm{coil},j} l_{\mathrm{coil},j} n_{\mathrm{s}} Q.
\end{split}
\end{equation}
Without loss of generality, define the x-axis as the central axis of each rotor or stator, i.e. x-axis coincides with the axial length of core $l_j$; consequently, define the y-axis and z-axis by following the right-hand rule and the two axes indicate the central radius.
All three axes originate at the centroid of the rotor or stator, i.e. the center of the axial length $l_j$.
Since each rotor is a solid cylinder, the moment of inertia about three principal axes of each rotor is given by:
\begin{equation} \label{eq:inertia_rotor}
\begin{split}
I_{\mathrm{xx,r},j} &= \textstyle\frac{1}{2}\rho_{\mathrm{iron}} \pi r_{\mathrm{ro},j}^4 l_j = \textstyle\frac{1}{2}m_{\mathrm{rotor},j} r_{\mathrm{ro},j}^2, \\
I_{\mathrm{yy,r},j} &= I_{\mathrm{zz,r},j} = \textstyle\frac{1}{12}\rho_{\mathrm{iron}} \pi r_{\mathrm{ro},j}^2 l_j(3r_{ro,j}^2+l_j^2).
\end{split}
\end{equation}
To simplify the inertia calculation for stators, each stator is simplified as a hollow cylinder with outer radius $r_{\mathrm{so},j}$ and inner radius $r_{\mathrm{ro},j}+\delta$.
Then the moment of inertia about three principal axes of each stator is given by:
\begin{equation} \label{eq:inertia_stator}
\begin{split}
I_{\mathrm{xx,s},j} &= \textstyle\frac{1}{2}m_{\mathrm{stator},j} (r_{\mathrm{so},j}^2 + (r_{\mathrm{ro},j}+\delta)^2), \\
I_{\mathrm{yy,s},j} &= I_{\mathrm{zz,s},j} = \textstyle\frac{1}{12}m_{\mathrm{stator},j}(3(r_{\mathrm{so},j}^2 + (r_{\mathrm{ro},j}+\delta)^2) + l_j^2).
\end{split}
\end{equation}
For a motor installed on a joint with a known screw axis, one can use $I_{\mathrm{xx,r},j},I_{\mathrm{xx,s},j}$ defined in \eqref{eq:inertia_rotor} and \eqref{eq:inertia_stator} as the inertia around the screw axis, and use the rest as the inertia around the other two axes.

\subsubsection{Resistance}
The resistance per tooth is given by $R_{1,j} = (n_{\mathrm{s}}^2 \rho_{\mathrm{e}} l_{\mathrm{coil},j}) / (A_{\mathrm{slot},j} f_f)$.
Using this, the phase resistance can be calculated as
\begin{equation} \label{eq:ap:Rj}
R_{j} = q_1 R_{1,j} / C_{\mathrm{p}}^2.
\end{equation}

\subsubsection{Permeance}
The permeance of the magnetic path across the air gap and the slot opening, denoted by $p_{\mathrm{g},j}$ and $p_{\mathrm{so},j}$, are given by:
\begin{equation*}
p_{\mathrm{g},j} = \textstyle\frac{2\pi r_{\mathrm{ro},j}\mu_0l/Q}{\delta+h_{\mathrm{m},j}/\mu_r}, \quad p_{\mathrm{so},j} = \textstyle\frac{\mu_0 h_{\mathrm{tip}} l_j}{b_{0,j}}.
\end{equation*}
The permeance of the magnetic path that curves from tip to tip is given by $p_{\mathrm{tt},j} = \textstyle\frac{\mu_0(\delta+h_{\mathrm{m},j})l_j}{\pi(\delta+h_{\mathrm{m},j})/2 + b_{0,j}}$.

\subsubsection{Inductance}
For an SPMSM, its d-axis and q-axis inductance are equivalent to each other and given by
\begin{equation} \label{eq:ap:inductance}
L_{\mathrm{d},j} = L_{\mathrm{q},j} = q_1n_{\mathrm{s}}^2 L_{1,j} / C_{\mathrm{p}}^2,
\end{equation}
where $L_{1,j}$ is the inductance per turn and per tooth, given by $L_{1,j} = p_{\mathrm{g},j} + 3p_{\mathrm{so},j} + 3p_{\mathrm{tt},j}$.

\subsubsection{Flux}
To proceed with the calculation, it is necessary to determine Carter’s coefficient denoted by $k_{\mathrm{C},j}$, given by
\begin{equation} \label{eq:ap:carter}
k_{\mathrm{C},j} = \textstyle\frac{t_{\mathrm{pitch},j}}{t_{\mathrm{pitch},j} - \gamma_j \delta}, \  \gamma_{j} = \textstyle\frac{(b_{0,j} / \delta)^2}{5 + b_{0,j} / \delta }, \ t_{\mathrm{pitch},j} = \textstyle\frac{2\pi r_{\mathrm{ro},j}}{Q}.
\end{equation}
Then, the magnetic flux density across the gap is given by $B_{\mathrm{g},j} = B_{\mathrm{r}} \textstyle\frac{h_{\mathrm{m},j}/\mu_r}{h_{\mathrm{m},j}/\mu_r + \delta k_{\mathrm{C},j}}$.
The flux density corresponding to the first harmonics can be calculated as $B_{\mathrm{g,1},j} = 4B_{\mathrm{g},j}/\pi$. Consequently, the flux per tooth per single turn is given by
\begin{equation} \label{eq:ap:phi_1}
\Phi_{1,j} = B_{\mathrm{g,1},j} l_j 2\pi r_{\mathrm{ro},j} / Q.
\end{equation}
In the absence of skewness, the permanent flux linkage is given by
\begin{equation} \label{eq:ap:phi_m}
\Phi_{\mathrm{pm},j} = k_{\mathrm{w}} n_{\mathrm{s}} \Phi_{1,j} q_1/C_{\mathrm{p}},
\end{equation}
where $k_{\mathrm{w}} = k_{\mathrm{p}} k_{\mathrm{d}}$ denotes the winding factor, and
\begin{equation} \label{eq:ap:k_p_k_d}
k_{\mathrm{p}} = \mathrm{sin}(\pi p/Q), \quad k_{\mathrm{d}} = \textstyle\frac{\mathrm{sin}(\pi/6)}{q_{\mathrm{pm}}\mathrm{sin}(\pi/(6q_{\mathrm{pm}}))}.
\end{equation}

\subsection{Torque Capacity Modeling} \label{appendix:torque_eff_model}
This subsection introduces the modeling of SPMSM torque capacity, which is crucial to motion planning and control of the mobile manipulator.  Particularly, the maximum and minimum torques are modeled as an analytical function of motor speed and design parameters.

The dynamics of a permanent magnet synchronous motor (PMSM) are governed by ordinary differential equations (ODEs) as follows:
\begin{equation} \label{eq:motor_dynamics_PMSM}
\begin{split}
\dot{i}_{\mathrm{d},j} &= (-R_j i_{\mathrm{d},j} + p\omega_j L_{\mathrm{q},j} i_{\mathrm{q},j} + u_{\mathrm{d},j}) / L_{\mathrm{d},j}, \\
\dot{i}_{\mathrm{q},j} &= (-R_j i_{\mathrm{q},j} - (L_{\mathrm{d},j} i_{\mathrm{d},j} + \Phi_{\mathrm{pm},j})p \omega_j + u_{\mathrm{q},j}) / L_{\mathrm{q},j}, \\
\tau_{j} &= 1.5p( \Phi_{\mathrm{pm},j}i_{\mathrm{q},j} + (L_{\mathrm{d},j}-L_{\mathrm{q},j}) i_{\mathrm{q},j}i_{\mathrm{d},j}), \\ 
\end{split}
\end{equation}
where $i_{\mathrm{d},j},i_{\mathrm{q},j}$ are the current in d- and q-axis; $\omega_j$ is the rotor speed of the motor (motor speed in short); $u_{\mathrm{d},j},u_{\mathrm{q},j}$ are the voltage in d- and q-axis; $\tau_{j}$ denotes the electric torque produced by the motor.

\begin{remark}
For an SPMSM, $L_{\mathrm{d},j} = L_{\mathrm{q},j}$ and thus the electric torque $\tau_{j}$ in~\eqref{eq:motor_dynamics_PMSM} can be simplified as $\tau_{j} = 1.5p \Phi_{\mathrm{pm},j} i_{\mathrm{q},j}$.
\end{remark}

For illustration purposes, this section presents the motor operation when $\tau_{j} >0$ and $\omega_j >0$, i.e. the first quadrant. The torque capacity for the other three quadrants is equivalent to the one in the first quadrant.
A motor control strategy must be determined first to model the motor torque capacity. The following assumption defines a common motor control strategy.
\begin{assumption} \label{assume:motor_ctrl_strategy}
For an arbitrary SPMSM, the current signals $i_{\mathrm{d}}$ and $i_{\mathrm{q}}$ and the DC bus voltage are measured. The motor controller first follows the maximum torque per ampere (MPTA) strategy before hitting voltage constraints. After the motor hits either the current or voltage constraint, it follows the maximum torque per voltage (MPTV) strategy.
\end{assumption}

For an arbitrary SPMSM, given the max DC bus voltage $V_{\mathrm{max},j}$ and the max current $I_{\mathrm{max},j}$, one can estimate the feasible torque region in the speed-torque plane. The max voltage dropping to overcome back-EMF (counter-electromotive force) can be calculated as
\begin{equation}
V_{\mathrm{dq,max},j} = V_{\mathrm{max},j} / \sqrt{3} - R_j I_{\mathrm{max},j} \geq 0.
\end{equation}
Without flux weakening, i.e. $i_{\mathrm{d},j} = 0$, the corner electric speed before hitting the voltage constraint while keeping the max torque ($i_{\mathrm{q},j} = I_{\mathrm{max},j}$) is given by
\begin{equation} \label{eq:wce_define}
\omega_{\mathrm{ce},j} = \frac{V_{\mathrm{dq,max},j}}{\sqrt{(L_{\mathrm{q},j}I_{\mathrm{max},j})^2+\Phi_{\mathrm{pm},j}^2}}.
\end{equation}
Given $i_{\mathrm{d},j}$, $i_{\mathrm{q},j}$ and $\omega_j$, the back-EMF is given by
\begin{equation} \label{eq:back_emf}
V_{\mathrm{bemf},j} = p \omega_j \sqrt{(L_{\mathrm{q},j} i_{\mathrm{q},j})^2 + (\Phi_{\mathrm{pm},j}+L_{\mathrm{d},j}i_{\mathrm{d},j})^2}.
\end{equation}

\begin{remark}
Determining the feasibility of an operating point $(\omega_j, \tau_j)$ is equivalent to find a feasible pair $(i_{\mathrm{d},j}, i_{\mathrm{q},j})$ such that the following voltage and current constraints hold:
\begin{equation}
\begin{split}
p\omega_j \sqrt{(L_{\mathrm{q},j} i_{\mathrm{q},j})^2 + (\Phi_{\mathrm{pm},j}+L_{\mathrm{d},j}i_{\mathrm{d},j})^2} &\leq V_{\mathrm{dq,max},j}, \\
i_{\mathrm{d},j}^2 + i_{\mathrm{q},j}^2 &\leq I_{\mathrm{max},j}^2, \\
\end{split}
\end{equation}
where $i_{\mathrm{q},j} = \tau_j / (1.5p \Phi_{\mathrm{pm},j}) \leq I_{\mathrm{max},j}$.
\end{remark}

Denote $\omega_{\mathrm{e},j} \triangleq p \omega_j$ as the electric speed.
If $\omega_{\mathrm{e},j} \leq \omega_{\mathrm{ce},j}$ and $\tau_j \leq 1.5p \Phi_{\mathrm{pm},j} I_{\mathrm{max},j}$, then the voltage constraint is always satisfied and the feasible solution with max torque per ampere is $i_{\mathrm{d},j} = 0$ and $i_{\mathrm{q},j} = \tau_j / (1.5p \Phi_{\mathrm{pm},j})$; if $\omega_{\mathrm{e},j} \leq \omega_{\mathrm{ce},j}$ and $\tau_j > 1.5p \Phi_{\mathrm{pm},j} I_{\mathrm{max},j}$, the operating point is infeasible. Denote the corner speed when the maximum constant torque $1.5p \Phi_{\mathrm{pm},j} I_{\mathrm{max},j}$ is about to decrease as
\begin{equation} \label{eq:w_corner_define}
\omega_{\mathrm{r},j} \triangleq \omega_{\mathrm{ce},j}/p,
\end{equation}
where $\omega_{\mathrm{ce},j}$ is given by \eqref{eq:wce_define}.

With $\omega_{\mathrm{e},j} > \omega_{\mathrm{ce},j}$, one needs to infer the max allowable q-axis current $i_{\mathrm{q,lim},j}$ which is restricted by either the max voltage or max current constraint. For the operating point $(\omega_j, \tau_j)$, it requires $i_{\mathrm{q},j} = \tau_j / (1.5p \Phi_{\mathrm{pm},j})$. We first consider the special case when both current and voltage constraints are active, the solution of which is obtained by solving two unknown $i_{\mathrm{d},j},i_{\mathrm{q},j}$ from two equations
\begin{equation} \label{eq:voltage_and_current_constraint}
\begin{split}
p \omega_j \sqrt{(L_{\mathrm{q},j} i_{\mathrm{q},j})^2 + (\Phi_{\mathrm{pm},j}+L_{\mathrm{d},j}i_{\mathrm{d},j})^2} &= V_{\mathrm{dq,max},j}, \\
i_{\mathrm{d},j}^2 + i_{\mathrm{q},j}^2 &= I_{\mathrm{max},j}^2. \\
\end{split}
\end{equation}
Its solution is given by
\begin{equation} \label{eq:current_limit_geq}
\begin{split}
i_{\mathrm{d,lim},j} &= \frac{ (V_{\mathrm{dq,max},j}/\omega_{\mathrm{e},j})^2 - (L_{\mathrm{d},j} I_{\mathrm{max},j})^2 - \Phi_{\mathrm{pm},j}^2 }{2\Phi_{\mathrm{pm},j}L_{\mathrm{d},j}}, \\
i_{\mathrm{q,lim},j} &= \sqrt{I_{\mathrm{max},j}^2 - i_{\mathrm{d,lim},j}^2}.\\
\end{split}
\end{equation}
When $\omega_j = \omega_{\mathrm{r},j}$, $i_{\mathrm{d,lim},j}=0$; when $\omega_j > \omega_{\mathrm{r},j}$, $i_{\mathrm{d,lim},j}$ is always negative to weaken the permanent flux.
Since the current is bounded by $i_{\mathrm{d},j}^2 + i_{\mathrm{q},j}^2 \leq I_{\mathrm{max},j}^2$, the magnitude of $i_{\mathrm{d},j}$ cannot exceed $I_{\mathrm{max},j}$.
According to \eqref{eq:current_limit_geq} with $\omega_{\mathrm{e},j} \coloneqq p \omega_j$, $i_{\mathrm{d,lim},j}(\omega_j)$ is a monotonically decreasing function to the motor speed $\omega_j$.
Therefore, one can analytically find the crossover point $\omega_{\mathrm{max},j}$ for $i_{\mathrm{d,lim},j}(\omega_{\mathrm{max},j}) = -I_{\mathrm{max},j}$, which yields
\begin{equation} \label{eq:omega_max}
\omega_{\mathrm{max},j} \triangleq \frac{V_{\mathrm{dq,max},j}}{p |\Phi_{\mathrm{pm},j}-L_{\mathrm{d},j}I_{\mathrm{max},j}|}.
\end{equation}
$\omega_{\mathrm{max},j}$ is the maximum motor speed given both the current and voltage constraints only when $\Phi_{\mathrm{pm},j}-L_{\mathrm{d},j}I_{\mathrm{max},j} >0$. From \eqref{eq:omega_max}, it is obvious that when $\Phi_{\mathrm{pm},j}-L_{\mathrm{d},j}I_{\mathrm{max},j} = 0$, $\omega_{\mathrm{max},j} = \infty$. In other words, the voltage and current constraints will not limit the motor speed.
When $\Phi_{\mathrm{pm},j}-L_{\mathrm{d},j}I_{\mathrm{max},j} <0$, $\omega_{\mathrm{max},j} = \infty$. The following two figures illustrate the reason as \eqref{eq:omega_max} cannot.

\begin{figure*}
\subfloat[$i_{\mathrm{d,lim},j}$ for case $\Phi_{\mathrm{pm},j}/L_{\mathrm{d},j} > I_{\mathrm{max},j}$.]
{\label{fig:id_lim:positive} \includegraphics[width=0.32\linewidth]{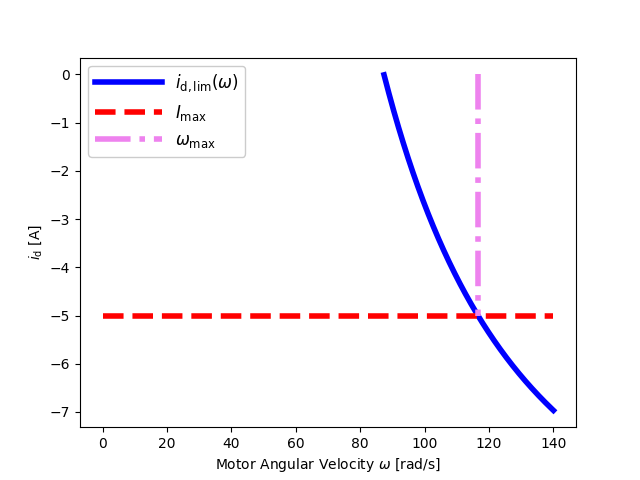}}
\hfill
\subfloat[$i_{\mathrm{d,lim},j}$ for case $\Phi_{\mathrm{pm},j}/L_{\mathrm{d},j} = I_{\mathrm{max},j}$.]
{\label{fig:id_lim:0} \includegraphics[width=0.32\linewidth]{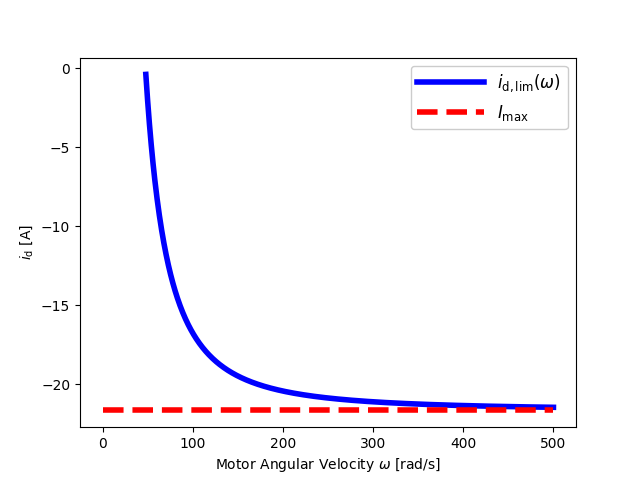}}
\hfill
\subfloat[$i_{\mathrm{d,lim},j}$ for case $\Phi_{\mathrm{pm},j}/L_{\mathrm{d},j} - I_{\mathrm{max},j} < 0$.]
{\label{fig:id_lim:negative} \includegraphics[width=0.32\linewidth]{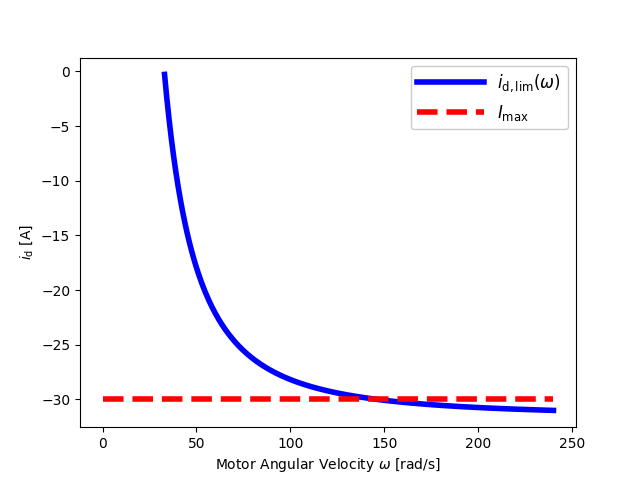}}
\caption{$i_{\mathrm{d,lim},j}$ as a function of $\omega_j$ in all cases. The blue lines represent the function $i_{\mathrm{d,lim},j}(\omega_j)$; the red dashed lines represent $I_{\mathrm{max},j}$; the pink dash-dot lines represent $\omega_{\mathrm{max},j}$ from \eqref{eq:omega_max}. The intersection in (a) indicates the maximum motor speed constrained by the maximum current $I_{\mathrm{max},j}$.} \label{fig:id_lim}
\end{figure*}

\begin{figure*}
\subfloat[Visualization of current and voltage constraints for case $\Phi_{\mathrm{pm},j}/L_{\mathrm{d},j} > I_{\mathrm{max},j}$.]
{\label{fig:current_voltage:positive} \includegraphics[width=0.31\linewidth]{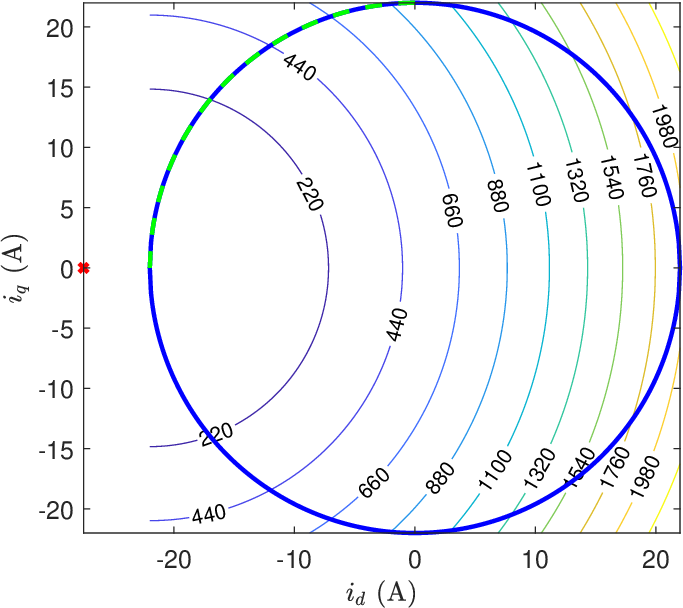}}
\hfill
\subfloat[Visualization of current and voltage constraints for case $\Phi_{\mathrm{pm},j}/L_{\mathrm{d},j} = I_{\mathrm{max},j}$.]
{\label{fig:current_voltage:0} \includegraphics[width=0.29\linewidth]{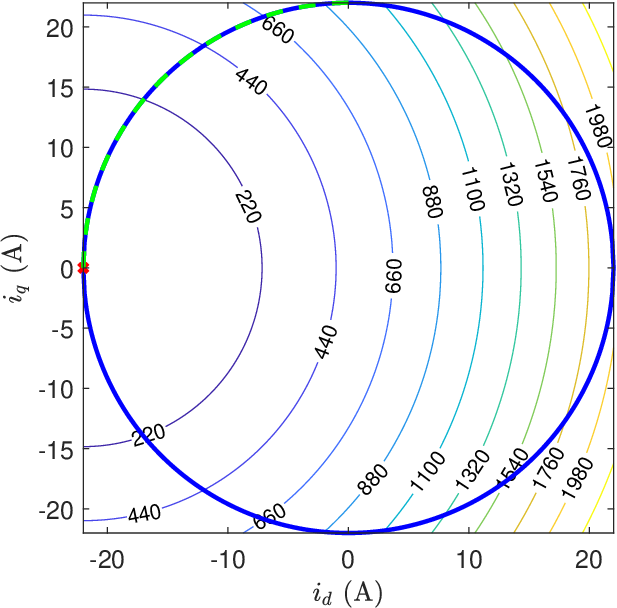}}
\hfill
\subfloat[Visualization of current and voltage constraints for case $\Phi_{\mathrm{pm},j}/L_{\mathrm{d},j} - I_{\mathrm{max},j} < 0$.]
{\label{fig:current_voltage:negative} \includegraphics[width=0.29\linewidth]{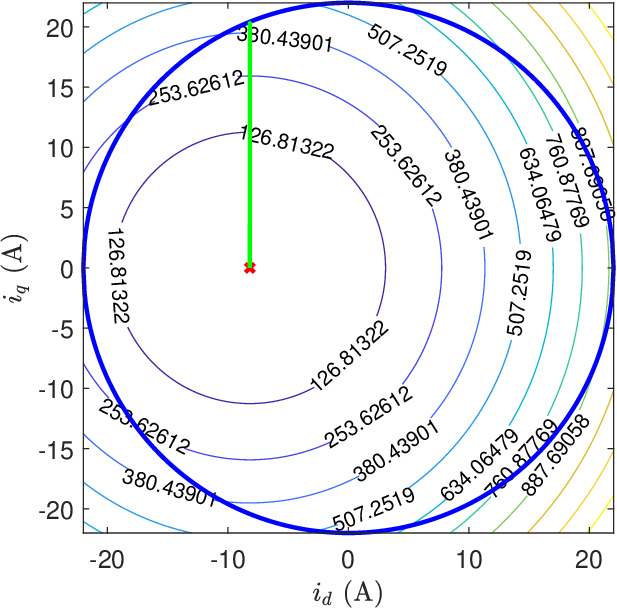}}
\caption{Visualization of current and voltage constraints in all cases given the same motor speed. Blue circles represent current constraints; contours represent voltage constraints; red cross marks represent centers of the voltage constraint contours; green (dashed) lines represent the max torque loci while increasing the motor speed.} \label{fig:current_voltage}
\end{figure*}

First, a concrete example of $i_{\mathrm{d,lim},j}$ with three different signs of $\Phi_{\mathrm{pm},j}-L_{\mathrm{d},j}I_{\mathrm{max},j}$ is shown in Fig. \ref{fig:id_lim}, where only $I_{\mathrm{max},j}$ changes as 5 A, 21.67 A, and 30 A for Fig.~\ref{fig:id_lim:positive} - Fig.~\ref{fig:id_lim:negative}, respectively. When $\Phi_{\mathrm{pm},j}-L_{\mathrm{d},j}I_{\mathrm{max},j} > 0$, there exists a finite $\omega_{\mathrm{max},j}$. When $\Phi_{\mathrm{pm},j}-L_{\mathrm{d},j}I_{\mathrm{max},j} = 0$, $i_{\mathrm{d,lim},j}(\omega_j)$ asymptotically converges to $I_{\mathrm{max},j}$ as $\omega_j$ goes to infinity.
In Fig.~\ref{fig:id_lim:negative}, $\omega_j$ exceeds the intersection (its value can be calculated by \eqref{eq:omega_max}) when $i_{\mathrm{d,lim},j}$ reaches $I_{\mathrm{max},j}= 30$ A. When $\omega_j$ keeps increasing after this intersection, by Assumption~\ref{assume:motor_ctrl_strategy}, the motor cannot operate at both the boundary of the current and voltage constraint. Thus, one cannot use \eqref{eq:voltage_and_current_constraint} to jointly determine $i_{\mathrm{d},j}$ and $i_{\mathrm{q},j}$, and its further derivation \eqref{eq:current_limit_geq} and \eqref{eq:omega_max} cannot be used as well.
Fig.~\ref{fig:current_voltage:negative} illustrates the reason in this particular case by showing the current constraint and the contours of the voltage constraint. Before explaining that, it is necessary to introduce the voltage constraint in the $i_{\mathrm{d},j}$-$i_{\mathrm{q},j}$ plane.

The max torque for a given $\omega_j \geq \omega_{\mathrm{r},j}$ always achieves at the boundary of the maximum voltage constraint. This is because if the max voltage constraint is inactive while the current constraint is hit, then from \eqref{eq:back_emf}, one can always reduce $i_{\mathrm{d},j}$ and increase $i_{\mathrm{q},j}$ to produce a larger torque while ensuring the current constraints. On the other hand, the max current constraint is not necessarily active.

\begin{remark}
The voltage constraint can be rearranged as follows
\begin{equation} \label{eq:torque_bound_voltage_constraint}
i_{\mathrm{q},j}^2 + (\frac{\Phi_{\mathrm{pm},j}}{L_{\mathrm{d},j}} + i_{\mathrm{d},j} )^2 = \frac{V_{\mathrm{dq,max},j}^2}{\omega_{\mathrm{e},j}^2 L_{\mathrm{d},j}^2},
\end{equation}
which is a circle centered at $(-\frac{\Phi_{\mathrm{pm},j}}{L_{\mathrm{d},j}},0)$ in the $i_{\mathrm{d},j}$-$i_{\mathrm{q},j}$ plane with a radius of $\frac{V_{\mathrm{dq,max},j}}{\omega_{\mathrm{e},j} L_{\mathrm{d},j}}$. On the other hand, the current constraint is a circle centered at the origin with a radius $I_{\mathrm{max},j}$.
\end{remark}

The contours of voltage constraints \eqref{eq:torque_bound_voltage_constraint} are shown in Fig. \ref{fig:current_voltage}, where the red crosses represent the center $(-\frac{\Phi_{\mathrm{pm},j}}{L_{\mathrm{d},j}},0)$; the blue circles represent the current constraints; green solid/dashed lines represent the max torque loci while increasing the motor speed. One can see that if $\Phi_{\mathrm{pm},j}/L_{\mathrm{d},j} - I_{\mathrm{max},j} \geq 0$, the maximum torque always achieves at the boundary of the current and voltage constraints because there are always intersections between the current constraint (blue circle) and the voltage constraint (contours).
When $\Phi_{\mathrm{pm},j}/L_{\mathrm{d},j} - I_{\mathrm{max},j} < 0$, the max torque does not achieve at the maximum current, but inside along the green line inside the blue circle. At some points, there is no intersection between the blue circle and the contours. 
This indicates that one cannot calculate $i_{\mathrm{d},j}$ and $i_{\mathrm{q},j}$ by \eqref{eq:voltage_and_current_constraint}, i.e. on both the voltage and current boundary, but only on the voltage boundary (every contour).

To summarize, one can determine the feasible operation region, i.e. the torque capacity, of a motor numerically using the following algorithms. Algorithm \ref{alg:tau_omg_to_idq} converts every arbitrary operating point $(\omega_j, \tau_j)$ into $i_{\mathrm{d},j}$ and $i_{\mathrm{q},j}$ when feasible, and returns a false flag when infeasible.
Note that given motor design parameters $\boldsymbol{\beta}_j$, one can calculates $R_j$, $L_{\mathrm{d},j}$, $L_{\mathrm{q},j}$, $\Phi_{\mathrm{pm},j}$ from Section~\ref{subsec:mec_modeling}.
Algorithm~\ref{alg:motor_feasible_region} presents how to generate a motor feasible operation map given an arbitrary range of motor speed $[0, \hat{\omega}_{\mathrm{max},j}]$ and torque $[0, \hat{\tau}_{\mathrm{max},j}]$.
Keeping all the parameters the same but given different $I_{\mathrm{max},j}$, Fig. \ref{fig:torque_map} presents the motor operation map and torque capacity given three different signs of $\Phi_{\mathrm{pm},j}-L_{\mathrm{d},j}I_{\mathrm{max},j}$. The red region represents the feasible operation range, whereas the white region represents the infeasible range. The boundary between the white and the red region is the maximum motor torque at each motor speed obtained numerically by Algorithm \ref{alg:motor_feasible_region}.
The dashed lines/curves are some analytical functions for torque bounds; the vertical dash-dot lines indicate some critical speeds; the details of the analytical torque bound functions are given in Section~\ref{subsec:torque_analytical_model}.

\begin{algorithm2e}
\caption{Conversion of motor speed and torque to motor currents}\label{alg:tau_omg_to_idq}
\DontPrintSemicolon
\KwIn{$\boldsymbol{\beta}_j$, $V_{\mathrm{max},j}$, $I_{\mathrm{max},j}$, number of pole pairs $p$}
\SetKwFunction{omgTau}{omgTau2currents}
\SetKwProg{Fn}{def}{:}{}
\Fn{\omgTau{$\omega_j$, $\tau_j$}}{
$V_{\mathrm{dq,max},j} \gets V_{\mathrm{max},j} / \sqrt{3} - R_j I_{\mathrm{max},j}$, flag $\gets$ True\;
$\omega_{\mathrm{r},j} \gets \frac{V_{\mathrm{dq,max},j}}{p\sqrt{(L_{\mathrm{q},j}I_{\mathrm{max},j})^2+\Phi_{\mathrm{pm},j}^2}}$, $\omega_{\mathrm{e},j} \gets p \omega_j$\;
\eIf{$\omega_j \leq \omega_{\mathrm{r},j}$}{
$i_{\mathrm{q},j} \gets \tau_j / (1.5p \Phi_{\mathrm{pm},j})$, $i_{\mathrm{d},j} \gets 0$\;
\lIf{$i_{\mathrm{q},j} > I_{\mathrm{max},j}$}
{
flag $\gets$ False, $i_{\mathrm{d},j},i_{\mathrm{q},j} \gets \text{None}$
}
}
{
$i_{\mathrm{d,lim},j} \gets \frac{ (V_{\mathrm{dq,max},j}/\omega_{\mathrm{e},j})^2 - (L_{\mathrm{d},j} I_{\mathrm{max},j})^2 - \Phi_{\mathrm{pm},j}^2 }{2\Phi_{\mathrm{pm},j}L_{\mathrm{d},j}}$\;
$i_{\mathrm{q,lim,V},j} \gets V_{\mathrm{dq,max},j}/(\omega_{\mathrm{e},j} L_{\mathrm{q},j})$, $i_{\mathrm{q,t},j} \gets \tau_j / (1.5p \Phi_{\mathrm{pm},j})$\;
\If{$i_{\mathrm{d,lim},j} \leq -I_{\mathrm{max},j}$}
{
\If{$i_{\mathrm{q,t},j} > i_{\mathrm{q,lim,V},j}$ or $\Phi_{\mathrm{pm},j}/L_{\mathrm{d},j} - I_{\mathrm{max},j} \geq 0$}
{
flag $\gets$ False, $i_{\mathrm{d},j},i_{\mathrm{q},j} \gets \text{None}$
}
\lElse
{
$i_{\mathrm{q},j} \gets i_{\mathrm{q,t},j}$, $i_{\mathrm{d},j} \gets -\Phi_{\mathrm{pm},j} / L_{\mathrm{d},j}$
}
}
\Else
{
$i_{\mathrm{d,lim},j} \gets \text{sign}(i_{\mathrm{d,lim},j}) \text{min}(|i_{\mathrm{d,lim},j}|, I_{\mathrm{max},j})$\;
\If{$i_{\mathrm{d,lim},j}L_{\mathrm{d},j} + \Phi_{\mathrm{pm},j} < 0$}
{
$i_{\mathrm{d,lim},j} \gets -\Phi_{\mathrm{pm},j} / L_{\mathrm{d},j}$\;
$i_{\mathrm{q,lim,C},j} \gets \sqrt{I_{\mathrm{max},j}^2 - i_{\mathrm{d,lim},j}^2}$\;
$i_{\mathrm{q,lim},j} \gets \text{min}(i_{\mathrm{q,lim,V},j},i_{\mathrm{q,lim,C},j})$\;
}
\lElse
{
$i_{\mathrm{q,lim},j} \gets \sqrt{I_{\mathrm{max},j}^2 - i_{\mathrm{d,lim},j}^2}$
}
\If{$i_{\mathrm{q,t},j} > i_{\mathrm{q,lim},j}$}
{
flag $\gets$ False, $i_{\mathrm{d},j},i_{\mathrm{q},j} \gets \text{None}$\;
}
\lElse
{
$i_{\mathrm{q},j} \gets i_{\mathrm{q,t},j}$, $i_{\mathrm{d},j} \gets i_{\mathrm{d,lim},j}$
}
}
}
\KwRet $i_{\mathrm{d},j}$, $i_{\mathrm{q},j}$, flag\;
}
\end{algorithm2e}

\begin{algorithm2e}[h!]
\DontPrintSemicolon
\KwIn{$\boldsymbol{\beta}_j$, $V_{\mathrm{max},j}$, $I_{\mathrm{max},j}$, number of pole pairs $p$, $N_{\omega} \in \mathbb{Z}_+$, $N_{\tau} \in \mathbb{Z}_+$, $\hat{\omega}_{\mathrm{max},j} >0$, $\hat{\tau}_{\mathrm{max},j} >0$}
$\boldsymbol{M} \gets \text{zeros}(N_{\tau}, N_{\omega})$, $\boldsymbol{v}_{\omega} \gets\text{linspace}(0, \hat{\omega}_{\mathrm{max},j}, N_{\omega})$, $\boldsymbol{v}_{\tau} \gets \text{linspace}(0, \hat{\tau}_{\mathrm{max},j}, N_{\tau})$\;
\For{$i_{\tau}$ $\emph{in range}(N_{\tau})$}{
\For{$i_{\omega}$ $\emph{in range}(N_{\omega})$}{
$\omega_j \gets \boldsymbol{v}_{\omega}[i_{\omega}]$, $\tau_j \gets \boldsymbol{v}_{\tau}[i_{\tau}]$\;
$i_{\mathrm{d},j}, i_{\mathrm{q},j}, \text{flag} \gets$ \omgTau{$\omega_j$, $\tau_j$}\;
\lIf{\emph{flag is True}}{
$\boldsymbol{M}[i_{\tau},i_{\omega}] = 1$
}
}
}
\Return{$\boldsymbol{M}$}
\caption{Generation of motor operation map and torque capacity}
\label{alg:motor_feasible_region}
\end{algorithm2e}

\subsection{Analytical Torque Bound Modeling} \label{subsec:torque_analytical_model}

Given the motor control strategy defined in Assumption \ref{assume:motor_ctrl_strategy} and the derivation in Section \ref{appendix:torque_eff_model}, the motor torque bound can be modeled analytically, which can be categorized into three cases, depending on the sign of $\Phi_{\mathrm{pm},j}/L_{\mathrm{d},j} - I_{\mathrm{max},j}$.

\begin{figure*}
\subfloat[Motor operation map for case $\Phi_{\mathrm{pm},j}/L_{\mathrm{d},j} > I_{\mathrm{max},j}$.]
{\label{fig:torque_map:positive} \includegraphics[width=0.32\linewidth]{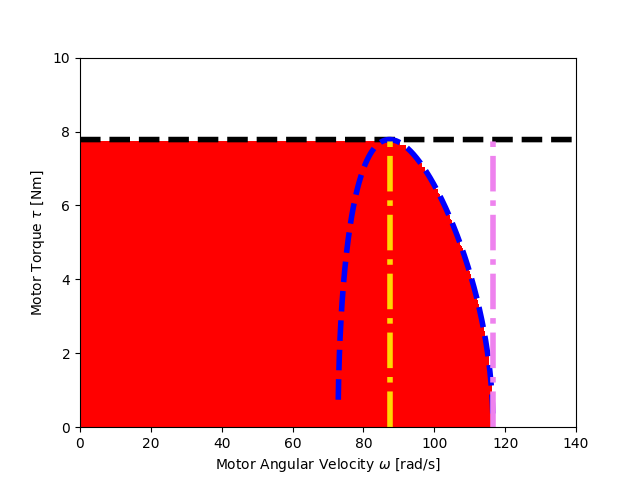}}
\hfill
\subfloat[Motor operation map for case $\Phi_{\mathrm{pm},j}/L_{\mathrm{d},j} = I_{\mathrm{max},j}$.]
{\label{fig:torque_map:0} \includegraphics[width=0.32\linewidth]{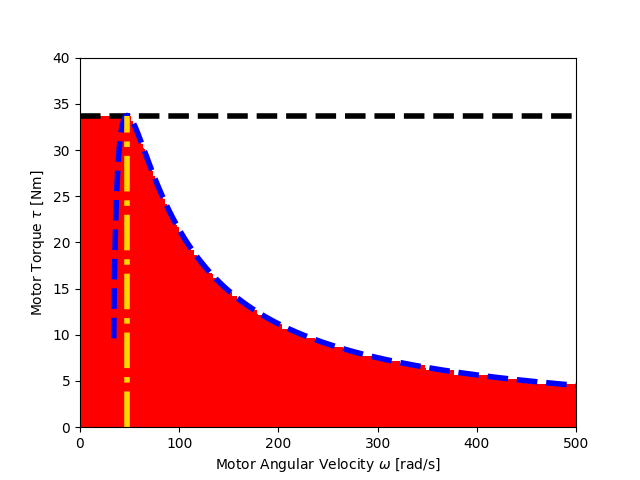}}
\hfill
\subfloat[Motor operation map for case $\Phi_{\mathrm{pm},j}/L_{\mathrm{d},j} - I_{\mathrm{max},j} < 0$.]
{\label{fig:torque_map:negative} \includegraphics[width=0.32\linewidth]{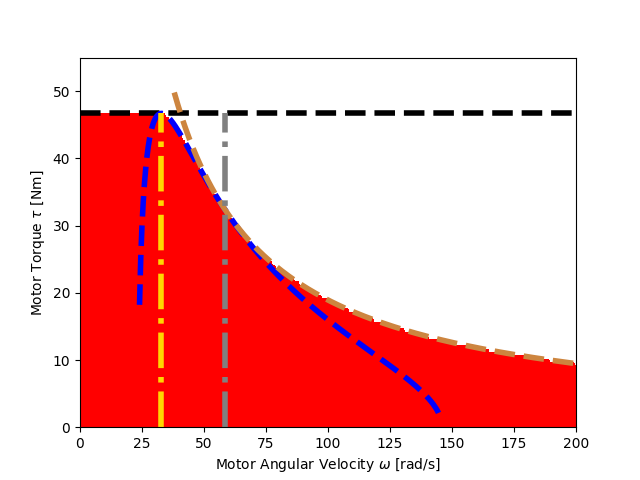}}
\caption{Motor operation map in all cases. The red region represents the feasible operation range; the white region represents the infeasible range. The meaning of the dashed and dash-dot lines/curves is given in Section \ref{subsec:torque_analytical_model}.} \label{fig:torque_map}
\end{figure*}

When $\Phi_{\mathrm{pm},j}/L_{\mathrm{d},j} - I_{\mathrm{max},j} > 0$, the maximum motor torque $\tau_{\mathrm{max},j}(\omega_j, \boldsymbol{\beta}_j)$, as a function of motor speed and motor design parameters, is given by
\begin{equation} \label{eq:motor_torque_max_positive}
\tau_{\mathrm{max},j} = \begin{cases}
1.5p \Phi_{\mathrm{pm},j}I_{\mathrm{max},j},& \text{if } |\omega_j| \leq \omega_{\mathrm{r},j}\\
1.5p \Phi_{\mathrm{pm},j}i_{\mathrm{q,lim},j},& \text{if } |\omega_j| \in [\omega_{\mathrm{r},j},\omega_{\mathrm{max},j}]
\end{cases}
\end{equation}
where $\omega_{\mathrm{r},j}$, $i_{\mathrm{q,lim},j}$, and $\omega_{\mathrm{max},j}$ are given by \eqref{eq:w_corner_define}, \eqref{eq:current_limit_geq} and \eqref{eq:omega_max}, respectively. This piecewise-defined function is visualized in Fig. \ref{fig:torque_map:positive}. The yellow dash-dot vertical line represents $\omega_{\mathrm{r},j}$, which is the motor corner speed to start decreasing the constant maximum torque. The purple dash-dot vertical line represents $\omega_{\mathrm{max},j}$, which is the maximum motor speed given current and voltage constraints.
The black dashed horizontal line represents the constant maximum torque given by the first row of \eqref{eq:motor_torque_max_positive}.
The blue dashed curve represents the maximum torque, which decreases as the motor speed increases, according to the second row of \eqref{eq:motor_torque_max_positive}. Fig. \ref{fig:torque_map:positive} verifies the analytical torque bound given the numerical result from Algorithm \ref{alg:motor_feasible_region}.
Even though the particular shape of the maximum torque in $[\omega_{\mathrm{r},j}, \omega_{\mathrm{max},j}]$ is concave in Fig. \ref{fig:torque_map:positive}, with $I_{\mathrm{max},j}$ increasing but still less than $\Phi_{\mathrm{pm},j}/L_{\mathrm{d},j}$, the shape could become convex.

When $\Phi_{\mathrm{pm},j}/L_{\mathrm{d},j} - I_{\mathrm{max},j} = 0$, the maximum motor torque is given by
\begin{equation} \label{eq:motor_torque_max_0}
\tau_{\mathrm{max},j} = \begin{cases}
1.5p \Phi_{\mathrm{pm},j}I_{\mathrm{max},j},& \text{if } |\omega_j| \leq \omega_{\mathrm{r},j}\\
1.5p \Phi_{\mathrm{pm},j}i_{\mathrm{q,lim},j},& \text{if } |\omega_j| \in [\omega_{\mathrm{r},j},\infty)
\end{cases}
\end{equation}
where $\omega_{\mathrm{r},j}$ and $i_{\mathrm{q,lim},j}$ are given by \eqref{eq:w_corner_define} and \eqref{eq:current_limit_geq}, respectively.
Note that the only difference between \eqref{eq:motor_torque_max_positive} and \eqref{eq:motor_torque_max_0} is whether there exists a maximum motor speed $\omega_{\mathrm{max},j}$ or not.
This piecewise-defined function is visualized in Fig. \ref{fig:torque_map:0}. The yellow dash-dot vertical line represents $\omega_{\mathrm{r},j}$.
According to \eqref{eq:omega_max}, there is no $\omega_{\mathrm{max},j}$ or $\omega_{\mathrm{max},j} = \infty$. 
The black dashed horizontal line represents the constant maximum torque given by the first row of \eqref{eq:motor_torque_max_0}.
The blue dashed curve represents the maximum torque based on the second row of \eqref{eq:motor_torque_max_0}. Fig. \ref{fig:torque_map:0} verifies the analytical torque bound given the numerical result from Algorithm \ref{alg:motor_feasible_region}.

When $\Phi_{\mathrm{pm},j}/L_{\mathrm{d},j} - I_{\mathrm{max},j} < 0$, one needs to find the critical motor speed $\omega_{\mathrm{s},j}$ that switches from the boundary of the current constraint to the boundary of the voltage constraint. This point is equivalent to the intersection point between the blue circle (current constraint) and the green line (voltage constraint) in Fig. \ref{fig:current_voltage:negative}.

There are two ways to find $\omega_{\mathrm{s},j}$ analytically.
Regarding the first method, when $\Phi_{\mathrm{pm},j}/L_{\mathrm{d},j} - I_{\mathrm{max},j} \geq 0$, there always holds $\Phi_{\mathrm{pm},j} - i_{\mathrm{d,lim},j} L_{\mathrm{d},j} \geq 0$ due to $-I_{\mathrm{max},j} \leq i_{\mathrm{d,lim},j} \leq 0$.
This means that $i_{\mathrm{d},j} = i_{\mathrm{d,lim},j}$ when the torque is maximum.
However, when $\Phi_{\mathrm{pm},j}/L_{\mathrm{d},j} - I_{\mathrm{max},j} < 0$, $i_{\mathrm{d,lim},j} L_{\mathrm{d},j} + \Phi_{\mathrm{pm},j}$ could vary from positive to negative, depending on the motor speed.
Note that this $i_{\mathrm{d,lim},j}$ is obtained by both the activated current and voltage constraint, according to \eqref{eq:current_limit_geq}.
When $\Phi_{\mathrm{pm},j} + i_{\mathrm{d,lim},j} L_{\mathrm{d},j} < 0$, to avoid the flux go to negative, the current limit on the d-axis is no longer given by \eqref{eq:current_limit_geq}, but given by $\Phi_{\mathrm{pm},j} + i_{\mathrm{d},j} L_{\mathrm{d},j} = 0$ to compensate the flux, thus
\begin{equation} \label{eq:id_limit_negative}
i_{\mathrm{d,lim,V},j} = -\Phi_{\mathrm{pm},j} / L_{\mathrm{d},j}.
\end{equation}
In this case, since the voltage constraint is activated, the maximum current on the q-axis, i.e. the maximum torque, is derived from the voltage constraint, which reads
\begin{equation} \label{eq:iq_limit_negative}
i_{\mathrm{q,lim,V},j} = V_{\mathrm{dq,max},j} / (\omega_{\mathrm{e},j}L_{\mathrm{q},j}).
\end{equation}
Therefore, the critical condition that determines whether the maximum torque is constrained by the voltage constraint only or by both the voltage and current constraints is $i_{\mathrm{d,lim},j} L_{\mathrm{d},j} + \Phi_{\mathrm{pm},j} = 0$ together with $i_{\mathrm{d,lim},j} = \frac{ (V_{\mathrm{dq,max},j}/\omega_{\mathrm{e},j})^2 - (L_{\mathrm{d},j} I_{\mathrm{max},j})^2 - \Phi_{\mathrm{pm},j}^2 }{2\Phi_{\mathrm{pm},j}L_{\mathrm{d},j}}$, i.e.
\begin{equation}
\frac{ (V_{\mathrm{dq,max},j}/\omega_{\mathrm{e},j})^2 - (L_{\mathrm{d},j} I_{\mathrm{max},j})^2 - \Phi_{\mathrm{pm},j}^2 }{2\Phi_{\mathrm{pm},j}L_{\mathrm{d},j}} L_{\mathrm{d},j} + \Phi_{\mathrm{pm},j} = 0,
\end{equation}
which gives the critical speed $\omega_{\mathrm{s},j}$ as
\begin{equation} \label{eq:omega_switch_define}
\omega_{\mathrm{s},j} \triangleq \frac{V_{\mathrm{dq,max},j}}{p\sqrt{(L_{\mathrm{d},j} I_{\mathrm{max},j})^2 - \Phi_{\mathrm{pm},j}^2}}.
\end{equation}
When $\omega_j \leq \omega_{\mathrm{r},j}$, the maximum torque is a constant as in the other two cases above; when $\omega_j \in [\omega_{\mathrm{r},j}, \infty)$, both the voltage and current constraints are activated, thus the maximum torque occurs when the currents are given by \eqref{eq:current_limit_geq}; when $\omega_j \in [\omega_{\mathrm{s},j}, \omega_{\mathrm{max},j}]$, only the voltage constraint is activated, thus the maximum torque occurs when the currents are given by \eqref{eq:id_limit_negative} and \eqref{eq:iq_limit_negative}.

As for the second method to determine $\omega_{\mathrm{s},j}$, inspired by Fig. \ref{fig:current_voltage:negative}, the critical speed occurs when the motor torque induced by both the current and voltage constraints is equal to the torque induced by only the voltage constraint.
The former is equal to $1.5p \Phi_{\mathrm{pm},j}i_{\mathrm{q,lim},j}$, where $i_{\mathrm{q,lim},j}$ is given by \eqref{eq:current_limit_geq}.
The latter is equal to $1.5p \Phi_{\mathrm{pm},j}i_{\mathrm{q,lim,V},j}$, where $i_{\mathrm{q,lim,V},j}$ is given by \eqref{eq:id_limit_negative}.
Thus one has
\begin{equation}
\begin{split}
&\frac{V_{\mathrm{dq,max},j}}{\omega_{\mathrm{e},j}L_{\mathrm{q},j}} = \sqrt{I_{\mathrm{max},j}^2 - i_{\mathrm{d,lim},j}^2}, \\
&i_{\mathrm{d,lim},j} = \frac{ (V_{\mathrm{dq,max},j}/\omega_{\mathrm{e},j})^2 - (L_{\mathrm{d},j} I_{\mathrm{max},j})^2 - \Phi_{\mathrm{pm},j}^2 }{2\Phi_{\mathrm{pm},j}L_{\mathrm{d},j}},
\end{split}
\end{equation}
which can be simplified as a quadratic-like equation, i.e.
\begin{equation} \label{eq:quartic_equation}
\begin{split}
&(L_{\mathrm{d},j}^2 I_{\mathrm{max},j}^2 - \Phi_{\mathrm{pm},j}^2)^2 \omega_{\mathrm{e},j}^4 - 2V_{\mathrm{dq,max},j}^2(L_{\mathrm{d},j}^2 I_{\mathrm{max},j}^2 - \\
&\Phi_{\mathrm{pm},j}^2) \omega_{\mathrm{e},j}^2 + V_{\mathrm{dq,max},j}^4 = 0.
\end{split}
\end{equation}
Since $\Phi_{\mathrm{pm},j}/L_{\mathrm{d},j} - I_{\mathrm{max},j} < 0$, one has $L_{\mathrm{d},j} I_{\mathrm{max},j} > \Phi_{\mathrm{pm},j} > 0$ and thus $L_{\mathrm{d},j}^2 I_{\mathrm{max},j}^2 - \Phi_{\mathrm{pm},j}^2 > 0$.
Then \eqref{eq:quartic_equation} can be written as
\begin{equation} \label{eq:qaratic_equation_2}
(\omega_{\mathrm{e},j}^2 - \frac{V_{\mathrm{dq,max},j}^2}{L_{\mathrm{d},j}^2 I_{\mathrm{max},j}^2 - \Phi_{\mathrm{pm},j}^2})^2 = 0.
\end{equation}
Together with $\omega_{\mathrm{e},j} = p \omega_j >0$, solving \eqref{eq:qaratic_equation_2} gives
\begin{equation*}
\omega_{\mathrm{s},j} = \frac{V_{\mathrm{dq,max},j}}{p\sqrt{(L_{\mathrm{d},j} I_{\mathrm{max},j})^2 - \Phi_{\mathrm{pm},j}^2}},
\end{equation*}
which verifies \eqref{eq:omega_switch_define} obtained by the first method.

\begin{figure*}
\subfloat[Motor torque bound for case $\Phi_{\mathrm{pm},j}/L_{\mathrm{d},j} > I_{\mathrm{max},j}$.]
{\label{fig:motor_torque_bound:positive} \includegraphics[width=0.32\linewidth]{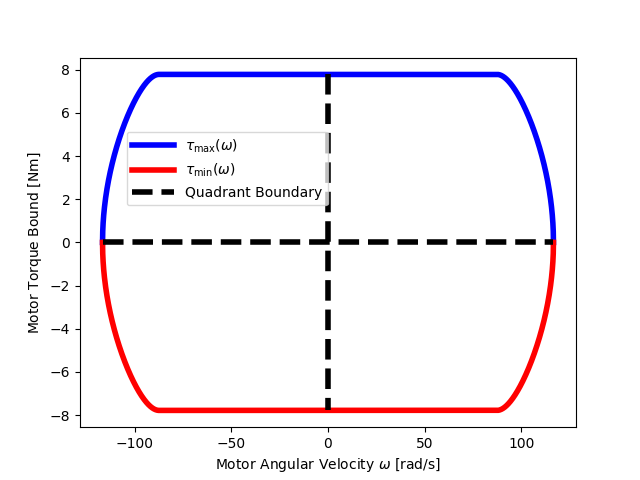}}
\hfill
\subfloat[Motor torque bound for case $\Phi_{\mathrm{pm},j}/L_{\mathrm{d},j} = I_{\mathrm{max},j}$.]
{\label{fig:motor_torque_bound:0} \includegraphics[width=0.32\linewidth]{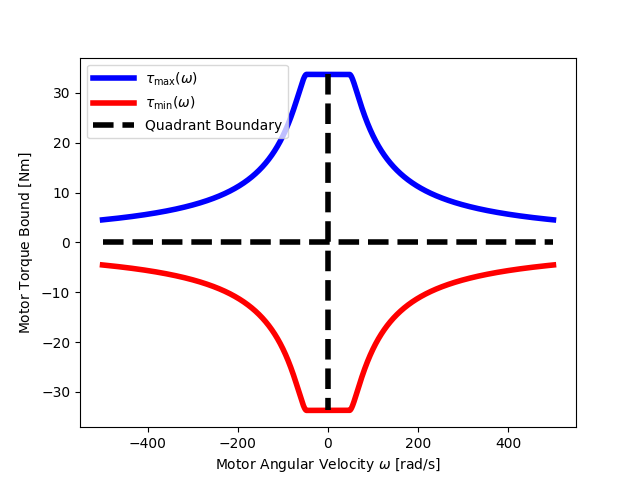}}
\hfill
\subfloat[Motor torque bound for case $\Phi_{\mathrm{pm},j}/L_{\mathrm{d},j} - I_{\mathrm{max},j} < 0$.]
{\label{fig:motor_torque_bound:negative} \includegraphics[width=0.32\linewidth]{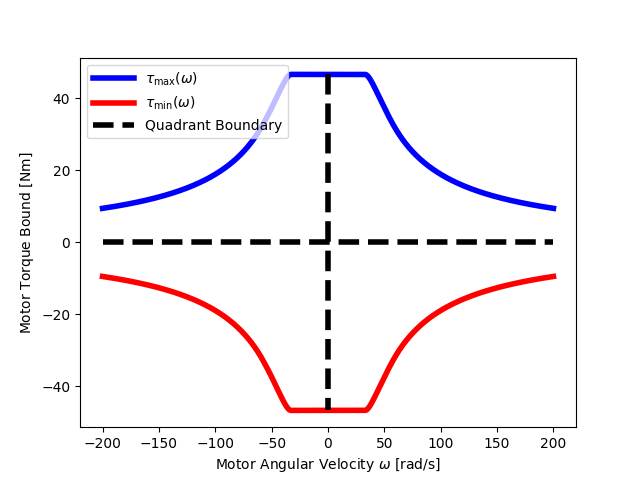}}
\caption{Motor torque bound in all three cases.} \label{fig:motor_torque_bound}
\end{figure*}

Thus, when $\Phi_{\mathrm{pm},j}/L_{\mathrm{d},j} - I_{\mathrm{max},j} < 0$, the maximum motor torque $\tau_{\mathrm{max},j}(\omega_j, \boldsymbol{\beta}_j)$, as a function of motor speed and motor design parameters, is given by
\begin{equation} \label{eq:motor_torque_max_negative}
\tau_{\mathrm{max},j} = \begin{cases}
1.5p \Phi_{\mathrm{pm},j}I_{\mathrm{max},j},& \text{if } |\omega_j| \leq \omega_{\mathrm{r},j}\\
1.5p \Phi_{\mathrm{pm},j}i_{\mathrm{q,lim},j},& \text{if } |\omega_j| \in [\omega_{\mathrm{r},j},\omega_{\mathrm{s},j}] \\
1.5p \Phi_{\mathrm{pm},j}i_{\mathrm{q,lim,V},j},& \text{if } |\omega_j| \in [\omega_{\mathrm{s},j},\infty)
\end{cases}
\end{equation}
where $\omega_{\mathrm{r},j}$, $i_{\mathrm{q,lim},j}$, $\omega_{\mathrm{s},j}$, and $i_{\mathrm{q,lim,V},j}$ are given by \eqref{eq:w_corner_define}, \eqref{eq:current_limit_geq}, \eqref{eq:omega_switch_define}, and \eqref{eq:iq_limit_negative}, respectively.
This piecewise-defined function is visualized in Fig. \ref{fig:torque_map:negative}.
The yellow dash-dot vertical line represents $\omega_{\mathrm{r},j}$.
The grey dash-dot vertical line represents $\omega_{\mathrm{s},j}$, which is the critical motor speed switching from activating both the current and voltage constraints to only the voltage constraint.
The black dashed horizontal line represents the constant maximum torque given by the first row of \eqref{eq:motor_torque_max_negative}.
The blue dashed curve represents the maximum torque given by the second row of \eqref{eq:motor_torque_max_negative}.
The brown dashed curve represents the maximum torque given by the third row of \eqref{eq:motor_torque_max_negative}.
Fig. \ref{fig:torque_map:negative} verifies the analytical torque bound given the numerical result from Algorithm \ref{alg:motor_feasible_region}.
Regardless of the sign of $\Phi_{\mathrm{pm},j}/L_{\mathrm{d},j} - I_{\mathrm{max},j}$, the minimum motor torque is given by
\begin{equation} \label{eq:motor_toque_lower_bound}
\tau_{\mathrm{min},j}(\omega_j, \boldsymbol{\beta}_j) \triangleq -\tau_{\mathrm{max},j}(\omega_j, \boldsymbol{\beta}_j).
\end{equation}
Therefore, for all allowable motor speeds, the lower and upper bound of motor torque is illustrated in Fig. \ref{fig:motor_torque_bound}, based on the analytical piecewise functions \eqref{eq:motor_torque_max_positive}, \eqref{eq:motor_torque_max_0}, and \eqref{eq:motor_torque_max_negative}.
The motor operation maps from Fig. \ref{fig:torque_map} verify the correctness of the analytical functions for the torque bounds.

\section{Dynamics Modeling} \label{sec:system_modeling}
This section introduces the modeling methodology for a mobile manipulator's forward and inverse dynamics.

\subsection{Inverse Dynamics} \label{subsec:inverse_dynamics}

This subsection presents the recursive Newton-Euler algorithm (RNEA) for the inverse dynamics of a mobile manipulator with motor parameterization. The details are summarized in Algorithm \ref{alg:inverse_dyn_mobile}.
Its derivation follows the methodology from \cite[Chapter~8.9.3]{lynch2017modern} but is revised accordingly for mobile manipulators with motor parameterization.
First, this paper adopts the following assumption on each motor's gearbox.
\begin{assumption}[\bf{Massless Gearbox}] \label{assume:gearbox_massless}
The mass of the gearbox for Motor $k$ ($k \geq 3$) is negligible.
\end{assumption}

Assumption~\ref{assume:gearbox_massless} is adopted due to the absence of the detailed inertia modeling of harmonic drives. Otherwise, one can lump the inertia of the primary gear, respectively the secondary gear, into its corresponding rotor, respectively its corresponding link.
Remark~\ref{remark:mobile_base} discusses the reason why no motors are considered for the base.

To recursively model the inverse dynamics, one first needs to compute all the necessary quantities recursively, including the twist, its time derivative, and the velocity-product accelerations of each rigid body.
This procedure is reflected in Line 3 - Line 10 of Algorithm \ref{alg:inverse_dyn_mobile}.
The twist $\boldsymbol{\mathcal{V}}_{L_k}$ of Link $k$ is the sum of the twist of Link $k-1$ expressed in frame $\{L_k\}$ and the twist due to the joint velocity $\dot{\boldsymbol{q}}_k$ or $\dot{\boldsymbol{\theta}}_k$, i.e.
\begin{equation} \label{eq:twist_update}
\boldsymbol{\mathcal{V}}_{L_k} = \mathrm{Ad}_{\boldsymbol{T}_{L_k, L_{k-1}}} \boldsymbol{\mathcal{V}}_{L_{k-1}} + \boldsymbol{A}_{L_k} \dot{\boldsymbol{q}}_k.
\end{equation}
The velocity-product acceleration $\boldsymbol{\zeta}_{L_k}$ of Link $k$ is related to the acceleration of Link $k$ by the equation of motion, i.e.
\begin{equation} \label{eq:bias_force_update}
\boldsymbol{\zeta}_{L_k} = \mathring{\boldsymbol{A}}_{L_k} \dot{\boldsymbol{q}}_{k} + \mathrm{ad}_{\boldsymbol{\mathcal{V}}_{L_k}}\boldsymbol{A}_{L_k}\dot{\boldsymbol{q}}_{k}.
\end{equation}
The acceleration $\dot{\boldsymbol{\mathcal{V}}}_{L_k}$ of Link $k$, is given by the time derivative of $\boldsymbol{\mathcal{V}}_{L_k}$, i.e.
\begin{equation} \label{eq:twist_dot_update}
\dot{\boldsymbol{\mathcal{V}}}_{L_k} = \mathrm{Ad}_{\boldsymbol{T}_{L_k, L_{k-1}}} \dot{\boldsymbol{\mathcal{V}}}_{L_{k-1}} + \boldsymbol{\zeta}_{L_k} + \boldsymbol{A}_{L_k} \ddot{\boldsymbol{q}}_k.
\end{equation}
Similarly, Line 5 - Line 7 hold for Rotor $k$, $k \geq 3$.
Denote $\boldsymbol{\mathcal{V}}_{L_0} = \boldsymbol{0}_6$ as the twist of the base frame $\{L_0\}$ expressed in frame $\{L_0\}$. Gravity is treated as an acceleration of the base in the opposite direction, and thus $\dot{\boldsymbol{\mathcal{V}}}_{L_0} = -\boldsymbol{a}_{\mathrm{g}} = \matt{0 & 0 & 0 & 0 & 0 & 9.81}^{\top}$,
where $\boldsymbol{a}_{\mathrm{g}}$ denotes the gravity expressed in frame $\{L_0\}$.

\begin{algorithm2e}[h!]
\DontPrintSemicolon
\textbf{Inputs}: $\boldsymbol{q}_1, \dot{\boldsymbol{q}}_1, \ddot{\boldsymbol{q}}_1;\boldsymbol{A}_{L_{k}}, \boldsymbol{G}_{L_{k}}, \mathring{\boldsymbol{A}}_{L_k}, \boldsymbol{T}_{L_{k}, L_{k-1}}, \ k \in \llbracket 1, n+2 \rrbracket$; $\theta_{k}, \dot{\theta}_{k}, \ddot{\theta}_{k}, \boldsymbol{A}_{R_k}, \mathring{\boldsymbol{A}}_{R_k}, \boldsymbol{G}_{R_k}, \boldsymbol{T}_{R_k, L_{k-1}}, Z_k, \ k \in \llbracket 3, n+2 \rrbracket$\;
$\boldsymbol{\mathcal{V}}_{L_0} = \boldsymbol{0}_6, \dot{\boldsymbol{\mathcal{V}}}_{L_0} = \matt{\boldsymbol{0}_{1\times 5} & 9.81}^{\top}, \boldsymbol{q}_2 = \dot{\boldsymbol{q}}_2 = \ddot{\boldsymbol{q}}_2 = 0$ \;
\For{$k=1$  \textbf{to} $n+2$}{
\If{$k \geq 3$}{
    $\boldsymbol{\mathcal{V}}_{R_k} \gets \mathrm{Ad}_{\boldsymbol{T}_{R_k, L_{k-1}}} \boldsymbol{\mathcal{V}}_{L_{k-1}} + \boldsymbol{A}_{R_k} \dot{\theta}_k$\;
    $\boldsymbol{\zeta}_{R_k} \gets \mathring{\boldsymbol{A}}_{R_k} \dot{\theta}_{k} +  \mathrm{ad}_{\boldsymbol{\mathcal{V}}_{R_k}} \boldsymbol{A}_{R_k}\dot{\theta}_{k}$\;
    $\dot{\boldsymbol{\mathcal{V}}}_{R_k} \gets \mathrm{Ad}_{\boldsymbol{T}_{R_k, L_{k-1}}} \dot{\boldsymbol{\mathcal{V}}}_{L_{k-1}} + \boldsymbol{\zeta}_{R_k} + \boldsymbol{A}_{R_k} \ddot{\theta}_k$
}

$\boldsymbol{\mathcal{V}}_{L_k} \gets \mathrm{Ad}_{\boldsymbol{T}_{L_k, L_{k-1}}} \boldsymbol{\mathcal{V}}_{L_{k-1}} + \boldsymbol{A}_{L_k} \dot{\boldsymbol{q}}_k$\;
$\boldsymbol{\zeta}_{L_k} \gets \mathring{\boldsymbol{A}}_{L_k} \dot{\boldsymbol{q}}_{k} + \mathrm{ad}_{\boldsymbol{\mathcal{V}}_{L_k}} \boldsymbol{A}_{L_k}\dot{\boldsymbol{q}}_{k}$\;
$\dot{\boldsymbol{\mathcal{V}}}_{L_k} \gets \mathrm{Ad}_{\boldsymbol{T}_{L_k, L_{k-1}}} \dot{\boldsymbol{\mathcal{V}}}_{L_{k-1}} + \boldsymbol{\zeta}_{L_k} + \boldsymbol{A}_{L_k} \ddot{\boldsymbol{q}}_k$\;
}
\For{$k=n+2$  \textbf{to} $1$}{
$\boldsymbol{\mathcal{F}}_{\mathrm{I}, L_k} \gets \boldsymbol{G}_{L_k} \dot{\boldsymbol{\mathcal{V}}}_{L_k} - \mathrm{ad}_{\boldsymbol{\mathcal{V}}_{L_k}}^{\top} \boldsymbol{G}_{L_k}\boldsymbol{\mathcal{V}}_{L_k}$\;
\If{$k \geq 3$}{
$\boldsymbol{\mathcal{F}}_{\mathrm{I}, R_k} \gets \boldsymbol{G}_{R_k} \dot{\boldsymbol{\mathcal{V}}}_{R_k} - \mathrm{ad}^{\top}_{\boldsymbol{\mathcal{V}}_{R_{k}}} \boldsymbol{G}_{R_{k}} \boldsymbol{\mathcal{V}}_{R_{k}}$\;
}
\uIf{$k \leq n+1$}
{
\lIf{$k \geq 2$}{
$\boldsymbol{\mathcal{F}}_{L_k} \gets \boldsymbol{\mathcal{F}}_{\mathrm{I}, L_k} + \mathrm{Ad}^{\top}_{\boldsymbol{T}_{L_{k+1}, L_k}} \boldsymbol{\mathcal{F}}_{L_{k+1}} + \mathrm{Ad}^{\top}_{\boldsymbol{T}_{R_{k+1},L_k}} \boldsymbol{\mathcal{F}}_{\mathrm{I}, R_{k+1}}$
}
\lElse
{
$\boldsymbol{\mathcal{F}}_{L_k} \gets \boldsymbol{\mathcal{F}}_{\mathrm{I}, L_k} + Ad^{\top}_{\boldsymbol{T}_{L_{k+1}, L_k}} \boldsymbol{\mathcal{F}}_{L_{k+1}}$
}
}
\lElse
{
$\boldsymbol{\mathcal{F}}_{L_k} \gets \boldsymbol{\mathcal{F}}_{\mathrm{I}, L_k}$
}
\uIf{$k \geq 3$}{
$\tau_{k,\mathrm{gear}} \gets \boldsymbol{A}^{\top}_{L_k} \boldsymbol{\mathcal{F}}_{L_k}$\;
$\tau_{k} \gets \tau_{k,\mathrm{gear}} / Z_k + \boldsymbol{A}_{R_k}^{\top} \boldsymbol{\mathcal{F}}_{\mathrm{I}, R_k}/Z_k$
}
\lElse
{
$\boldsymbol{f}_k \gets \boldsymbol{A}^{\top}_{L_k} \boldsymbol{\mathcal{F}}_{L_k}$
}
}
\Return{$\boldsymbol{f}_1, \tau_{k}, k=3, \cdots, n+2$}
\caption{Recursive Newton-Euler algorithm for the inverse dynamics of a mobile manipulator}
\label{alg:inverse_dyn_mobile}
\end{algorithm2e}

\begin{figure}[!ht]
\centering
\includegraphics[width=0.40\linewidth]{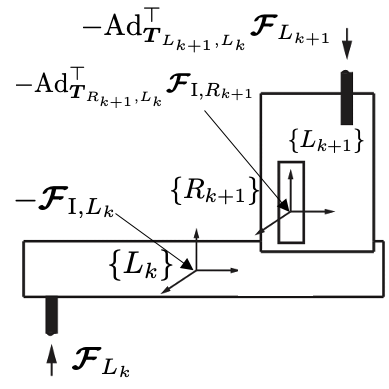}
\caption{Free-body diagram for Link $k$ and Link $k+1$. Revised based on Fig. 8.12 of \cite{lynch2017modern}.}
\label{fig:force_gear_arm}
\end{figure}

Secondly, one needs to identify all the wrenches applied on Link $k$ and Link $k+1$ ($k \geq 2$) with the existence of Rotor $k+1$. This procedure is reflected in Line 11 - Line 18 of Algorithm \ref{alg:inverse_dyn_mobile}.
The free-body diagram for Link $k$ and Link $k+1$ is illustrated in Fig. \ref{fig:force_gear_arm}. Link $k$ receives a wrench $\boldsymbol{\mathcal{F}}_{L_k}$ from Motor $k$'s gearhead (expressed in frame $\{L_k\}$) and a fictitious force (or inertia force) due to the inertia of Link $k$ (expressed in frame $\{L_k\}$). Link $k+1$ receives a wrench $\boldsymbol{\mathcal{F}}_{L_{k+1}}$ from Motor $k+1$'s gearhead (expressed in frame $\{L_{k+1}\}$) and a fictitious force due to the inertia of Rotor $k+1$ (expressed in frame $\{R_{k+1}\}$).
To represent all the wrenches in the same frame $\{L_k\}$, the adjoint representation of $\boldsymbol{T}_{L_k, L_{k+1}}$ and $\boldsymbol{T}_{L_k, R_{k+1}}$ are used to convert the wrenches from frame $\{L_{k+1}\}$ to frame $\{L_k\}$, and from $\{R_{k+1}\}$ to frame $\{L_k\}$, respectively. Thus, the following equation, equivalent to Line 16, holds for each pair of Link $k$ and Link $k+1$, $k \geq 2$.
\begin{equation} \label{eq:diagram_k_kp1}
\boldsymbol{\mathcal{F}}_{L_k} - \boldsymbol{\mathcal{F}}_{\mathrm{I}, L_k} - \mathrm{Ad}^{\top}_{\boldsymbol{T}_{L_{k+1}, L_k}} \boldsymbol{\mathcal{F}}_{L_{k+1}} - \mathrm{Ad}^{\top}_{\boldsymbol{T}_{R_{k+1},L_k}} \boldsymbol{\mathcal{F}}_{\mathrm{I}, R_{k+1}} = \boldsymbol{0}.
\end{equation}
Note that when $k=1$, since there is no rotor, \eqref{eq:diagram_k_kp1} is reduced to
\begin{equation} \label{eq:diagram_1_2}
\boldsymbol{\mathcal{F}}_{L_k} - \boldsymbol{\mathcal{F}}_{\mathrm{I}, L_k} - \mathrm{Ad}^{\top}_{\boldsymbol{T}_{L_{k+1}, L_k}} \boldsymbol{\mathcal{F}}_{L_{k+1}} = \boldsymbol{0},
\end{equation}
which is equivalent to Line 17.
The fictitious forces of Link $k$ and Rotor $k$ are given by Line 12 and Line 14, respectively.

Finally, the actuator at each Joint $k$ only has to provide force or torque in its motion subspace, which results in Line 20 and Line 22 of Algorithm \ref{alg:inverse_dyn_mobile}. The torque for Motor $k$ ($k \geq 3$) is the torque at Rotor $k$, which is given by the sum of the torque transmitted from the gearhead and the fictitious force of Rotor $k$ projected on Rotor $k$'s motion subspace, i.e. Line 21. $\boldsymbol{f}_1 \triangleq \text{col}\{ \tau_{1,\mathrm{z}}, f_{1,\mathrm{x}}, f_{1,\mathrm{y}} \} \in \mathbb{R}^3$ denotes the input wrench applied on Joint 1, whose vector element represents the torque applied around Joint 1's z-axis and the forces applied along Joint 1's x- and y-axis. $\tau_{k}$ $(k=3, \cdots, n+2)$ denotes the input motor torque applied on Joint $3, \cdots, n+2$. The input of the entire mobile manipulator is denoted by $\boldsymbol{u} \triangleq \text{col}\{ \boldsymbol{f}_1, \tau_{3}, \cdots, \tau_{n+2}\} \in \mathbb{R}^{n+3}$.

\subsection{Forward Dynamics} \label{subsec:forward_dynamics}

This subsection presents the articulated-body algorithm (ABA) for the forward dynamics of a mobile manipulator with motor parameterization. The details are summarized in Algorithm \ref{alg:ABA_mobile}.
The derivation of Algorithm \ref{alg:ABA_mobile} follows the methodology from \cite[Chapter~7.3]{featherstone2014rigid}, but is revised accordingly for mobile manipulators with motor parameterization.

First, one needs to define how the rigid bodies are articulated together since it determines how bias forces and spatial inertia matrices are updated recursively when using ABA to propagate forward dynamics. As illustrated in Fig. \ref{fig:geared_motor_scheme}, for $k \geq 3$, Rotor $k$ is articulated with Link $k$ via its gearbox, and Link $k-1$ is articulated with Link $k$ via its gearbox. Link 1 is articulated (or rigidly connected) with Link 2 via Joint 2, a 0-DOF joint.

Then, Line 3 - Line 12 of Algorithm \ref{alg:ABA_mobile} recursively calculates and initializes the twists, velocity-product accelerations, spatial inertia matrices, and bias forces of each link and rotor from the base to the last link. Specifically, Line 11 and Line 12 initialize the spatial inertia matrices $\boldsymbol{\mathcal{I}}^A_{L_k}$ and the bias forces $\boldsymbol{p}^A_{L_k}$ for each link, respectively. Line 7 and Line 8 initialize the same quantities for each rotor, i.e. $\boldsymbol{\mathcal{I}}^A_{R_k}$ and $\boldsymbol{p}^A_{R_k}$, respectively.

Thirdly, Line 13 - Line 34 recursively updates the spatial inertia matrices and bias forces from the last link to the first link.
Specifically, Line 14 - Line 22 defines some intermediate quantities that will be used later. Note that this part skips Link 2 because there is no input applying on Link 2 and Link 2 is rigidly connected with Link 1.
The reason for not combining Link 1 and Link 2 as one rigid body is for the ease of modularization.
Line 24 - Line 27 updates Link 1's spatial inertia matrix $\boldsymbol{\mathcal{I}}^A_{L_1}$ and bias force $\boldsymbol{p}^A_{L_1}$ from Link 2.
Line 31 - Line 32 updates Rotor $k$'s spatial inertia matrix and bias force given Link $k$'s effect.
Line 33 - Line 34 updates Link $k-1$'s spatial inertia matrix and bias force given Link $k$'s effect.

Finally, Line 35 - Line 40 recursively calculates the joint accelerations from the base to the last link. Note that the joint acceleration for Joint 2 is constantly zero since Link 2 is rigidly connected with Link 1.

\begin{algorithm2e}
\textbf{Inputs}: $\boldsymbol{q}_1, \dot{\boldsymbol{q}}_1, \boldsymbol{f}_1; \boldsymbol{A}_{L_{k}}, \boldsymbol{G}_{L_{k}}, \mathring{\boldsymbol{A}}_{L_k}, \boldsymbol{T}_{L_k,L_{k-1}}, \ k \in \llbracket 1, n+2 \rrbracket$; $\theta_{k}, \dot{\theta}_{k}, \tau_{k}, \boldsymbol{A}_{R_k}, \mathring{\boldsymbol{A}}_{R_k}, \boldsymbol{G}_{R_k}, \boldsymbol{T}_{R_k,L_{k-1}}, Z_k, \ k \in \llbracket 3, n+2 \rrbracket$\;
$\boldsymbol{\mathcal{V}}_{L_0} = \boldsymbol{0}_6, \dot{\boldsymbol{\mathcal{V}}}_{L_0} = \matt{\boldsymbol{0}_{1\times 5} & 9.81}^{\top}, \boldsymbol{q}_2 = \dot{\boldsymbol{q}}_2 = 0$ \;
\For{$k=1$ \textbf{to} $n+2$}{
	\If{$k \geq 3$}{
		$\boldsymbol{\mathcal{V}}_{R_k} \gets \mathrm{Ad}_{\boldsymbol{T}_{R_k,L_{k-1}}} \boldsymbol{\mathcal{V}}_{L_{k-1}} + \boldsymbol{A}_{R_k}\dot{\theta}_{k}$\;
		$\boldsymbol{\zeta}_{R_k} \gets \mathring{\boldsymbol{A}}_{R_k} \dot{\theta}_{k} +  \mathrm{ad}_{\boldsymbol{\mathcal{V}}_{R_k}} \boldsymbol{A}_{R_k}\dot{\theta}_{k}$\;
		$\boldsymbol{\mathcal{I}}^A_{R_k} \gets \boldsymbol{G}_{R_k}$\;
		$\boldsymbol{p}^A_{R_k} \gets \mathrm{ad}_{\boldsymbol{\mathcal{V}}_{R_k}} \boldsymbol{\mathcal{I}}^A_{R_k} \boldsymbol{\mathcal{V}}_{R_k}$\;
	}
	$\boldsymbol{\mathcal{V}}_{L_k} \gets \mathrm{Ad}_{\boldsymbol{T}_{L_k,L_{k-1}}} \boldsymbol{\mathcal{V}}_{L_{k-1}} + \boldsymbol{A}_{L_k}\dot{\boldsymbol{q}}_k$\;
	$\boldsymbol{\zeta}_{L_k} \gets \mathring{\boldsymbol{A}}_{L_k} \dot{\boldsymbol{q}}_{k} + \mathrm{ad}_{\boldsymbol{\mathcal{V}}_{L_k}} \boldsymbol{A}_{L_k}\dot{\boldsymbol{q}}_{k}$\;
	$\boldsymbol{\mathcal{I}}^A_{L_k} \gets \boldsymbol{G}_{L_k}$\;
	$\boldsymbol{p}^A_{L_k} \gets \mathrm{ad}_{\boldsymbol{\mathcal{V}}_{L_k}} \boldsymbol{\mathcal{I}}^A_{L_k} \boldsymbol{\mathcal{V}}_{L_k}$\;
}
\For{$k=n+2$  \textbf{to} $1$}{
	$\boldsymbol{U}_{L_k} \gets \boldsymbol{\mathcal{I}}^A_{L_k} \boldsymbol{A}_{L_k}$\;
	\uIf{$k = 1$}
	{
        $\boldsymbol{D}_k \gets (\boldsymbol{A}_{L_k}^\top \boldsymbol{U}_{L_k})^{-1}$\;
        $\boldsymbol{\mu}_{L_k} \gets \boldsymbol{f}_k - \boldsymbol{A}_{L_k}^{\top} \boldsymbol{p}^A_{L_k}$\;}
	\ElseIf{$k \geq 3$}
	{
        $\boldsymbol{\mu}_{L_k} \gets Z_k \tau_{k} - \boldsymbol{A}_{L_k}^{\top} \boldsymbol{p}^A_{L_k}$\;
	$\boldsymbol{U}_{R_k} \gets \boldsymbol{\mathcal{I}}^A_{R_k} \boldsymbol{A}_{R_k}$\;
	$\boldsymbol{\mu}_{R_k} \gets  Z_k \tau_{k} - \boldsymbol{A}_{R_k}^{\top} \boldsymbol{p}^A_{R_k}$\;
	$\boldsymbol{D}_k \gets (\boldsymbol{A}_{L_k}^\top \boldsymbol{U}_{L_k} + \boldsymbol{A}_{R_k}^\top \boldsymbol{U}_{R_k})^{-1}$\;
	}

    \uIf{$k = 2$}
    {
    $\boldsymbol{I}_{L_k}^a = \boldsymbol{\mathcal{I}}^A_{L_k}$\; $\boldsymbol{p}_{L_k}^a = \boldsymbol{p}^A_{L_k} + \boldsymbol{I}_{L_k}^a \boldsymbol{\zeta}_{L_k}$\;

    $\boldsymbol{\mathcal{I}}^A_{L_{k-1}} \gets \boldsymbol{\mathcal{I}}^A_{L_{k-1}} + \mathrm{Ad}_{\boldsymbol{T}_{L_k,L_{k-1}}}^\top \boldsymbol{I}_{L_k}^a \mathrm{Ad}_{\boldsymbol{T}_{L_k,L_{k-1}}}$\;
    $\boldsymbol{p}^A_{L_{k-1}} \gets \boldsymbol{p}^A_{L_{k-1}} + \mathrm{Ad}_{\boldsymbol{T}_{L_k,L_{k-1}}}^\top \boldsymbol{p}_{L_k}^a$\;
    }
    \ElseIf{$k \geq 3$}
    {
    $\boldsymbol{I}_{L_k}^a \gets \boldsymbol{\mathcal{I}}^A_{L_k} - \boldsymbol{U}_{L_k} \boldsymbol{D}_k \boldsymbol{U}_{L_k}^{\top}$\;
    $\boldsymbol{p}_{L_k}^a \gets \boldsymbol{p}^A_{L_k} + \boldsymbol{I}_{L_k}^a \boldsymbol{\zeta}_{L_k} + \boldsymbol{U}_{L_k} \boldsymbol{D}_k\boldsymbol{\mu}_{L_k}$\;	
    $\boldsymbol{I}_{R_k}^a \gets \boldsymbol{\mathcal{I}}^A_{R_k} - \boldsymbol{U}_{R_k} \boldsymbol{D}_k \boldsymbol{U}_{R_k}^{\top}$\;
    $\boldsymbol{p}_{R_k}^a \gets \boldsymbol{p}^A_{R_k} + \boldsymbol{I}_{R_k}^a \boldsymbol{\zeta}_{R_k} + \boldsymbol{U}_{R_k} \boldsymbol{D}_k\boldsymbol{\mu}_{R_k}$\;

    $\boldsymbol{\mathcal{I}}^A_{L_{k-1}} \gets \boldsymbol{\mathcal{I}}^A_{L_{k-1}} + \mathrm{Ad}_{\boldsymbol{T}_{L_k,L_{k-1}}}^\top \boldsymbol{I}_{L_k}^a \mathrm{Ad}_{\boldsymbol{T}_{L_k,L_{k-1}}} + \mathrm{Ad}_{\boldsymbol{T}_{R_k,L_{k-1}}}^\top \boldsymbol{I}_{R_k}^a \mathrm{Ad}_{\boldsymbol{T}_{R_k,L_{k-1}}}$\;
    $\boldsymbol{p}^A_{L_{k-1}} \gets \boldsymbol{p}^A_{L_{k-1}} + \mathrm{Ad}_{\boldsymbol{T}_{L_k,L_{k-1}}}^\top \boldsymbol{p}_{L_k}^a + \mathrm{Ad}_{\boldsymbol{T}_{R_k,L_{k-1}}}^\top \boldsymbol{p}_{R_k}^a$\;
    }
} 
\For{$k=1$  \textbf{to} $n+2$}{
$\hat{\boldsymbol{a}}_k \gets \mathrm{Ad}_{\boldsymbol{T}_{L_k,L_{k-1}}} \dot{\boldsymbol{\mathcal{V}}}_{L_{k-1}}$\;

\lIf{$k = 1$}
{$\ddot{\boldsymbol{q}}_{k} \gets \boldsymbol{D}_k(\boldsymbol{\mu}_{L_k} - \boldsymbol{U}_{L_k}^{\top}(\hat{\boldsymbol{a}}_k + \boldsymbol{\zeta}_{L_k}))$}
\lElseIf{$k = 2$}
{$\ddot{\boldsymbol{q}}_{k} \gets 0$}
\lElse{$\ddot{\theta}_{k} \gets \boldsymbol{D}_k(\boldsymbol{\mu}_{L_k} - \boldsymbol{U}_{L_k}^{\top}(\hat{\boldsymbol{a}}_k + \boldsymbol{\zeta}_{L_k}) - \boldsymbol{U}_{R_k}^{\top}(\hat{\boldsymbol{a}}_k + \boldsymbol{\zeta}_{R_k}) - \boldsymbol{A}_{R_k}^{\top} \boldsymbol{p}^A_{R_k})$}

$\dot{\boldsymbol{\mathcal{V}}}_{L_k} \gets \hat{\boldsymbol{a}}_k + \boldsymbol{\zeta}_{L_k} + \boldsymbol{A}_{L_k}\ddot{\boldsymbol{q}}_{k}$\;
} 
\textbf{Return}: $\ddot{\boldsymbol{q}}_{1}, \ddot{\theta}_{k}, k = 3, \cdots, n+2$\;
\caption{Articulated-body algorithm for the forward dynamics of a mobile manipulator}
\label{alg:ABA_mobile}
\end{algorithm2e}

\subsection{Modeling Validation}

This subsection presents the modeling validation by comparing the forward dynamics and the inverse dynamics obtained from Algorithm \ref{alg:ABA_mobile} and Algorithm \ref{alg:inverse_dyn_mobile}, respectively.
Since both algorithms are derived by different methodologies, it is reasonable to infer that the modeling is correct if there is no difference or only numerical error between the numerical results from the two algorithms. The definition of the so-called numerical results is given below.

First, the continuous-time forward dynamics of the mobile manipulator can be obtained symbolically by Algorithm \ref{alg:ABA_mobile}.
One could obtain a discrete-time trajectory of states and inputs by solving a motion planning problem, which will be introduced in Section \ref{sec:joint_locomotion_planning}.
Then at each time instance, one could compute the joint accelerations $\ddot{\boldsymbol{q}}_1, \ddot{\boldsymbol{\theta}}_k, k \geq 3$ given the current state, i.e. the joint positions $\boldsymbol{q}_1, \boldsymbol{\theta}_k$ and the velocities $\dot{\boldsymbol{q}}_1, \dot{\boldsymbol{\theta}}_k$, and the current input $\boldsymbol{f}_1, \tau_{k}, k \geq 3$ with the continuous-time forward dynamics.
Finally, at each time instance, with Algorithm \ref{alg:inverse_dyn_mobile}, given $\boldsymbol{q}_1, \boldsymbol{\theta}_k$, $\dot{\boldsymbol{q}}_1, \dot{\boldsymbol{\theta}}_k$, and $\ddot{\boldsymbol{q}}_1, \ddot{\boldsymbol{\theta}}_k$, one could compute the desired input and compare the inputs from the forward dynamics and the inverse dynamics.
Since this process only evaluates the state and input by continuous-time forward dynamics at each time instance, how the continuous-time forward dynamics are discretized will not affect the comparison accuracy. In other words, the comparison does not involve forward propagating the continuous-time dynamics and thus there is no accumulated error regarding how to integrate the continuous-time dynamics. The only reason why this process involves motion planning is to generate a reasonable trajectory of states and inputs for point-wise evaluation.

The statistics about the error between the forward and inverse dynamics are summarized in Table \ref{table:error_forward_inverse}, where the percentage error is defined by the error percentage regarding the maximum absolute value of this variable's trajectory. Based on Table \ref{table:error_forward_inverse}, the error might only be a relatively large round-off error. According to \cite[Chapter~10]{featherstone2014rigid}, large round-off errors can arise during dynamics calculations for the following reasons: (1) using a far-away coordinate system; (2) large velocities; (3) large inertia ratios.
For reason (1), a coordinate system is far away from a rigid body if the distance between the origin and the center of mass is many times the body’s radius of gyration. Ideally, this distance should be no more than about one or two radii of gyration \cite[Chapter~10]{featherstone2014rigid}. Thus this paper conjectures that using a global coordinate system for mobile manipulators can cause large round-off errors.
For reason (3), one generally requires the base to have a larger inertia than the manipulator for stability, and thus a large inertia ratio is also a possible reason for large round-off errors.
How to compute mobile manipulator dynamics in a more numerically stable coordinate system requires further investigation.

\begin{table}
\centering
\begin{threeparttable}
\caption{Error between Forward and Inverse Dynamics} \label{table:error_forward_inverse}
\begin{tabular}{c c c c c}
\toprule
Name & mean $\pm$ std & quartiles$^\dagger$ & max & mean \\
\midrule
$\tau_{1,\mathrm{z}}$ & $0.154\pm0.113$ & 0.051, 0.149, 0.229 & 0.417 & 0.41 \% \\
$f_{1,\mathrm{x}}$ & $0.188\pm0.114$ & 0.102, 0.193, 0.235 & 0.492 & 0.23 \% \\
$f_{1,\mathrm{y}}$ & $0.226\pm0.173$ & 0.104, 0.196, 0.281 & 0.706 & 0.24 \% \\
$\tau_3$ & $0.003\pm0.002$ & 0.001, 0.003, 0.004 & 0.008 & 1.26 \% \\
$\tau_4$ & $0.002\pm0.002$ & 0.001, 0.001, 0.003 & 0.013 & 0.07 \% \\
$\tau_5$ & $0.002\pm0.003$ & 0.001, 0.001, 0.003 & 0.013 & 0.21 \% \\
$\tau_6$ & $0.003\pm0.003$ & 0.001, 0.002, 0.005 & 0.013 & 1.97 \% \\
$\tau_7$ & $0.002\pm0.002$ & 0.000, 0.001, 0.002 & 0.011 & 0.37 \% \\
$\tau_8$ & $0.003\pm0.002$ & 0.001, 0.002, 0.004 & 0.008 & 1.57 \% \\
\bottomrule
\end{tabular}
\begin{tablenotes}
\small
\item[$\dagger$] 25th, 50th, 75th percentile
\item[$\ddagger$] If no indication, unit for torque is Nm; unit for force is N
\end{tablenotes}
\end{threeparttable}
\centering
\end{table}

\section{Integrated Locomotion and Manipulation Planning} \label{sec:joint_locomotion_planning}

This section introduces an optimization-based integrated locomotion and manipulation planning given the differentiable dynamics of a mobile manipulator with motor parameterization.
Consider a mobile manipulator with $n = 6$.
The motor design variables for manipulator's $j$-th motor, $j \in \llbracket 3, n+2 \rrbracket$, are denoted by $l_{j}$, $r_{\mathrm{ro},j}$, $r_{\mathrm{so},j}$, $h_{\mathrm{m},j}$, $h_{\mathrm{sy},j}$, $w_{\mathrm{tooth},j}$, $b_{0,j}$.
And further denote
\begin{equation*}
\begin{split}
\boldsymbol{\beta}_j &\triangleq \text{col}\{l_{j}, r_{\mathrm{ro},j}, r_{\mathrm{so},j}, h_{\mathrm{m},j}, h_{\mathrm{sy},j}, w_{\mathrm{tooth},j}, b_{0,j}\}, \\
\boldsymbol{\beta} &\triangleq \text{col}\{ \boldsymbol{\beta}_{3}, \cdots, \boldsymbol{\beta}_{n+2}\} \in \mathbb{R}^{7n},
\end{split}
\end{equation*}
where $\boldsymbol{\beta}$ indicates all the motor design parameters of the mobile manipulator.
Denote $\boldsymbol{\theta} \triangleq \text{col}\{ \theta_3, \cdots, \theta_{n+2} \} \in \mathbb{R}^n$ as the joint angular positions of the arm and denote $\dot{\boldsymbol{\theta}} \in \mathbb{R}^n$ as the joint velocity similarly.
The complete forward dynamics for the mobile manipulator, detailed in Algorithm \ref{alg:ABA_mobile},  are abstracted as
\begin{equation} \label{eq:mani_dyn}
\dot{\boldsymbol{x}} = \boldsymbol{f}_{\mathrm{c}}(\boldsymbol{x}(t), \boldsymbol{u}(t), \boldsymbol{\beta}),
\end{equation}
where $\boldsymbol{x} \triangleq \text{col} \{ \boldsymbol{q}_1, \boldsymbol{\theta}, \dot{\boldsymbol{q}}_1, \dot{\boldsymbol{\theta}} \} \in \mathbb{R}^{2(n+3)}$ and $\boldsymbol{u} \triangleq \text{col}\{ \boldsymbol{f}_1, \tau_{3}, \cdots, \tau_{n+2}\} \in \mathbb{R}^{n+3}$.

Given a task, a trajectory planning optimization can return its optimal trajectories of states and controls that align with the task requirements:
\begin{mini!}|s|
{\boldsymbol{x}(t), \boldsymbol{u}(t), t_{\mathrm{f}}}{ J(\boldsymbol{x}(t), \boldsymbol{u}(t), t_{\mathrm{f}}) \label{oc:obj}}
{\label{oc}}{}
\addConstraint{ \dot{\boldsymbol{x}} = \boldsymbol{f}_{\mathrm{c}}(\boldsymbol{x}(t), \boldsymbol{u}(t), \boldsymbol{\beta}) \label{oc:dynamics}}
\addConstraint{\forall t \in [0, t_{\mathrm{f}}] \  \text{with given } \boldsymbol{x}(0) \label{oc:initial_cond}}
\addConstraint{\text{other constraints on } \boldsymbol{x}(t), \boldsymbol{u}(t),  \label{oc:other_con}}
\end{mini!}
where $t_{\mathrm{f}} > 0$ denotes the final time for trajectory planning, which could either be a decision variable or a prescribed parameter.
Note that \eqref{oc} adopts the most general form, which represents time-optimal trajectory planning, trajectory tracking, or energy-optimal trajectory planning. For the latter two cases, $t_{\mathrm{f}}$ is a fixed prescribed parameter.
Given a particular value of $\boldsymbol{\beta}$, the optimal states $\boldsymbol{x}^*(t)$, controls $\boldsymbol{u}^*(t)$, and final time $t_{\mathrm{f}}^*$ (if applicable) optimize the cost function $J$.

\begin{remark}
One can combine the motor dynamics \eqref{eq:motor_dynamics_PMSM} with the manipulator dynamics \eqref{eq:mani_dyn} together as an entire system with its state $\boldsymbol{q}_1, \boldsymbol{\theta}, \dot{\boldsymbol{q}}_1, \dot{\boldsymbol{\theta}}, i_{\mathrm{d},j},  i_{\mathrm{q},j}$ and its input $\boldsymbol{f}_1, \tau_{3}, \cdots, \tau_{n+2}, u_{\mathrm{d},j}, u_{\mathrm{q},j}$.
However, the motor dynamics typically operate at frequencies in the thousands of Hertz, while the arm dynamics often run at frequencies in the hundreds of Hertz.
The frequency discrepancy necessitates discretizing the entire system with a time step much smaller than 1 millisecond, which can lead to an unnecessary computational burden.
In practice, it is reasonable to assume that between every two consecutive time steps of the manipulator dynamics, the motor dynamics have enough time to adjust their speed and torque to match the desired values, considering control constraints.
\end{remark}

\subsection{Discretization by Direct Collocation}

Discretization is typically used to efficiently solve the continuous-time problem \eqref{oc} with numerical solvers. Among different discretization methods, the direct collocation method minimizes the error between state derivatives from continuous-time dynamics and state derivatives from approximated polynomial differentiation on every control interval. For systems with fast and nonlinear dynamics with running frequencies higher than hundreds of Hz, direct collocation generally outperforms the Euler method (ODE1) and the Runge–Kutta methods such as ODE4.
To discretize the optimal control problem \eqref{oc} with direct collocation, the continuous-time dynamics $\dot{\boldsymbol{x}} = \boldsymbol{f}_{\mathrm{c}}(\boldsymbol{x}(t), \boldsymbol{u}(t), \boldsymbol{\beta})$ is replaced by a set of discrete-time collocation equations.

First, discretize the entire time horizon $[0, t_{\mathrm{f}}]$ into $N \in \mathbb{Z}_+$ uniform intervals, then the discretized time grids are given by $t_k = k \cdot t_{\mathrm{f}} / N$, $k = 0, \cdots, N$.
Denote the uniform time step as $\Delta \triangleq t_{\mathrm{f}} / N$ and
 the state at time $t_k$ as $\boldsymbol{x}_k \triangleq \boldsymbol{x}(t_k)$.
Let the control over the time interval $[t_k, t_{k+1})$ as constant, i.e. $\boldsymbol{u}(t_k) = \boldsymbol{u}_k, \ \forall t \in [t_k, t_{k+1}), \ k = 0, \cdots, N$.

Secondly, one needs to compute the coefficients of the polynomials on each interval $[t_k, t_{k+1})$ to ensure that the continuous-time dynamics (i.e. some ODEs) are exactly satisfied at the collocation points $t_{k,i} = t_k + c_i \Delta$, $i = 1, \cdots, n_{\mathrm{p}}$, where $n_{\mathrm{p}} \in \mathbb{Z}_+$ denotes the polynomial degree.
The collocation points $c_1 < \cdots < c_{n_{\mathrm{p}}}$ are defined in an open interval $(0,1)$. For completeness, the left endpoint $c_0 = 0$ is also included and thus $0 = c_0 < c_1 < \cdots < c_{n_{\mathrm{p}}} < 1$.
The choice of the collocation points determines the accuracy and numerical stability of the discretization and the optimal control solution.
This paper chooses the Gauss collocation points, obtained as the roots of a shifted Gauss-Jacobi polynomial, thanks to the good numerical stability and the attribute of living in the open interval $(0,1)$ \cite{magnusson2012collocation}.

Thirdly, to use the polynomials to approximate the continuous-time dynamics, one needs to define the basis of polynomials.
This paper utilizes the  Lagrange interpolation polynomial basis.
Define the basis for polynomials as $\ell_i: \mathbb{R} \mapsto \mathbb{R}$, $i = 0,\cdots,n_{\mathrm{p}}$, and
\begin{equation}
\ell_i(t) = \textstyle\prod_{j=0, i \neq j}^{n_{\mathrm{p}}} \frac{t - c_j}{c_i - c_j}, \ i\in \llbracket 0, n_{\mathrm{p}} \rrbracket.
\end{equation}
Then the approximated state trajectory $\boldsymbol{q}_k(t)$ on each time interval $[t_k, t_{k+1})$ is given by a linear combination of the basis functions:
\begin{equation*}
\boldsymbol{q}_k(t) = \textstyle\sum_{j=0}^{n_{\mathrm{p}}} \ell_j(\frac{t-t_k}{\Delta}) \boldsymbol{x}_{k,j},
\end{equation*}
where $\boldsymbol{x}_{k,j} \in \mathbb{R}^{2(n+3)}$, $\forall k \in \llbracket 0, N \rrbracket, j = \llbracket 1, n_{\mathrm{p}} \rrbracket$ is the new decision variable for the discretized optimal control problem.
By differentiation, the approximated time derivative of the state at each collocation point over the interval $[t_k, t_{k+1})$ is given by
\begin{equation} \label{eq:collocation_qk_dot}
\dot{\boldsymbol{q}}_k(t_{k,i}) = \frac{1}{\Delta} \sum_{j=0}^{n_{\mathrm{p}}} \dot{\ell}_j(c_i) \boldsymbol{x}_{k,j} \coloneqq \frac{1}{\Delta} \sum_{j=0}^{n_{\mathrm{p}}} \boldsymbol{C}_{j,i} \boldsymbol{x}_{k,j}, i \in \llbracket 0, n_{\mathrm{p}} \rrbracket,
\end{equation}
where $\dot{\ell}_j(c_i) = \frac{d \ell_j(t)}{d t}|_{t = c_i}$;
$\boldsymbol{C}_{j,i} \in \mathbb{R}$ is a constant given a particular $n_{\mathrm{p}}$, polynomial basis and collocation points;
denote $\boldsymbol{C} \in \mathbb{R}^{(n_{\mathrm{p}}+1) \times (n_{\mathrm{p}}+1)}$ as the differentiation matrix, where its element on the $j$-th row and $i$-th column is $\boldsymbol{C}_{j,i}$.
Note that even though $\boldsymbol{q}_k(t)$ is dependent on $\Delta \coloneqq t_{\mathrm{f}} / N$, according to \eqref{eq:collocation_qk_dot}, the constant matrix $\boldsymbol{C}$ is independent on $t_{\mathrm{f}}$ and $N$.
The state at the end of each time interval $[t_k, t_{k+1})$, i.e. at time $t_{k+1}$, is given by
\begin{equation}
\boldsymbol{x}_{k+1} = \textstyle\sum_{j=0}^{n_{\mathrm{p}}} \ell_j(1) \boldsymbol{x}_{k,j} \coloneqq \sum_{j=0}^{n_{\mathrm{p}}} \boldsymbol{D}_j \boldsymbol{x}_{k,j},
\end{equation}
where $\boldsymbol{D}_{j} \in \mathbb{R}$ is a constant;
denote $\boldsymbol{D} \in \mathbb{R}^{n_{\mathrm{p}}+1}$ as the continuity vector, where its element on the $j$-th row is $\boldsymbol{D}_{j}$.

In many applications, the objective function \eqref{oc:obj} typically includes an integration of a stage cost and a terminal cost, i.e.
\begin{equation*}
J = \boldsymbol{h}(\boldsymbol{x}(t_{\mathrm{f}}), t_{\mathrm{f}}) + \int_{0}^{t_{\mathrm{f}}} \boldsymbol{c}(\boldsymbol{x}(t), \boldsymbol{u}(t)) dt,
\end{equation*}
where $\boldsymbol{h}: \mathbb{R}^{2(n+3)} \times \mathbb{R}_+ \mapsto \mathbb{R}$ is the terminal cost function.; $\boldsymbol{c}: \mathbb{R}^{2(n+3)} \times \mathbb{R}^{n+3} \mapsto \mathbb{R}$ is the stage cost function.
Then, using the approximation $\boldsymbol{q}_k(t)$, one can integrate the stage cost over each time interval $[t_k, t_{k+1})$ given constant control $\boldsymbol{u}_k$ and obtain the following:
\begin{equation}
\begin{split}
&\int_{t_k}^{t_{k+1}} \textstyle\sum_{j=0}^{n_{\mathrm{p}}} \ell_j(\frac{t-t_k}{\Delta}) \boldsymbol{c}(\boldsymbol{x}_{k,j}, \boldsymbol{u}_k) dt = \\
& \Delta \sum_{j=0}^{n_{\mathrm{p}}} \int_{0}^{1} \ell_j(t) dt \  \boldsymbol{c}(\boldsymbol{x}_{k,j}, \boldsymbol{u}_k) \coloneqq \Delta \sum_{j=0}^{n_{\mathrm{p}}} \boldsymbol{B}_j \ \boldsymbol{c}(\boldsymbol{x}_{k,j}, \boldsymbol{u}_k),
\end{split}
\end{equation}
where $\boldsymbol{B}_{j} \in \mathbb{R}$ is a constant;
denote $\boldsymbol{B} \in \mathbb{R}^{n_{\mathrm{p}}+1}$ as the quadrature vector, where its element on the $j$-th row is $\boldsymbol{B}_{j}$.

Therefore, with the direction collocation, the continuous-time optimal control problem can be rewritten in discrete-time:
\begin{mini!}|s|
{(*)}{ \boldsymbol{h}(\boldsymbol{x}_{N,0}, t_{\mathrm{f}}) + \frac{t_{\mathrm{f}}}{N} \textstyle\sum_{k=0}^{N-1} \sum_{j=0}^{n_{\mathrm{p}}} \boldsymbol{B}_j \ \boldsymbol{c}(\boldsymbol{x}_{k,j}, \boldsymbol{u}_k) \label{oc_col:obj}}
{\label{oc_col}}{}
\addConstraint{ \frac{t_{\mathrm{f}}}{N} \boldsymbol{f}_{\mathrm{c}}(\boldsymbol{x}_{k,i}, \boldsymbol{u}_k, \boldsymbol{\beta}) - \textstyle\sum_{j=0}^{n_{\mathrm{p}}} \boldsymbol{C}_{j,i} \boldsymbol{x}_{k,j} = \boldsymbol{0} \label{oc_col:derivative_state}}
\addConstraint{ \boldsymbol{x}_{k+1,0} - \textstyle\sum_{j=0}^{n_{\mathrm{p}}} \boldsymbol{D}_{j} \boldsymbol{x}_{k,j} = \boldsymbol{0} \label{oc_col:continuity_state}}
\addConstraint{\forall k \in \llbracket 0, N-1 \rrbracket, i \in \llbracket 1, n_{\mathrm{p}} \rrbracket, \  \text{given } \boldsymbol{x}_0 \label{oc_col:initial_cond}}
\addConstraint{\text{other constraints on decision variables},  \label{oc_col:other_con}}
\end{mini!}
where $(*)$ denotes all the decision variables, i.e. $t_{\mathrm{f}} \in \mathbb{R}_+$, $\boldsymbol{x}_{k,i} \in \mathbb{R}^{2(n+3)}, \forall k \in \llbracket 0, N-1 \rrbracket, i \in \llbracket 1, n_{\mathrm{p}} \rrbracket$, $\boldsymbol{u}_k \in \mathbb{R}^{n+3}, \forall k \in \llbracket 0, N-1 \rrbracket$, and $\boldsymbol{x}_{N,0}$.

\section{Numerical Experiments} \label{sec:numerical_exp}

This section presents some numerical experiments on a mobile manipulator, including an example of integrated locomotion and manipulation planning and a comparison with a benchmark planning method.
This section aims to emphasize that the dynamic modeling of a mobile manipulator enables the integrated planning of both the base and the manipulator.
This integrated approach leads to faster movement compared to a benchmark planning method, where locomotion and manipulation planning are performed sequentially.

\subsection{Proposed: Integrated Planning}

This subsection introduces a concrete example of the integrated locomotion and manipulation planning problem \eqref{oc_col} with direct collocation:
\begin{mini!}|s|
{(*)}{ \frac{t_{\mathrm{f}}}{N} \textstyle\sum_{k=0}^{N-1} \sum_{j=0}^{n_{\mathrm{p}}} \boldsymbol{B}_j ||\boldsymbol{u}_k||_2^2 \label{oc_col_example:obj}}
{\label{oc_col_example}}{}
\addConstraint{ \frac{t_{\mathrm{f}}}{N} \boldsymbol{f}_{\mathrm{c}}(\boldsymbol{x}_{k,i}, \boldsymbol{u}_k, \boldsymbol{\beta}) - \textstyle\sum_{j=0}^{n_{\mathrm{p}}} \boldsymbol{C}_{j,i} \boldsymbol{x}_{k,j} = \boldsymbol{0} \label{oc_col_example:derivative_state}}
\addConstraint{ \boldsymbol{x}_{k+1,0} - \textstyle\sum_{j=0}^{n_{\mathrm{p}}} \boldsymbol{D}_{j} \boldsymbol{x}_{k,j} = \boldsymbol{0} \label{oc_col_example:continuity_state}}
\addConstraint{ \underline{\boldsymbol{x}} \leq \boldsymbol{x}_{k,i} \leq \overline{\boldsymbol{x}} }
\addConstraint{\forall k \in \llbracket 0,N-1 \rrbracket, i \in \llbracket 1, n_{\mathrm{p}} \rrbracket, \  \text{given } \boldsymbol{x}_0 \label{oc_col_example:initial_cond}}
\addConstraint{ \tau_{k,r} - \tau_{\mathrm{max},r}(Z_r\theta_{k,i,r}, \boldsymbol{\beta}_r) \leq 0 }
\addConstraint{ \tau_{\mathrm{min},r}(Z_r\theta_{k,i,r}, \boldsymbol{\beta}_r) - \tau_{k,r} \leq 0 }
\addConstraint{ \underline{\boldsymbol{f}_1} \leq \boldsymbol{f}_{k,1} \leq \overline{\boldsymbol{f}_1} }
\addConstraint{ \forall  k \in \llbracket 0,N-1 \rrbracket, i \in \llbracket 1, n_{\mathrm{p}} \rrbracket, r \in \llbracket 3,n+2 \rrbracket }
\addConstraint{ ||\boldsymbol{p}_{J_{\mathrm{e}}}(\boldsymbol{x}_{N,0}) - \boldsymbol{p}_{J_{\mathrm{e}},\mathrm{des}}||_2 = 0, \label{oc_col_example:terminal_constraint}}
\end{mini!}
where $(*)$ denotes all the decision variables, i.e. $\boldsymbol{x}_{k,i} \in \mathbb{R}^{2(n+3)}, \forall k \in \llbracket 0,N-1 \rrbracket, i \in \llbracket 1,n_{\mathrm{p}} \rrbracket$, $\boldsymbol{u}_k \in \mathbb{R}^{n+3}, \forall k \in \llbracket 0,N-1 \rrbracket$, and $\boldsymbol{x}_{N,0}$; $n_{\mathrm{p}} = 1$.
$\underline{\boldsymbol{x}}$ and $\overline{\boldsymbol{x}}$ represent the lower and upper bound of states, respectively.
$\tau_{k,r}$ represents Motor $r$'s torque from the input $\boldsymbol{u}_{k}$.
$\theta_{k,i,r}$ represents Joint $r$'s position from the state $\boldsymbol{x}_{k,i}$; $Z_r=50$ represents the gear ratio of Joint $r$.
$\tau_{\mathrm{min},r}$ and $\tau_{\mathrm{max},r}$ are given by \eqref{eq:motor_torque_max_positive}, \eqref{eq:motor_torque_max_0}, or \eqref{eq:motor_torque_max_negative}, depending on the sign of $\Phi_{\mathrm{pm},r}/L_{\mathrm{d},r} - I_{\mathrm{max},r}$.
$\underline{\boldsymbol{f}_1}$ and $\overline{\boldsymbol{f}_1}$ represent the lower and upper bound of the base's control input, respectively.
$\boldsymbol{f}_{k,1}$ represents the base's control from the input $\boldsymbol{u}_{k}$.
Eq. \eqref{oc_col_example:terminal_constraint} constrains the terminal state $\boldsymbol{x}_{N,0}$, where $\boldsymbol{p}_{J_{\mathrm{e}}}(\cdot)$ is a mapping from the state (joint positions) to the end effector position in Euclidean space and $\boldsymbol{p}_{J_{\mathrm{e}},\mathrm{des}} \in \mathbb{R}^3$ represents the desired end effector position. The details on $\boldsymbol{p}_{J_{\mathrm{e}}}(\cdot)$ are given by Section \ref{subsec:homo_transformation_define}.

The parameters related to the mobile manipulator are given as follows. $n=6$. The base mass is 90 kg, the manipulator mass (including the motors) is 47.03 kg, and the end effector is carrying a 5 kg solid iron ball. $\overline{\boldsymbol{f}_1} = - \underline{\boldsymbol{f}_1} = \matt{ 150 \text{ Nm} & 150 \text{ N} & 150 \text{ N} }^{\top}$, which can generate an acceleration of approximately 1 $\text{m/s}^2 \approx 0.1$ g in the x- and y-axis. Given some empirical motor design parameters, the constant maximum torque is 13 Nm and starts decreasing at 168 rad/s until reaching the maximum speed at 301 rad/s. The base's initial yaw angle is 0 rad; the base's initial positions in the x- and y-axis are 0 m; the initial and the desired terminal velocities for every joint of the mobile manipulator are all 0.

\begin{figure}[!ht]
\centering
\includegraphics[width=0.95\linewidth]{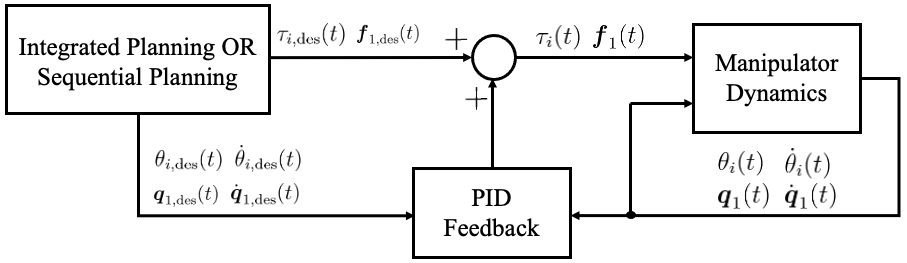}
\caption{The closed-loop system with PID feedback for both the integrated motion planning approach and the sequential planning approach.}
\label{fig:closed_loop_sys}
\end{figure}

During the simulation, after an optimal trajectory is obtained by solving \eqref{oc_col_example}, a PID (Proportional–Integral–Derivative) controller is adopted to track the desired torques from the optimal trajectory, i.e. for $i \geq 3$,
\begin{equation} \label{eq:pid_arm}
\begin{split}
&\tau_{i}(t) \coloneqq  \tau_{i,\mathrm{des}}(t) + K_{\mathrm{P},i}\Big(\theta_{i,\mathrm{des}}(t) - \theta_i(t)\Big) + K_{\mathrm{I},i} \\
& \int_0^t \Big(\theta_{i,\mathrm{des}}(t) - \theta_i(t)\Big) dt + K_{\mathrm{D},i}\Big(\dot{\theta}_{i,\mathrm{des}}(t) - \dot{\theta}_i(t)\Big),
\end{split}
\end{equation}
where $K_{\mathrm{P},i},  K_{\mathrm{I},i},  K_{\mathrm{D},i} > 0$ are the proportional, integral, and derivative gains, respectively; $\theta_{i,\mathrm{des}}(t)$ and $\dot{\theta}_{i,\mathrm{des}}(t)$ are the desired joint position and velocity from the optimal trajectory. Similarly, for the base, a similar PID controller is used, i.e.
\begin{equation} \label{eq:pid_base}
\begin{split}
&\boldsymbol{f}_1(t) \coloneqq \boldsymbol{f}_{1,\mathrm{des}}(t) + \boldsymbol{K}_{\mathrm{P},1}\Big(\boldsymbol{q}_{1,\mathrm{des}}(t) - \boldsymbol{q}_1(t)\Big) + \boldsymbol{K}_{\mathrm{I},1} \\
& \int_0^t \Big(\boldsymbol{q}_{1,\mathrm{des}}(t) - \boldsymbol{q}_1(t)\Big) dt + \boldsymbol{K}_{\mathrm{D},1}\Big(\dot{\boldsymbol{q}}_{1,\mathrm{des}}(t) - \dot{\boldsymbol{q}}_1(t)\Big),
\end{split}
\end{equation}
where $\boldsymbol{K}_{\mathrm{P},1}, \boldsymbol{K}_{\mathrm{I},1}, \boldsymbol{K}_{\mathrm{D},1} \in \mathbb{R}^{3 \time 3}$ are the PID gain diagonal matrices with positive diagonal elements. The closed-loop system result from the integrated motion planning approach is illustrated in Fig.~\ref{fig:closed_loop_sys}, where the PID feedback is given in \eqref{eq:pid_arm} and \eqref{eq:pid_base}. Note that for the integrated motion planning, $\tau_{i,\mathrm{des}}(t)$ and $\boldsymbol{f}_{1,\mathrm{des}}(t)$ in \eqref{eq:pid_arm} and \eqref{eq:pid_base} are the feed-forward terms and given by the optimal solution of \eqref{oc_col_example}.
The integrated motion planning problems with different parameters are denoted as:
\begin{enumerate}[{(1)}]
\item P1: the proposed integrated planning \eqref{oc_col_example} with final time $t_{\mathrm{f}}=5.0$ s and number of intervals $N = 500$;
\item P2: the same as P1, but $t_{\mathrm{f}}=3.5$ s, $N = 350$;
\item P3: the same as P1, but $t_{\mathrm{f}}=2.0$ s, $N = 220$.
\end{enumerate}
Note that $t_{\mathrm{f}}=2.0$ s is the minimal dynamically feasible time for this particular example given the full dynamics of the mobile manipulator. This is obtained by solving a time optimal integrated motion planning problem, which is similar to \eqref{oc_col_example}, but adding $t_{\mathrm{f}}$ as one decision variable and revising the objective function \eqref{oc_col_example:obj} as just $t_{\mathrm{f}}$.

\subsection{Comparison of Integrated Motion Planning with Analytical and Numerical Derivatives} \label{subsec:compare_numerical_grad}

To demonstrate the importance of using differentiable dynamical modeling, this subsection provides a comparison between solving the integrated motion planning optimization using the analytical gradients from the differentiable dynamic model and using the numerical gradients from the finite difference method.
Specifically, this paper uses CasADi \cite{andersson2019casadi} to construct the analytical dynamic model with Algorithm \ref{alg:ABA_mobile}, and employs the interface between CasADi and the nonlinear optimization solver IPOPT \cite{wachter2006implementation} to automatically provide the analytical gradients of the objective function and the equality/inequality constraints (including the analytical gradients of the forward dynamics), as well as the analytical Hessian matrix of the Lagrangian, to the IPOPT solver.
As for the numerical gradient approach, the numerical derivatives are provided by the central finite difference method in CasADi, and the Hessian matrix approximation is given by the limited-memory quasi-Newton approximation provided by IPOPT. The maximum iteration for both approaches is 1500 and the optimization initial guess for both approaches is a zero vector.

The desired end effector position is randomly generated given the constant distance of 2.25 m and the constant height of 0.5 m to the initial base position. Each approach is performed for the same random end effector desired position for 20 different trials and the result is summarized in Table \ref{table:optimization_grad_compare}, where Iteration Count indicates the total IPOPT iterations; n\_call\_nlp\_f indicates the number of function calls for the objective function; similarly, n\_call\_nlp\_g for the constraints, n\_call\_nlp\_grad\_f for the objective function gradient, n\_call\_nlp\_jac\_g for the constraints' gradients, n\_call\_nlp\_hess\_l for the Hessian matrix of the Lagrangian; $||\boldsymbol{u}||_2$ indicates the total summation of control magnitude over time; Error indicates the error between the final actual end effector position and the desired one on in millimeters given the closed-loop PID tracking controllers; Success indicates the success rate of solving optimization successfully.

Table \ref{table:optimization_grad_compare} demonstrates the significance of using a differentiable dynamics model for solving motion planning optimization problems, regarding computational efficiency. The differentiable dynamics model could provide the exact Hessian matrix and exact gradients for the optimization, which results in less total amount of function calls and thereby reduces the total computing time. Moreover, for interior-point methods, providing the exact Hessian matrix makes the algorithm converge faster than using the Quasi-Newton method with the limited memory BFGS (Broyden-Fletcher-Goldfarb-Shanno) update \cite{wachter2006implementation}. As for the optimality, according to the same result from $||\boldsymbol{u}||_2$ and Error, the optimization with analytical and numerical gradients in this comparison yields the same optimality, i.e. the same control effort.

\begin{table}
\centering
\begin{threeparttable}
\caption{Motion Planning with Analytical and Numerical Gradients}\label{table:optimization_grad_compare}
\begin{tabular}{c | c c}
\toprule
Item & P2 & P2$^\dagger$ \\
\midrule
IPOPT Time [s] & $\bm{28.13\pm1.07}$ & 267.21$\pm$24.42 \\
Iteration Count & $\bm{57.6\pm2.4}$ & 679.6$\pm$50.6 \\
n\_call\_nlp\_f & $\bm{758\pm375}$ & 16795$\pm$8315 \\
n\_call\_nlp\_g & $\bm{758\pm375}$ & 16795$\pm$8315 \\
n\_call\_nlp\_grad\_f & $\bm{630\pm340}$ & 7267$\pm$3863 \\
n\_call\_nlp\_jac\_g & $\bm{630\pm340}$ & 7267$\pm$3863 \\
n\_call\_nlp\_hess\_l & 610$\pm$329 & $\bm{0}$ \\
$||\boldsymbol{u}||_2$ & 14716$\pm$453 & 14716$\pm$453 \\
Error [mm] & 1.09$\pm$0.02 & 1.09$\pm$0.02 \\
Success [\%] & $\bm{100}$ & $\bm{100}$ \\
\bottomrule
\end{tabular}
\begin{tablenotes}
\small
\item[$\dagger$] with numerical gradients
\end{tablenotes}
\end{threeparttable}
\centering
\end{table}

\subsection{Benchmark: Sequential Planning}

\begin{figure}[!ht]
\centering
\includegraphics[width=0.45\linewidth]{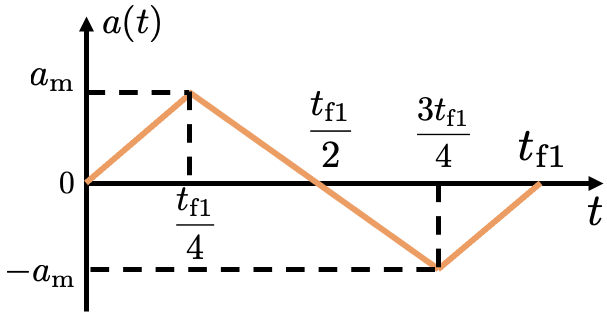}
\caption{An illustration on the acceleration parameterization.}
\label{fig:sequential_acceleration_traj}
\end{figure}

This subsection presents a benchmark approach, i.e., sequentially doing locomotion and manipulation planning. The sequential planning approach has two phases. In the first phase, the base moves to a location within a circle centering at the projection of $\boldsymbol{p}_{J_{\mathrm{e}},\mathrm{des}}$ on the ground (XOY plane). The base's desired position, velocity, and acceleration trajectory at the x- and y-axis are parameterized by a prescribed maximum acceleration magnitude $a_{\mathrm{m}} > 0$ and a desired final time $t_{\mathrm{f1}}$. The dynamic coupling between the base and the manipulator is essentially a wrench (or a spatial force) applied on both parts.
To reduce the effect of this wrench, as discussed inSection~\ref{subsec:literature_motion_planning}, the motion must minimize the jerk caused by sudden movements.
The trajectory parameterization is given by \eqref{eq:para_traj:1} - \eqref{eq:para_traj:3}, and its acceleration parameterization is illustrated in Fig. \ref{fig:sequential_acceleration_traj}.
The acceleration increases from 0 initially and decreases to 0 finally for smooth movement.

Without loss of generality (WLOG), the desired base movement at each axis is given by 
\begin{equation} \label{eq:para_traj:1}
a(t) = \begin{cases}
\frac{4a_{\mathrm{m}}}{t_{\mathrm{f1}}} t,& \text{if } t \in [0, \ \frac{t_{\mathrm{f1}}}{4}]\\
\frac{-4a_{\mathrm{m}}}{t_{\mathrm{f1}}} t + 2a_{\mathrm{m}},& \text{if } t \in [\frac{t_{\mathrm{f1}}}{4}, \ \frac{3t_{\mathrm{f1}}}{4}] \\
\frac{4a_{\mathrm{m}}}{t_{\mathrm{f1}}} t - 4a_{\mathrm{m}},& \text{if } t \in [\frac{3t_{\mathrm{f1}}}{4}, \ t_{\mathrm{f1}}]
\end{cases}
\end{equation}
\begin{equation} \label{eq:para_traj:2}
v(t) = \begin{cases}
\frac{2a_{\mathrm{m}}}{t_{\mathrm{f1}}} t^2,& \text{if } t \in [0, \ \frac{t_{\mathrm{f1}}}{4}]\\
\frac{-2a_{\mathrm{m}}}{t_{\mathrm{f1}}} (t-\frac{1}{2}t_{\mathrm{f1}})^2+\frac{a_{\mathrm{m}}t_{\mathrm{f1}}}{4},& \text{if } t \in [\frac{t_{\mathrm{f1}}}{4}, \ \frac{3t_{\mathrm{f1}}}{4}] \\
\frac{2a_{\mathrm{m}}}{t_{\mathrm{f1}}} (t-t_{\mathrm{f1}})^2,& \text{if } t \in [\frac{3t_{\mathrm{f1}}}{4}, \ t_{\mathrm{f1}}]
\end{cases}
\end{equation}
\begin{equation} \label{eq:para_traj:3}
p(t) = \begin{cases}
\frac{2a_{\mathrm{m}}}{3t_{\mathrm{f1}}} t^3,& \text{if } t \in [0, \ \frac{t_{\mathrm{f1}}}{4}]\\
\frac{-2a_{\mathrm{m}}}{3t_{\mathrm{f1}}} (t-\frac{1}{2}t_{\mathrm{f1}})^3+\frac{4a_{\mathrm{m}}t_{\mathrm{f1}} t - a_{\mathrm{m}}t_{\mathrm{f1}}^2}{16},& \text{if } t \in [\frac{t_{\mathrm{f1}}}{4}, \ \frac{3t_{\mathrm{f1}}}{4}] \\
\frac{2a_{\mathrm{m}}}{3t_{\mathrm{f1}}} (t-t_{\mathrm{f1}})^3 + \frac{a_{\mathrm{m}}t_{\mathrm{f1}}^2}{8},& \text{if } t \in [\frac{3t_{\mathrm{f1}}}{4}, \ t_{\mathrm{f1}}]
\end{cases}
\end{equation}
Note that $a(0) = a(t_{\mathrm{f1}}) = 0$; $v(0) = v(t_{\mathrm{f1}}) = 0$; WLOG, $v(t) > 0 \ \forall t \in (0, t_{\mathrm{f1}})$, $p(t) = 0$ and $p(t)$ is monotonically increasing in $[0, t_{\mathrm{f1}}]$.
Given a particular final position $p_{\mathrm{f}}$, the parameterized trajectory in \eqref{eq:para_traj:3} yields $t_{\mathrm{f1}} = \sqrt{8 p_{\mathrm{f}}/a_{\mathrm{m}}}$.
The desired yaw angle in the first phase is always the initial yaw angle, so as the desired yaw velocity, to reduce the influence of the base movement on the manipulator.

Since the dynamic coupling between two bodies is essentially an unmodeled wrench, this wrench can be viewed as a disturbance and one can design a feedback controller to reject this disturbance. Hence, a PID controller, similar to \eqref{eq:pid_base}, is used to track the desired trajectories for the base, where the feed-forward force on the x- and y-axis, i.e. the second and third element of $\boldsymbol{f}_{1,\mathrm{des}}(t)$, are given by the total mass of the robot multiplied by the desired acceleration defined in \eqref{eq:para_traj:1}; the feed-forward torque around the z-axis, i.e. the first element of $\boldsymbol{f}_{1,\mathrm{des}}(t)$, is zero.
To stabilize the manipulator while the base is moving, the desired motor torques are computed by the inverse dynamics of a fixed manipulator under the same actuator parameterization given desired zero joint velocities and accelerations.
A PID controller, similar to \eqref{eq:pid_arm}, is adopted to track the desired torques for the manipulator, where $\theta_{i,\mathrm{des}}(t) = \theta_{i}(0)$ and $\dot{\theta}_{i,\mathrm{des}}(t) = 0$.

In the second phase, once the base arrives at the desired location, a motion planning problem, similar to \eqref{oc_col_example}, is solved given the fixed manipulator dynamics and the prescribed time duration $t_{\mathrm{f2}}$. The PID controller for the manipulator in the first phase is used to track the desired torques returned by the optimization, where the feed-forward torques are the optimal torques returned by the fixed manipulator motion planning optimization. The PID controller for the base in the first phase is used to stabilize the base position at its desired one, with the feed-forward term $\boldsymbol{f}_{1,\mathrm{des}}(t)$ being zero. The closed-loop system of the sequential planning approach is also illustrated in Fig.~\ref{fig:closed_loop_sys}, where the difference between two approaches is how the feed-forward terms, i.e. $\tau_{i,\mathrm{des}}(t)$ and $\boldsymbol{f}_{1,\mathrm{des}}(t)$, are calculated.
The sequential planning problems with different parameters are denoted by:
\begin{enumerate}[{(1)}]
\item B1: the sequential planning with $\boldsymbol{a}_{\mathrm{m}} = \matt{0.25 & 0.25}^{\top}$ m/s$^2$;
\item B2: the same as B1, but $\boldsymbol{a}_{\mathrm{m}} = \matt{0.50 & 0.50}^{\top}$ m/s$^2$;
\item B3: the same as B1, but $\boldsymbol{a}_{\mathrm{m}} = \matt{0.75 & 0.75}^{\top}$ m/s$^2$;
\item B4: the same as B1, but $\boldsymbol{a}_{\mathrm{m}} = \matt{1.00 & 1.00}^{\top}$ m/s$^2$.
\end{enumerate}

\subsection{Comparison of Integrated and Sequential Planning Methods} \label{subsec:compare_two}

This subsection compares the motion planning results from the proposed integrated planning approach and the sequential planning approach given different parameters.
Each approach is tested for 20 trials, where the desired end effector position is randomly generated in the same way as Section \ref{subsec:compare_numerical_grad} does.
The results are summarized in Table \ref{table:planning_result}, where the column Time indicates the completion time in seconds. For the integrated motion planning method, only one set of PID gains is used for all $t_{\mathrm{f}}$. For the sequential planning method, only one set of PID gains is used throughout two phases, for all $\boldsymbol{a}_{\mathrm{m}}$; $t_{\mathrm{f2}}=2.5$ s is obtained by solving the time optimal motion planning problem given the fixed arm dynamics. Each set of gains is manually tuned to minimize the trajectory tracking error and the end effector position error.

According to Table \ref{table:planning_result}, the integrated planning method achieves time-optimal motion with millimeter-level end effector positioning errors, compared to the sequential planning. When compared with B1, the sequential planning approach with the smallest acceleration, P1 reduces the completion time by almost half while consuming similar control effort.
In fact, the integrated planning method with $t_{\mathrm{f}}=6$ s completes the task with a control effort of approximately 4166, consuming less energy than B1 while still being more time-efficient.
The optimal time of $t_{\mathrm{f}}=2$ s is obtained by solving the time-optimal integrated motion planning given the full dynamics of the mobile manipulator, which guarantees dynamical feasibility.
Thus, the motion planning with a final time greater than 2 s yields a 100\% success rate in solving the motion planning optimization.
By tuning the final time within the feasible range, one can customize motion planning to balance time and energy efficiency.

\begin{table}
\centering
\begin{threeparttable}
\caption{Results From Multiple Planning Problems} \label{table:planning_result}
\begin{tabular}{c c c c c}
\toprule
Name & Time [s] & $||\boldsymbol{u}||_2$ & Error [mm] & Success [\%] \\
\midrule
P1 & $5.0$ & 6691$\pm$171 & 1.17$\pm$1.10 & $\bm{100}$ \\
P2 & 3.5 & 13993$\pm$439 & $\bm{0.27\pm0.15}$ & $\bm{100}$ \\
P3 & $\bm{2.0}$ & 85705$\pm$6254 & 3.00$\pm$1.66 & $\bm{100}$ \\
B1 & 9.35$\pm$0.22 & $\bm{5453\pm65}$ & 38.68$\pm$9.83 & 90 \\
B2 & 7.34$\pm$0.16 & 15289$\pm$157 & 36.26$\pm$11.19 & 80 \\
B3 & 6.45$\pm$0.13 & 28017$\pm$264 & 34.64$\pm$11.93 & 85 \\
B4 & 5.93$\pm$0.11 & 43090$\pm$423 & 34.35$\pm$6.98 & 90 \\
\bottomrule
\end{tabular}
\end{threeparttable}
\centering
\end{table}

\begin{figure*}
\centering
\subfloat[P1.]
{\label{fig:2d_traj:5.0} \includegraphics[width=0.33\linewidth]{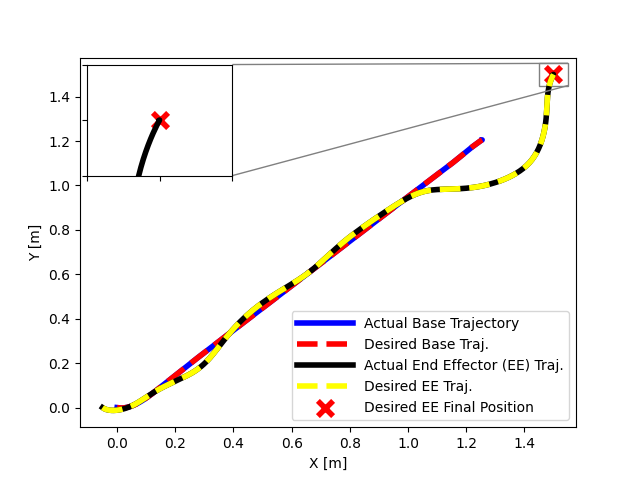}}
\subfloat[P2.]
{\label{fig:2d_traj:3.5} \includegraphics[width=0.33\linewidth]{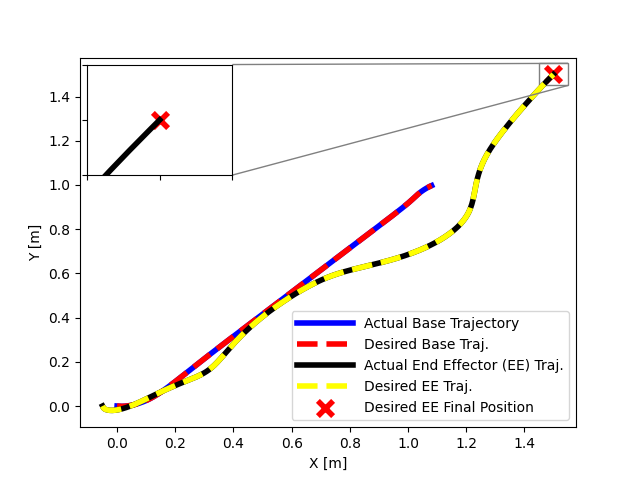}}
\subfloat[P3.]
{\label{fig:2d_traj:2.0} \includegraphics[width=0.33\linewidth]{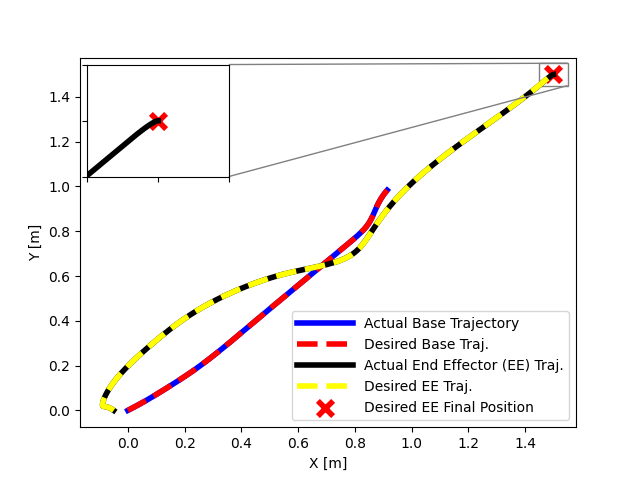}}
\hfill
\subfloat[B1.]
{\label{fig:2d_traj:a_0.25} \includegraphics[width=0.25\linewidth]{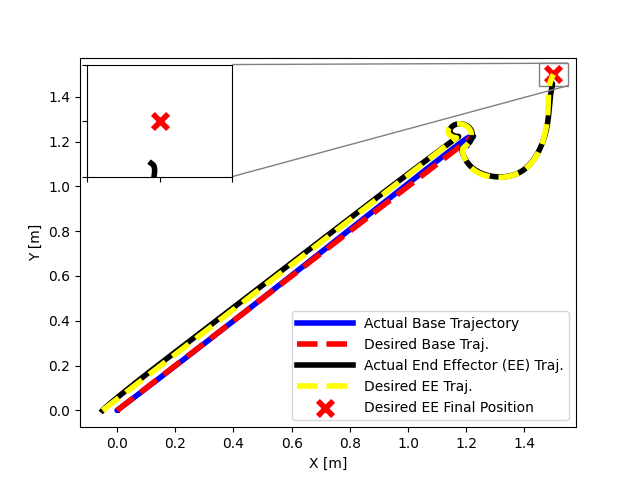}}
\subfloat[B2.]
{\label{fig:2d_traj:a_0.50} \includegraphics[width=0.25\linewidth]{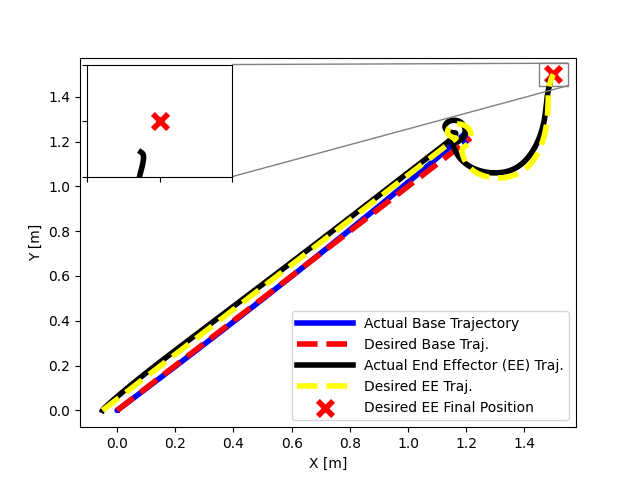}}
\subfloat[B3.]
{\label{fig:2d_traj:a_0.75} \includegraphics[width=0.25\linewidth]{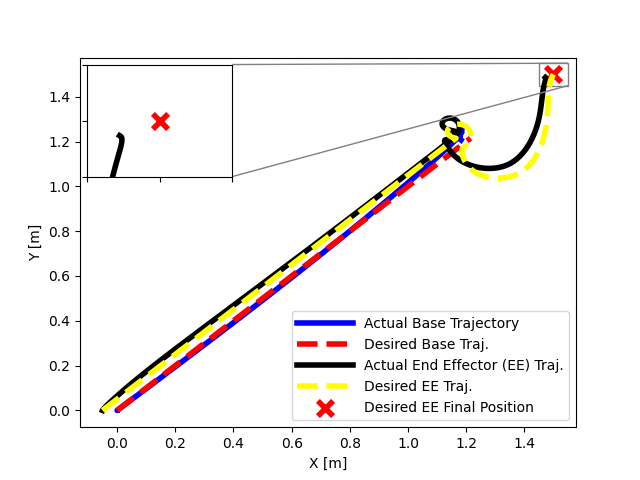}}
\subfloat[B4.]
{\label{fig:2d_traj:a_1.00} \includegraphics[width=0.25\linewidth]{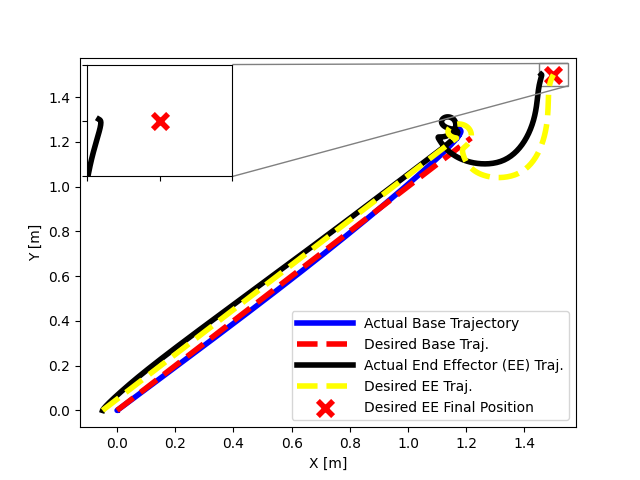}}
\caption{The desired and actual position trajectories of the base and the end effector on the XOY plane for different approaches.} \label{fig:2d_traj}
\end{figure*}

Fig. \ref{fig:2d_traj} illustrates how the mobile manipulator moves given different approaches and parameters. From Fig. \ref{fig:2d_traj:5.0} to Fig. \ref{fig:2d_traj:2.0}, the integrated planning with different final times all have precise trajectory tracking, and the zoom-in areas in all three subfigures show that the end effector reaches the desired goal position, even with the most aggressive but feasible task time of $t_{\mathrm{f}}=2$ s.
From Fig. \ref{fig:2d_traj:a_0.25} to Fig. \ref{fig:2d_traj:a_1.00}, where the sequential planning method is used, the manipulator fails to reach the desired position with the same precision.
This is because the computation of the parameterized motion does not account for the dynamic coupling between the base and the manipulator. Thus, the dynamical coupling is assumed small enough so that it, being treated as a disturbance, can be compensated by a tracking controller. However, this assumption holds only if the desired motion has a small acceleration.
Consequently, it is challenging for the sequential planning method to adjust the manipulator's configuration while the base is moving.
Therefore, when a more time-efficient movement is required, the sequential planning method cannot perform as well as the integrated planning method.

\begin{figure*}
\subfloat[Base Yaw, P1.]
{\label{fig:showcase:yaw_P1} \includegraphics[width=0.24\linewidth]{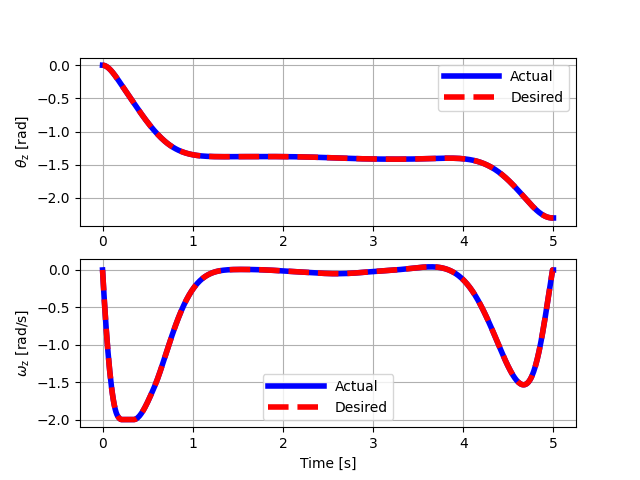}}
\hfill
\subfloat[Base Yaw, P3.]
{\label{fig:showcase:yaw_P3} \includegraphics[width=0.24\linewidth]{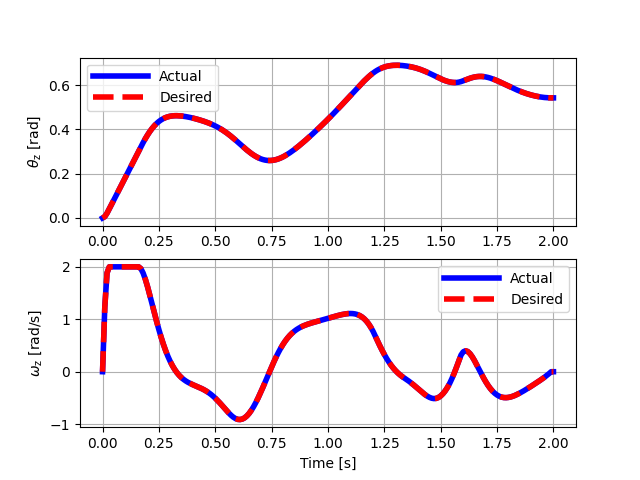}}
\hfill
\subfloat[Base Yaw, B1.]
{\label{fig:showcase:yaw_B1} \includegraphics[width=0.24\linewidth]{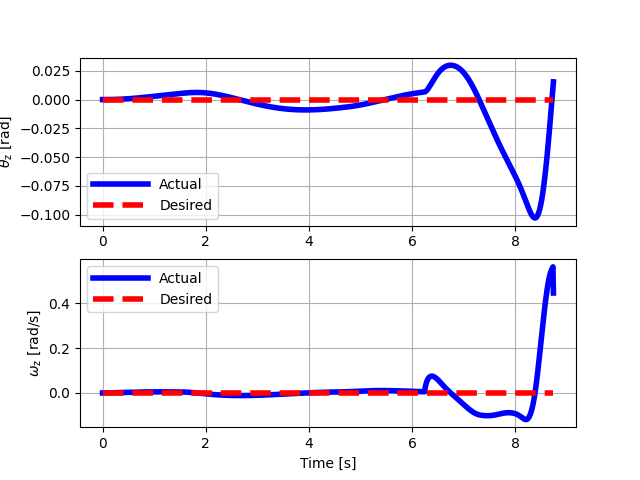}}
\hfill
\subfloat[Base Yaw, B4.]
{\label{fig:showcase:yaw_B4} \includegraphics[width=0.24\linewidth]{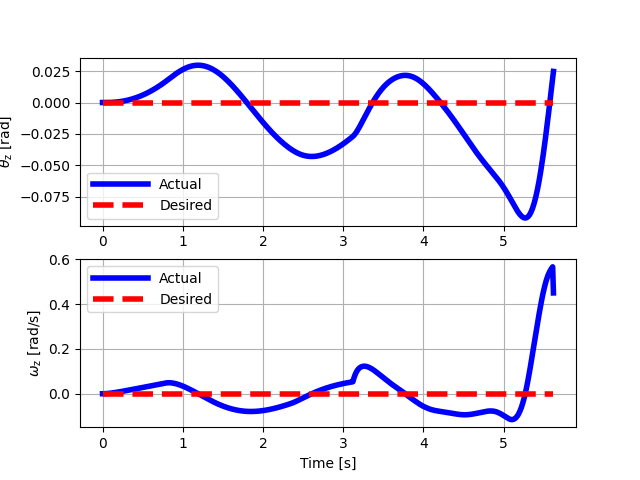}}
\hfill
\subfloat[Base X, P1.]
{\label{fig:showcase:x_P1} \includegraphics[width=0.24\linewidth]{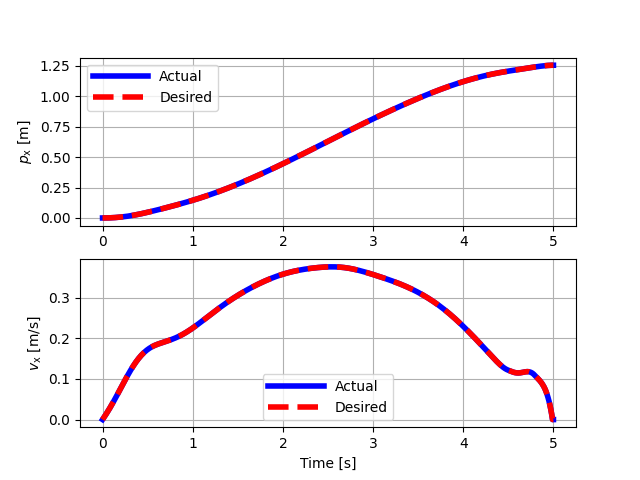}}
\hfill
\subfloat[Base X, P3.]
{\label{fig:showcase:x_P3} \includegraphics[width=0.24\linewidth]{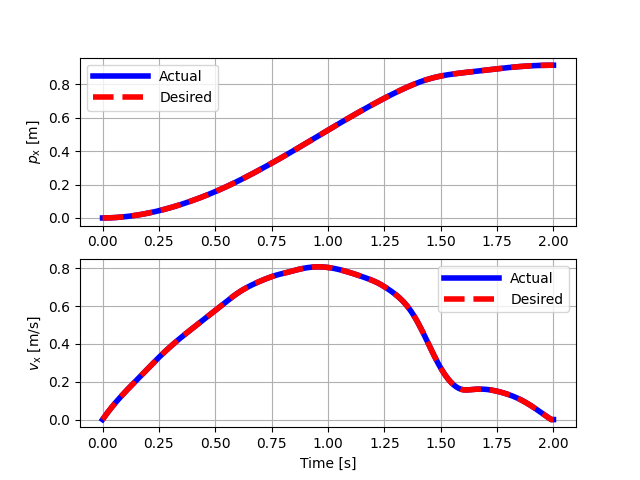}}
\hfill
\subfloat[Base X, B1.]
{\label{fig:showcase:x_B1} \includegraphics[width=0.24\linewidth]{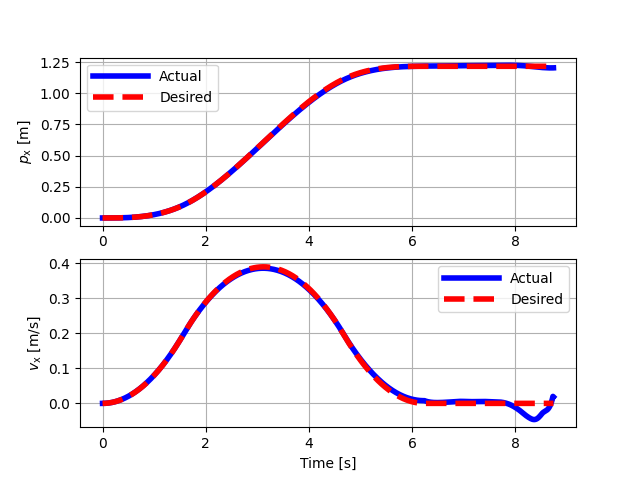}}
\hfill
\subfloat[Base X, B4.]
{\label{fig:showcase:x_B4} \includegraphics[width=0.24\linewidth]{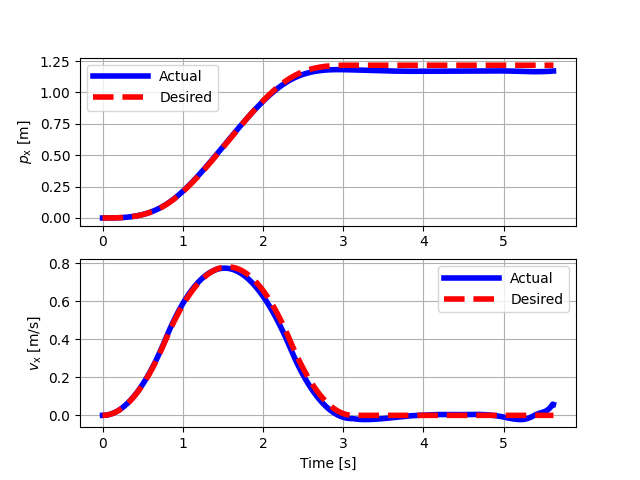}}
\hfill
\subfloat[Base Y, P1.]
{\label{fig:showcase:y_P1} \includegraphics[width=0.24\linewidth]{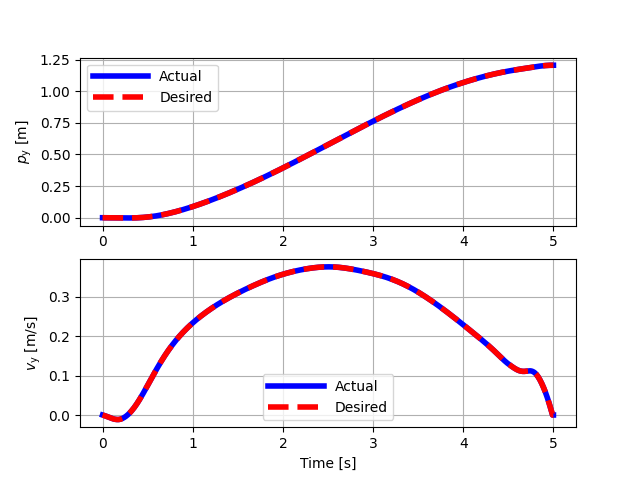}}
\hfill
\subfloat[Base Y, P3.]
{\label{fig:showcase:y_P3} \includegraphics[width=0.24\linewidth]{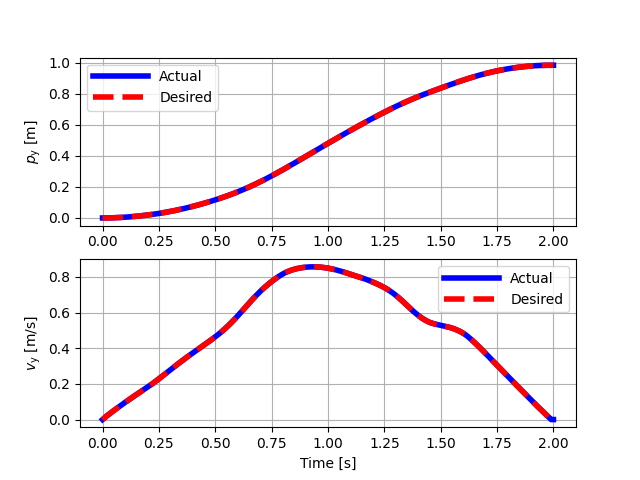}}
\hfill
\subfloat[Base Y, B1.]
{\label{fig:showcase:y_B1} \includegraphics[width=0.24\linewidth]{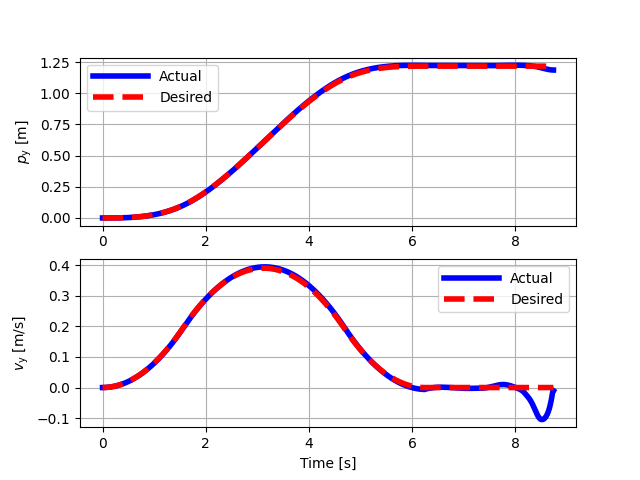}}
\hfill
\subfloat[Base Y, B4.]
{\label{fig:showcase:y_B4} \includegraphics[width=0.24\linewidth]{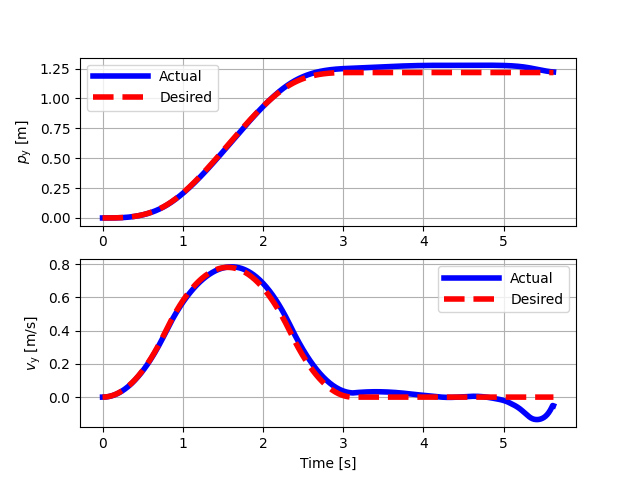}}
\hfill
\subfloat[Joint 4, P1.]
{\label{fig:showcase:j4_P1} \includegraphics[width=0.24\linewidth]{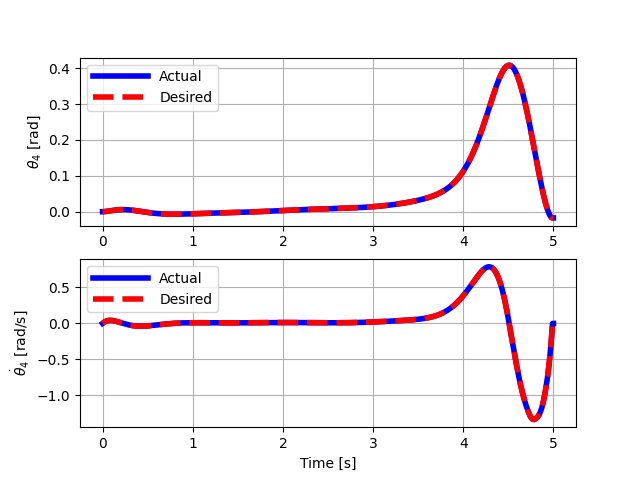}}
\hfill
\subfloat[Joint 4, P3.]
{\label{fig:showcase:j4_P3} \includegraphics[width=0.24\linewidth]{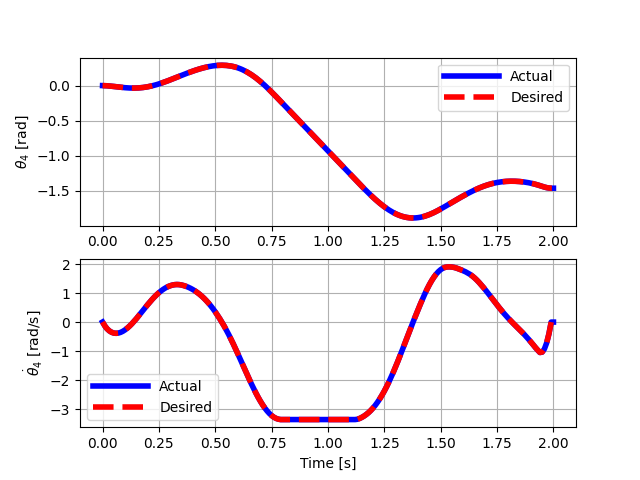}}
\hfill
\subfloat[Joint 4, B1.]
{\label{fig:showcase:j4_B1} \includegraphics[width=0.24\linewidth]{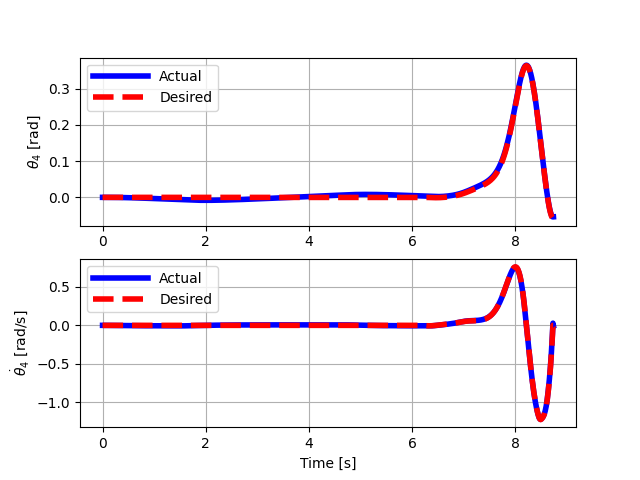}}
\hfill
\subfloat[Joint 4, B4.]
{\label{fig:showcase:j4_B4} \includegraphics[width=0.24\linewidth]{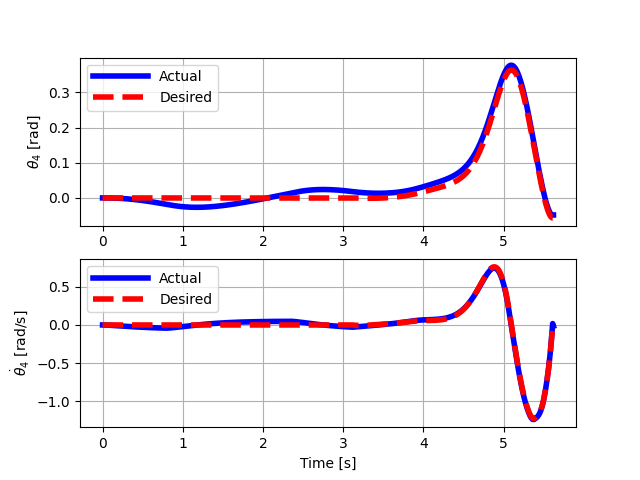}}
\hfill
\subfloat[Base Inputs, P1.]
{\label{fig:showcase:input_P1} \includegraphics[width=0.24\linewidth]{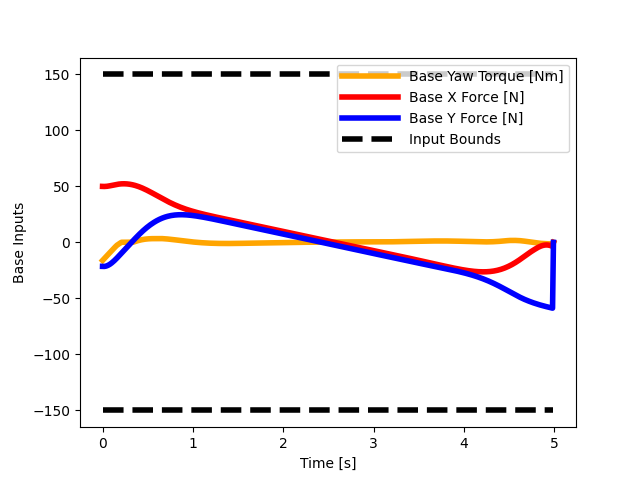}}
\hfill
\subfloat[Base Inputs, P3.]
{\label{fig:showcase:input_P3} \includegraphics[width=0.24\linewidth]{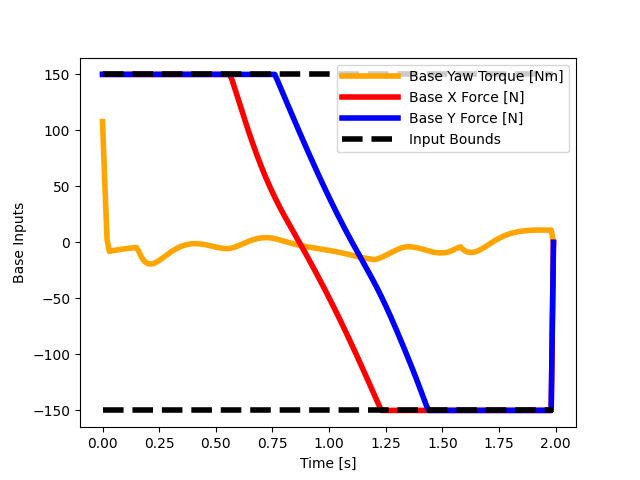}}
\hfill
\subfloat[Base Inputs, B1.]
{\label{fig:showcase:input_B1} \includegraphics[width=0.24\linewidth]{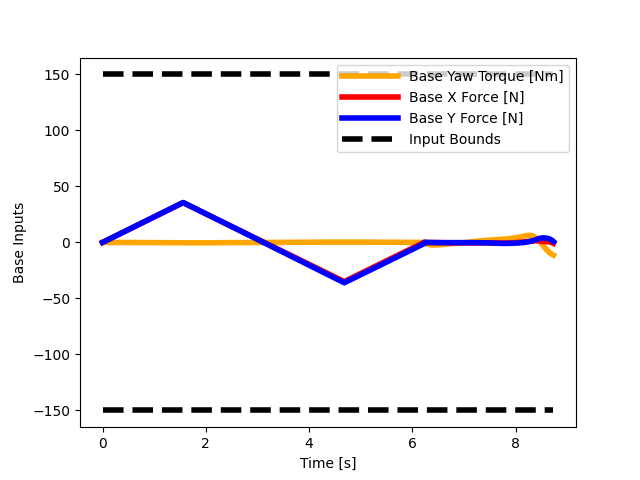}}
\hfill
\subfloat[Base Inputs, B4.]
{\label{fig:showcase:input_B4} \includegraphics[width=0.24\linewidth]{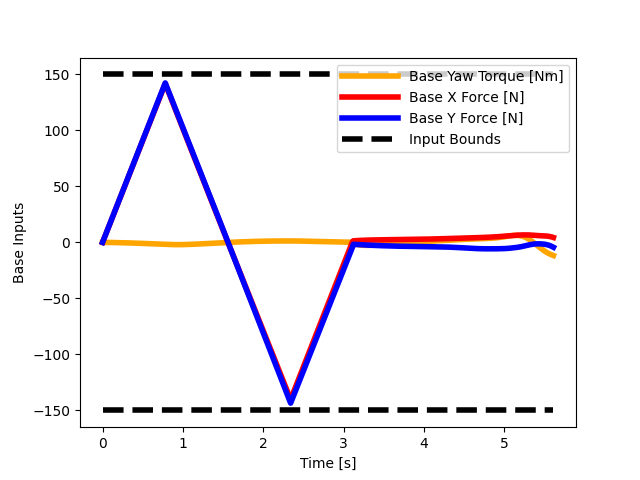}}
\caption{The state and control trajectories of some joints by different approaches.} \label{fig:showcase}
\end{figure*}

Fig. \ref{fig:showcase} demonstrates the state and control trajectories of some joints given different approaches. When comparing the base movement in Fig. \ref{fig:showcase:yaw_P1} - Fig. \ref{fig:showcase:y_B4}, the sequential planning method struggles to precisely track the desired state trajectory in the second phase.
This is due to the significant dynamical influence from the manipulator motion to the base, which is not considered and compensated for in the base motion.
Similarly, in Fig. \ref{fig:showcase:j4_P1} - Fig. \ref{fig:showcase:j4_B4}, the sequential planning method fails to precisely track Joint 4's desired trajectory due to the dynamical influence from the base motion to the manipulator. Note that Joint 4 is the first joint of the manipulator that needs to generate torque to compensate for the gravity of the entire manipulator. Thus, any small tracking error or dynamics mismatch in Joint 4 can be amplified for the rest of the joints.
Fig. \ref{fig:showcase:input_P1} - Fig. \ref{fig:showcase:input_B4} illustrates the control trajectory for different methods. Specifically, Fig. \ref{fig:showcase:input_P1} - Fig. \ref{fig:showcase:input_P3} demonstrates that the base yaw torque is constantly compensating for the dynamical effect between the base and the manipulator, especially for the most aggressive P3. This shows that for the sequential planning method, the desired yaw torque, i.e. the feed-forward term in the PID tracking controller, in both phases cannot constantly equal zero. Without modeling the dynamics of the entire mobile manipulator, one cannot calculate how much yaw torque is required to compensate for the motion.
In other words, the base yaw torque in Fig. \ref{fig:showcase:input_P3} compensates for the dynamic coupling wrench, whereas the sequential planning cannot compensate it well because it assumes that the desired wrench, i.e. the feed-forward term, is zero.
Part 1 of the supplementary video demonstrates how the mobile manipulator behaves given different approaches.

\subsection{Comparison with Measurement Noise}

\begin{figure*}
\centering
\subfloat[P1.]
{\label{fig:2d_traj_noise:5.0} \includegraphics[width=0.33\linewidth]{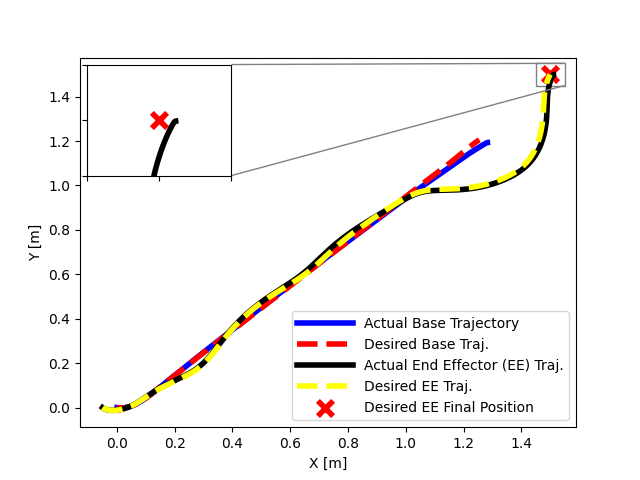}}
\subfloat[P2.]
{\label{fig:2d_traj_noise:3.5} \includegraphics[width=0.33\linewidth]{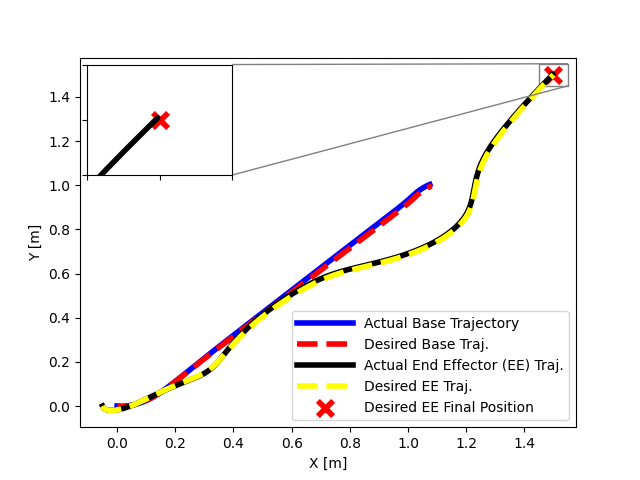}}
\subfloat[P3.]
{\label{fig:2d_traj_noise:2.0} \includegraphics[width=0.33\linewidth]{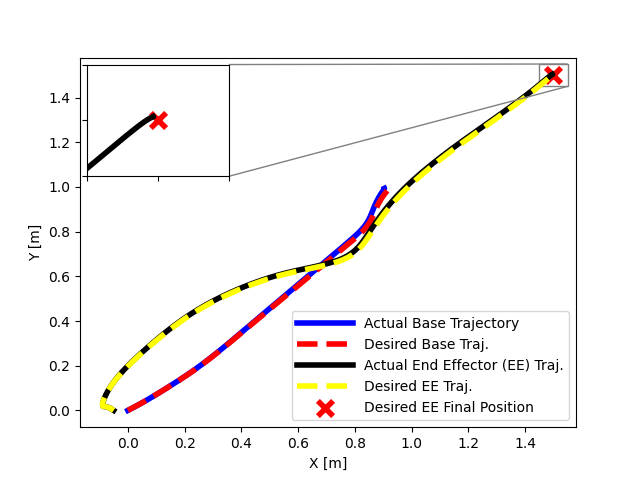}}
\hfill
\subfloat[B1.]
{\label{fig:2d_traj_noise:a_0.25} \includegraphics[width=0.25\linewidth]{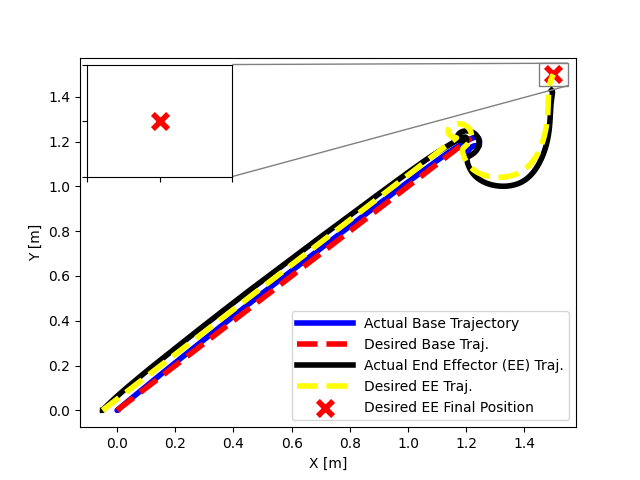}}
\subfloat[B2.]
{\label{fig:2d_traj_noise:a_0.50} \includegraphics[width=0.25\linewidth]{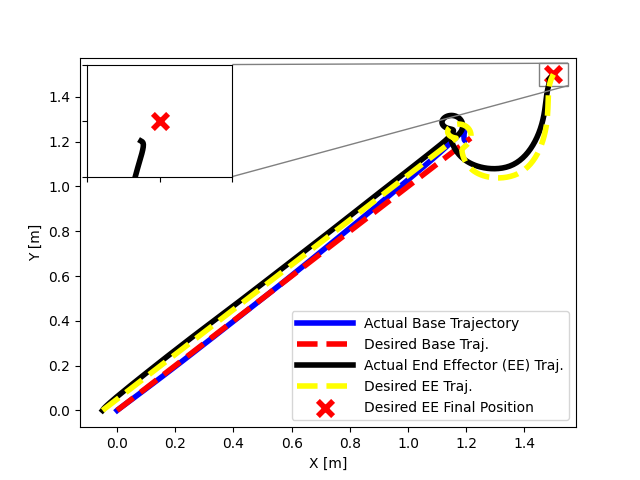}}
\subfloat[B3.]
{\label{fig:2d_traj_noise:a_0.75} \includegraphics[width=0.25\linewidth]{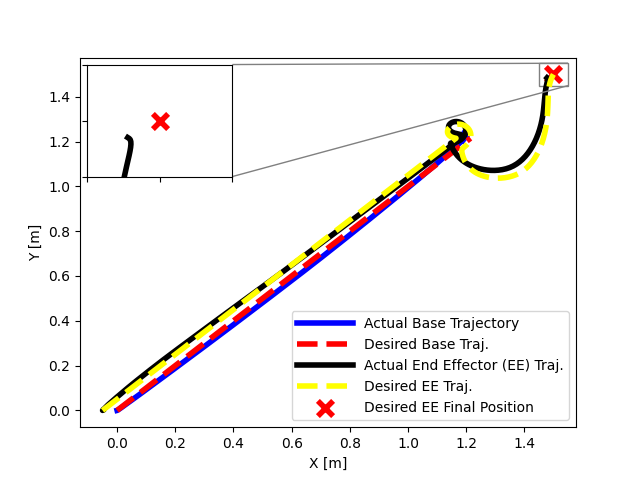}}
\subfloat[B4.]
{\label{fig:2d_traj_noise:a_1.00} \includegraphics[width=0.25\linewidth]{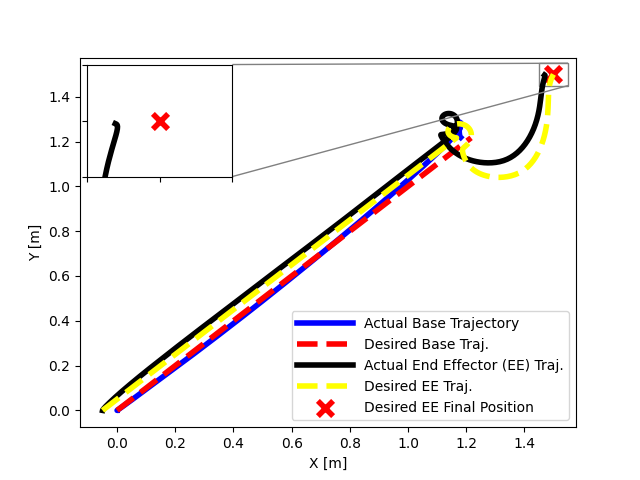}}
\caption{The desired and actual position trajectories of the base and the end effector on the XOY plane for different approaches with measurement noise.} \label{fig:2d_traj_noise}
\end{figure*}

This subsection compares the motion planning results given measurement noise.
The state observed at each time instance equals the true state plus a multivariate Gaussian noise with zero mean and a diagonal covariance matrix. In other words, each entry of the state follows an independent Gaussian distribution.
Particularly, the localization and pose estimation of the mobile base are less accurate than the measurement of the manipulator joints' angular positions and velocities.
The variance for joint angular positions and velocities is $1\times 10^{-5}$ rad due to accurate measurement. The variances for the base yaw angular position and velocity are $5.24\times 10^{-3}$ rad (0.3 degrees) and $1.75\times 10^{-3}$ rad/s (0.1 deg/s), respectively. The variances for the base position (along the x- and y-axis) and velocity are 0.02 m and 0.006 m/s, respectively.
Note that the measurement for the base is quite noisy, as 99.7\% of the position measurements in each axis lie within an error range of $\pm60$ millimeters.
The measured state is directly fed into the tracking controller of each approach, without any filtering or estimator.
Each approach is performed for 20 different trials which are randomly generated in the same way as Section \ref{subsec:compare_two} does.
The result is summarized in Table \ref{table:planning_noise}, where P2 and P3 perform similarly despite measurement noise.
Given that 99.7\% of the position measurements lie within an error range of $\pm60$ millimeters, the end effector error of P2 and P3 is acceptable.
Fig. \ref{fig:2d_traj_noise} illustrates how the mobile manipulator moves to reach the desired position given measurement noise.
Part 2 of the supplementary video demonstrates how the mobile manipulator tracks the desired trajectories under measurement noise.
Adopting a noise filter or a state estimator such as an Extended Kalman Filter (EKF) for trajectory tracking might further reduce the tracking error.

\begin{table}
\centering
\begin{threeparttable}
\caption{Results With Measurement Noise} \label{table:planning_noise}
\begin{tabular}{c c c c c}
\toprule
Name & Time [s] & $||\boldsymbol{u}||_2$ & Error [mm] & Success [\%] \\
\midrule
P1 & $5.0$ & 6697$\pm$173 & 45.48$\pm$23.82 & $\bm{100}$ \\
P2 & 3.5 & 13994$\pm$438 & 11.10$\pm$9.03 & $\bm{100}$ \\
P3 & $\bm{2.0}$ & 85551$\pm$6245 & $\bm{6.23\pm5.30}$ & $\bm{100}$ \\
B1 & 9.35$\pm$0.22 & $\bm{5990\pm329}$ & 57.96$\pm$26.04 & 90 \\
B2 & 7.34$\pm$0.16 & 15569$\pm$440 & 39.63$\pm$16.76 & 80 \\
B3 & 6.45$\pm$0.13 & 28261$\pm$363 & 35.68$\pm$12.85 & 85 \\
B4 & 5.93$\pm$0.11 & 43344$\pm$528 & 33.65$\pm$12.13 & 90 \\
\bottomrule
\end{tabular}
\end{threeparttable}
\centering
\end{table}

\section{Simultaneous Actuator Design and Integrated Motion Planning} \label{sec:simultaneous_design}

This section illustrates the effectiveness of the proposed modeling approach for mobile manipulator co-design, particularly by showcasing a simultaneous actuator design and motion planning framework. Specifically, simultaneous design involves solving a motion planning problem similar to \eqref{oc_col_example}, while also treating the motor design parameters $\boldsymbol{\beta}$ as decision variables, along with certain motor design constraints. The following subsections present the simultaneous design formulation and the corresponding numerical results.

\subsection{Formulation}

This subsection introduces the simultaneous design formulation.
First, Motor $j$'s motor design parameter $\boldsymbol{\beta}_j$ is subject to the following design constraints:
\begin{subequations} \label{eq:motor_design_constraints}
\begin{align}
&l_j \in [20, 100], r_{\mathrm{ro},j} \in [10, 100], r_{\mathrm{so},j} \in [10, 100], h_{\mathrm{m},j} \in [1, 5],\label{eq:design_bound:1}\\
&h_{\mathrm{sy},j} \in [5, 10], w_{\mathrm{tooth},j} \in [5, 20], b_{0,j} \in [1, 10], \label{eq:design_bound:2}\\
&h_{\mathrm{ss},j} > 0 \text{ mm}, \quad D_{\mathrm{wire},j} \geq 0.6 \text{ mm}, \quad  k_{\mathrm{C},j} > 0, \label{eq:con:4}\\
&0 < \arcsin(\textstyle\frac{w_{\mathrm{tooth},j}}{2(r_{\mathrm{ro},j}+\delta)}) + \arcsin(\textstyle\frac{b_{0,j}}{2(r_{\mathrm{ro},j}+\delta)}) \leq \textstyle\frac{\pi}{Q}, \label{eq:con_tooth_width}\\
&0 < m_{\mathrm{stator},j} + m_{\mathrm{rotor},j} \leq 3 \text{ kg}, \quad m_{\mathrm{stator},j} > 0 \text{ kg}, \label{eq:con_weight} \\
&0 \text{ T} < \textstyle\frac{k_{\mathrm{p}}\Phi_{1,j}}{w_{\mathrm{tooth},j}l_{j}} \leq 1.5\text{ T}, \quad 0 \text{ T} < \textstyle\frac{k_{\mathrm{p}}\Phi_{1,j}}{\sqrt{3}h_{\mathrm{sy},j}l_j} \leq 1.5 \text{ T}, \label{eq:con_magn_flux_yoke} \\
&A_{\mathrm{slot},j} > 0 \text{ mm}^2, \quad A_{\mathrm{so},j} > 0 \text{ mm}^2, \label{eq:con_slot_and_so_area}
\end{align}
\end{subequations}
where \eqref{eq:con:4} describe the minimal slot height, motor's minimal wire diameter, and minimal Carter's coefficient; \eqref{eq:con_tooth_width} and \eqref{eq:con_weight} describe the tooth width bound and the motor weight bound, respectively;
\eqref{eq:con_magn_flux_yoke} describes the magnetic flux bounds in the tooth and the stator yoke;
\eqref{eq:con_slot_and_so_area} describes the minimal slot area and the minimal cross-section area of the stator core.
$\delta$ is a constant defined in Section \ref{subsec:mec_modeling};
$h_{\mathrm{ss},j}$, $D_{\mathrm{wire},j}$, $m_{\mathrm{stator},j}$, $m_{\mathrm{rotor},j}$, $\Phi_{1,j}$, $k_{\mathrm{p}}$, $k_{\mathrm{C},j}$ can be calculated by \eqref{eq:ap:hss} - \eqref{eq:ap:mass_motor} and \eqref{eq:ap:carter} - \eqref{eq:ap:k_p_k_d} from Section \ref{subsec:mec_modeling}.
The design constraints \eqref{eq:design_bound:1} - \eqref{eq:design_bound:2} can be summarized as a closed convex set, i.e. $\boldsymbol{\beta}_j \in \boldsymbol{\mathcal{B}} \subset \mathbb{R}^7$; the design constraints \eqref{eq:con:4} - \eqref{eq:con_slot_and_so_area} can be written as an inequality constraint, i.e. $\boldsymbol{g}_{\mathrm{d}}(\boldsymbol{\beta}_j) \leq \boldsymbol{0}$.

Thus, the simultaneous actuator design and motion planning problem can be written as follows:
\begin{mini!}|s|
{(*)}{ \frac{t_{\mathrm{f}}}{N} \textstyle\sum_{k=0}^{N-1} \sum_{j=0}^{n_{\mathrm{p}}} \boldsymbol{B}_j ||\boldsymbol{u}_k||_2^2 \label{sim_design_motion:obj}}
{\label{sim_design_motion}}{}
\addConstraint{ \boldsymbol{\beta}_r \in \boldsymbol{\mathcal{B}}, \ \boldsymbol{g}_{\mathrm{d}}(\boldsymbol{\beta}_r) \leq \boldsymbol{0}}
\addConstraint{ \frac{t_{\mathrm{f}}}{N} \boldsymbol{f}_{\mathrm{c}}(\boldsymbol{x}_{k,i}, \boldsymbol{u}_k, \boldsymbol{\beta}) - \textstyle\sum_{j=0}^{n_{\mathrm{p}}} \boldsymbol{C}_{j,i} \boldsymbol{x}_{k,j} = \boldsymbol{0} \label{sim_design_motion:derivative_state}}
\addConstraint{ \boldsymbol{x}_{k+1,0} - \textstyle\sum_{j=0}^{n_{\mathrm{p}}} \boldsymbol{D}_{j} \boldsymbol{x}_{k,j} = \boldsymbol{0} \label{sim_design_motion:continuity_state}}
\addConstraint{ \underline{\boldsymbol{x}} \leq \boldsymbol{x}_{k,i} \leq \overline{\boldsymbol{x}} }
\addConstraint{\forall k \in \llbracket 0,N-1 \rrbracket, i \in \llbracket 1, n_{\mathrm{p}} \rrbracket, \  \text{given } \boldsymbol{x}_0 \label{sim_design_motion:initial_cond}}
\addConstraint{ \tau_{k,r} - \tau_{\mathrm{max},r}(Z_r\theta_{k,i,r}, \boldsymbol{\beta}_r) \leq 0 \label{sim_design_motion:max_tau}}
\addConstraint{ \tau_{\mathrm{min},r}(Z_r\theta_{k,i,r}, \boldsymbol{\beta}_r) - \tau_{k,r} \leq 0 \label{sim_design_motion:min_tau}}
\addConstraint{ \underline{\boldsymbol{f}_1} \leq \boldsymbol{f}_{k,1} \leq \overline{\boldsymbol{f}_1} }
\addConstraint{ \forall  k \in \llbracket 0,N-1 \rrbracket, i \in \llbracket 1, n_{\mathrm{p}} \rrbracket, r \in \llbracket 3,n+2 \rrbracket }
\addConstraint{ ||\boldsymbol{p}_{J_{\mathrm{e}}}(\boldsymbol{x}_{N,0}) - \boldsymbol{p}_{J_{\mathrm{e}},\mathrm{des}}||_2 = 0, \label{sim_design_motion:terminal_constraint}}
\end{mini!}
where $(*)$ denotes all the decision variables, i.e. $\boldsymbol{\beta} \in \mathbb{R}^{7n}$, $\boldsymbol{x}_{k,i}, \forall k \in \llbracket 0,N-1 \rrbracket, i \in \llbracket 1,n_{\mathrm{p}} \rrbracket$, $\boldsymbol{u}_k, \forall k \in \llbracket 0,N-1 \rrbracket$, and $\boldsymbol{x}_{N,0}$.

Since the simultaneous design optimization \eqref{sim_design_motion} includes motor design parameters $\boldsymbol{\beta}$ as decision variables, the objective function and the constraints of \eqref{sim_design_motion} need to be differentiable to $\boldsymbol{\beta}$. According to Section~\ref{subsec:torque_analytical_model}, the piece-wise analytical motor max/min torque functions from \eqref{eq:motor_torque_max_positive}, \eqref{eq:motor_torque_max_0} and \eqref{eq:motor_torque_max_negative} are not differentiable to $\boldsymbol{\beta}$.
However, as discussed in Section \ref{appendix:torque_eff_model}, when the motor speed exceeds $\omega_{\mathrm{r},j} = \omega_{\mathrm{ce},j}(\boldsymbol{\beta}_j)/p$, the maximum torque starts to decrease from its constant value.
This decrease occurs because either the voltage or the current increases, activating the voltage and/or current constraint.
From the perspective of minimizing the motor input power (i.e. the product of current and voltage), it is reasonable to constrain the motor speed to be less than $\omega_{\mathrm{ce},j}(\boldsymbol{\beta}_j)/p$.
Consequently, the motor's maximum torque remains constant, determined by a function of $\boldsymbol{\beta}$, regardless of the sign of $\Phi_{\mathrm{pm},j}/L_{\mathrm{d},j} - I_{\mathrm{max},j}$.
Therefore, for all Motor $r$, $r \in \llbracket 3, n+2 \rrbracket$, the motor torque constraints \eqref{sim_design_motion:max_tau} and \eqref{sim_design_motion:min_tau} are reduced to
\begin{equation}
\tau_{k,r} - \tau_{\mathrm{max},r}(\boldsymbol{\beta}_r) \leq 0, \quad -\tau_{\mathrm{max},r}(\boldsymbol{\beta}_r) - \tau_{k,r} \leq 0,
\end{equation}
where $\tau_{\mathrm{max},r} \coloneqq 1.5p \Phi_{\mathrm{pm},r}I_{\mathrm{max},r}$. For every time instance and collocation point, i.e. $\forall  k \in \llbracket 0,N-1 \rrbracket, i \in \llbracket 1, n_{\mathrm{p}} \rrbracket$, the joint speed is constrained by
\begin{equation}
\theta_{k,i,r} -\frac{\omega_{\mathrm{ce},r}(\boldsymbol{\beta}_r)}{Z_r p} \leq 0, \quad -\frac{\omega_{\mathrm{ce},r}(\boldsymbol{\beta}_r)}{Z_r p} -\theta_{k,i,r} \leq 0,
\end{equation}
where $\omega_{\mathrm{ce},r}(\boldsymbol{\beta}_r)$ is given by \eqref{eq:wce_define}.

\subsection{Numerical Results}

\begin{figure*}[h]
\centering
\includegraphics[width=0.95\linewidth]{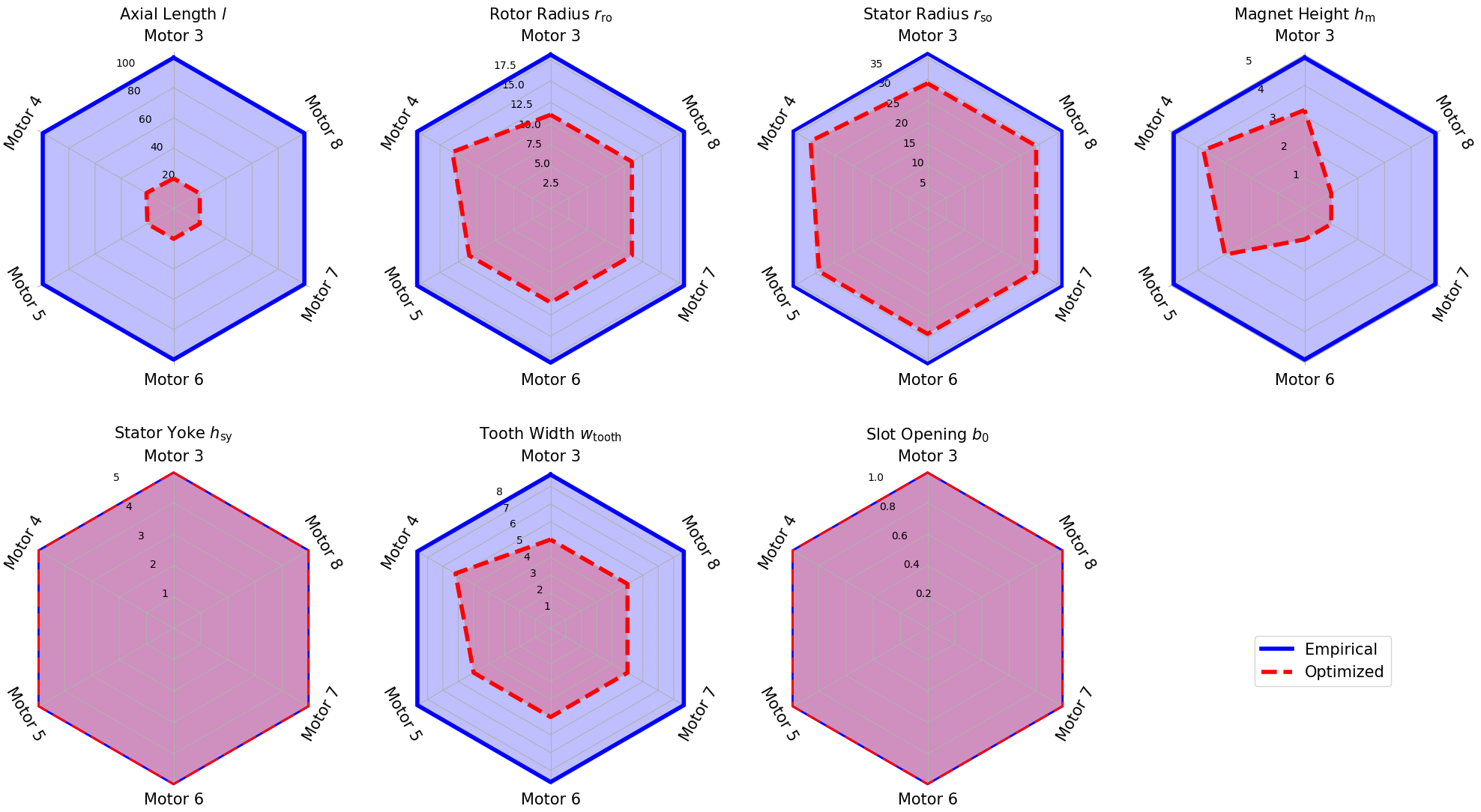}
\caption{Empirical and optimized motor design parameters.}
\label{fig:design_compare}
\end{figure*}

This subsection presents the numerical results of the simultaneous actuator design and motion planning.
The mobile manipulator is desired to carry a 5 kg solid iron ball to reach a terminal position. The final time is 2 seconds.
The initial guess of the motor design parameters is the empirical design used in Section \ref{sec:numerical_exp}.
The empirical and optimized motor designs are visualized in Fig. \ref{fig:design_compare}, where the magnet height and tooth width of all motors are decreased to reduce the unnecessary permanent flux $\Phi_{\mathrm{m}}$ and thereby motor torque capacity; the axial length, rotor radius, and stator radius of all motors are decreased to reduce the unnecessary motor mass and inertia.
Since Motor 4 is the heaviest motor in the arm to compensate for the entire arm's gravity, the rotor radius, magnet height, and tooth width of Motor 4 are greater than the others.

The motion planning results based on empirical and optimized motor designs are summarized in Table \ref{table:sim_design}. The total motor mass of the arm and the required control effort are significantly reduced by 86.44\% and 28.58\%, respectively, with the optimized design.
Comparing both the integrated motion planning and sequential planning summarized in Table \ref{table:planning_result} given the empirical design, the integrated motion planning with the optimized design consumes a similar amount of control effort as B4. But it achieves a more time-efficient motion, with its final time being almost one-third of B4's.
These results verify the effectiveness of the simultaneous design.
Part 3 of the supplementary video demonstrates how the mobile manipulator reaches the desired position given the empirical and optimized motor design.

\begin{table}
\centering
\begin{threeparttable}
\caption{Motion Planning Results with Empirical and Optimized Motor Design} \label{table:sim_design}
\begin{tabular}{c | c c c}
\toprule
Item & Empirical & Optimized & Reduced by \\
\midrule
Motor Mass [kg] & 18.00 & 2.44 & 86.44\% \\
$||\boldsymbol{u}||_2$ & 68667 & 49039 & 28.58\% \\
\bottomrule
\end{tabular}
\end{threeparttable}
\centering
\end{table}

While this specific simultaneous design scheme requires differentiability on design parameters, other co-design schemes, like the sequential co-design scheme in \cite{stein2023application}, do not.
For instance, one can formulate two optimization problems and solve them sequentially. The first is the motion planning optimization with specific motor design parameters, while the second is a motor design optimization to minimize operational efficiency based on the optimal trajectory obtained from the motion planning optimization.
In this case, the differentiability of the motor torque and speed constraints to the design parameters is relaxed for the motion planning optimization.
This section highlights the capability of the proposed modeling approach for robot co-design by showcasing the simultaneous actuator design and motion planning framework.

\section{Conclusions and Future Work} \label{sec:conclusion}

This paper investigates differentiable dynamic modeling for mobile manipulators, where actuators are parameterized by physically meaningful motor geometry parameters.
These parameters affect various aspects of the manipulator, such as link mass, inertia, center-of-mass, control constraints (motor torque capacity), and state constraints (motor maximum speed), all of which influence the performance of the motion planning and closed-loop control system.
This paper presents an analytical model for the motor's maximum torque and speed and describes how design parameters impact the dynamics, enabling differentiable and analytical dynamic modeling.
Furthermore, using the proposed differentiable dynamics and motor parameterization, this paper formulates an integrated locomotion and manipulation planning problem, discretized by direct collocation.
The effectiveness of the differentiable dynamics is demonstrated through numerical experiments, showing improvements in computation efficiency, task completion time, and energy consumption, compared to the prevailing sequential motion planning approach.
Lastly, this paper introduces a framework for simultaneous actuator design and motion planning, offering numerical results to validate the effectiveness of the proposed differentiable modeling approach for co-design problems.

The limitations of the proposed modeling approach can be summarized in three main aspects.
Firstly, this paper simplifies the mobile base as one rigid body with one 3-DOF planar joint on $\mathrm{SE}(2)$.
This choice is made to focus on the methodology and algorithm for modeling the forward and inverse dynamics of the entire chain of rigid bodies based on motor parameterization.
However, to model the full 6-degree-of-freedom motion of a mobile base, including pitch and roll behavior, a specific mechanical configuration for the mobile base, including wheel-ground contact, must be defined.
This would introduce additional modeling complexity beyond the scope of this paper.
For example, \cite{seegmiller2016high} investigate the high-fidelity modeling of a four-wheeled car-like mobile robot, including the modeling of contact forces between the wheels and the ground, as well as how the terrain affects the robot's dynamics. A future improvement of the proposed modeling approach could include similar techniques used in \cite{seegmiller2016high}.

Secondly, this paper neglects the dynamics of the gearboxes, typically harmonic drives, in the manipulator.
While harmonic drives are generally compact, and their dynamics are often negligible compared to the manipulator's dynamics, modeling a harmonic drive with major design parameters could increase the modeling fidelity for the entire robot. This would further enable the integrated design of motors and gearboxes.

Thirdly, the proposed modeling approach does not account for friction in each joint.
Actual friction occurs at the contact surface between each motor's rotor and its corresponding gearhead.
While some literature \cite{kermani2007friction,simoni2019modelling,pagani2020evaluation} models a manipulator's joint friction as a function of joint velocity, joint friction is also affected by the temperature of the contact surface, and any friction-related parameters could vary depending on the actual lubrication condition \cite{simoni2019modelling,pagani2020evaluation}.
Therefore, from the perspective of motion planning, it would be beneficial to consider unmodeled joint friction as an unknown disturbance. Designing planning and control techniques to compensate for this unknown disturbance is necessary for high-precision motion.
From the perspective of robot co-design, reasonably modeling joint friction could make the co-design result more realistic.
For instance, friction, as a function of joint velocity, may restrict fast joint movement to save energy or improve energy efficiency.

Regarding the proposed integrated locomotion and manipulation planning, this paper emphasizes that considering the entire dynamics of mobile manipulators could yield better motion performance compared to the most common sequential planning approach.
Modeling a motor's maximum torque/speed as analytical functions ensures that the corresponding motion does not violate the control authority of the motors.
Even though the proposed integrated motion planning is not a real-time algorithm, it provides insights into how the proposed modeling approach could benefit both offline and online motion planning.
A differentiable dynamic model is necessary for many state-of-the-art online motion planning algorithms, according to this paper's literature review.






\bibliography{ref}







\end{document}